\PassOptionsToPackage{table}{xcolor}
\PassOptionsToPackage{sort&compress}{natbib}
\documentclass[a4paper,onecolumn,10pt]{holoct-preprint}
\usepackage{amsmath,amssymb,amsfonts}
\usepackage{array,tabularx,adjustbox,ifoddpage}
\usepackage{longtable,hhline,makecell,pdflscape,ragged2e,tocloft}
\usepackage{chapterbib}
\setcitestyle{numbers,super,sort&compress,open={},close={}}
\providecommand{\bibcommenthead}{}
\makeatletter
\def\input@path{{supplementary/}}
\AtBeginDocument{%
    \patchcmd{\LT@output}{\vss}{\vfil}{}{}%
    \patchcmd{\LT@output}{\vss}{\vfil}{}{}%
}
\makeatother
\newcommand{\figref}[2][]{Fig.~\ref{#2}#1}
\newcommand{\edfigref}[2][]{Extended Data Fig.~\ref{#2}#1}

\newcommand{\supplink}[2]{\hyperlink{table.#1}{#2}}
\newcommand{\suppref}[1]{Supplementary Table~\supplink{#1}{#1}}
\newcommand{\supprefrange}[2]{Supplementary Tables~\supplink{#1}{#1}--\supplink{#2}{#2}}
\newcommand{\suppreftwo}[2]{Supplementary Tables~\supplink{#1}{#1} and \supplink{#2}{#2}}
\DeclareCaptionFont{importedportraitcaption}{%
    \fontsize{7.60pt}{8.70pt}\selectfont
}

\DeclareCaptionFont{importedlandscapecaption}{%
    \fontsize{6.20pt}{6.80pt}\selectfont
}

\DeclareCaptionLabelSeparator{tablepipe}{\enspace|\enspace}

\newcommand{\NaturalPageCenteredGraphic}[1]{%
    \centering\includegraphics[width=\linewidth]{#1}\par
}
\newcommand{\BeginNaturalFigurePages}{%
    \clearpage
    \newgeometry{left=13.5mm,right=13.5mm,top=15mm,bottom=17mm,
        headheight=0pt,headsep=0pt,footskip=8mm,bindingoffset=0mm}%
    \pagestyle{plain}%
    \captionsetup[figure]{format=plain,labelsep=period,font={small,stretch=1},
        labelfont=bf,justification=raggedright,singlelinecheck=false,skip=5pt}%
}
\newcommand{\EndNaturalFigurePages}{%
    \clearpage\restoregeometry\pagestyle{plain}%
}

\title{A 3D foundation model unifying CT interpretation and segmentation}
\author[1,2]{Yulu Xie}
\author[2]{Zilong Li}
\author[3,2]{Shen Xu}
\author[2]{Yifan Bai}
\author[4]{Liuhong Shi}
\author[5]{Jiao Chen}
\author[5]{Xianglei Kong}
\author[5]{Wenchao Zhu}
\author[6]{Na Shen}
\author[7]{Xianghua Ye}
\author[8]{Yang Yan}
\author[8]{Jiaju Zheng}
\author[9]{Yunsong Peng}
\author[2]{Yanzhou Su}
\author[2]{Chaohui Yu}
\author[2,10,11,\dagger]{Hao Luo}
\author[2,\dagger]{Fan Wang}
\affiliation[1]{Department of Statistics and Data Science, Tsinghua University, Beijing, China}
\affiliation[2]{Alibaba DAMO Academy, Hangzhou, China}
\affiliation[3]{School of Psychological and Cognitive Sciences and Beijing Key Laboratory of Behavior and Mental Health, Peking University, Beijing, China}
\affiliation[4]{Department of Head and Neck Surgery, Sir Run Run Shaw Hospital, Zhejiang University School of Medicine, Hangzhou, China}
\affiliation[5]{Department of Radiology, Sir Run Run Shaw Hospital, Zhejiang University School of Medicine, Hangzhou, China}
\affiliation[6]{Department of Otolaryngology-Head \& Neck Surgery, Zhongshan Hospital, Fudan University, Shanghai, China}
\affiliation[7]{Department of Radiation Oncology, The First Affiliated Hospital, Zhejiang University School of Medicine, Hangzhou, China}
\affiliation[8]{Department of Radiology, The Affiliated Yangming Hospital of Ningbo University, Yuyao, China}
\affiliation[9]{Guizhou Province International Science and Technology Cooperation Base for Precision Imaging Diagnosis and Treatment, Key Laboratory of Advanced Medical Imaging and Intelligent Computing of Guizhou Province, Department of Radiology, Guizhou Provincial People's Hospital, Guiyang, China}
\affiliation[10]{College of Computer Science and Technology, Zhejiang University, Hangzhou, China}
\affiliation[11]{Hupan Laboratory, Hangzhou, China}
\contribution[\dagger]{Corresponding authors}
\metadata[Code]{%
  \url{https://github.com/alibaba-damo-academy/HoloCT}%
}

\abstract{
Radiologists interpret computed tomography (CT) scans by jointly determining what a finding represents and where it lies in three-dimensional (3D) anatomy.
However, deep learning models for CT have largely treated CT interpretation and 3D segmentation as separate tasks, leaving textual findings spatially unverifiable and overlooking potential synergy between clinical understanding and spatial perception.
Here we present HoloCT, a generalist 3D foundation model unifying CT interpretation and voxel-level segmentation by jointly modelling whole-volume context and fine-grained anatomical detail, performing diverse clinical tasks while providing 3D masks as evidence.
To develop and evaluate HoloCT at scale, we curated HoloDATA, comprising 195,921 CT scans from 73 public datasets and six clinical cohorts, with 6.2 million instruction dialogues and 2.8 million CT--mask pairs.
Across 346 tasks spanning segmentation, lesion screening, visual question answering, and report generation on 61 internal and 12 external datasets, HoloCT outperformed both specialist and foundation models.
In multicentre validation across 11,455 clinical CT scans from five hospitals, HoloCT outperformed competing models in lesion segmentation and report generation by 18.2 and 15.3 points.
It also achieved 85.1\% mean accuracy in screening thyroid tumours and colon lesions, exceeding all four participating radiologists in a retrospective reader study.
Further analyses revealed bidirectional semantic--spatial synergy: language-guided segmentation delineated unseen anatomical structures and lesions as well as diffuse free-text findings, outperforming the leading approach by 22.1 points, whereas prompting the model to generate corresponding 3D masks alongside its textual outputs significantly improved interpretation performance.
In summary, HoloCT shows that clinical understanding and spatial perception can be learned together within a single foundation model and offers a path towards more accurate and reliable CT analysis.
}

\hypersetup{pdftitle={A 3D foundation model unifying CT interpretation and segmentation},pdfauthor={Yulu Xie et al.}}

\AddToHookNext{shipout/foreground}{%
    \begin{tikzpicture}[remember picture,overlay]
        \node[anchor=north east,inner sep=0pt]
            at ([xshift=-25mm,yshift=-7mm]current page.north east)
            {\includegraphics[height=8mm]{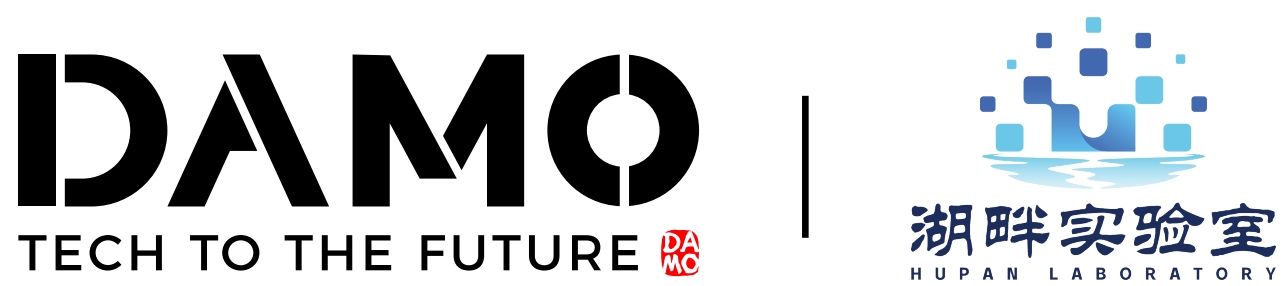}};
            \draw[black,line width=1pt]
    ([xshift=25mm,yshift=-19mm]current page.north west) --
    ([xshift=-25mm,yshift=-19mm]current page.north east);
    \end{tikzpicture}%
}

\begin{document}
\maketitle
\section{Introduction}\label{sec1}\label{sec:introduction}

Computed tomography (CT) provides rapid, high-resolution volumetric imaging for clinical screening, diagnosis, and treatment planning~\citep{smithbindman2019trends}.
When interpreting a scan, radiologists continuously link \emph{what} a finding represents to \emph{where} it lies in three-dimensional (3D) anatomy.
Reliable CT interpretation requires both: textual descriptions without anatomical localization are difficult to verify across hundreds of slices, and segmented regions without clinical characterization carry limited diagnostic meaning.

Existing deep learning systems have largely developed these two capabilities in isolation.
Segmentation networks delineate organs and abnormalities at the voxel level yet do not interpret their clinical significance~\citep{Isensee2021nnUNet,He2024VISTA3DAU,Zhao2023LargevocabularySF,Rokuss2025VoxTellFP}.
Conversely, CT vision--language models (VLMs) generate diagnostic text without providing voxel-level localization, making the anatomical basis of their predictions difficult to inspect~\citep{blankemeier2026merlin,Hamamci2024GeneralistFM,Wu2025TowardsGF}.
Recent grounded medical models connect language and segmentation at the output level~\citep{CandidateBiomedParse2025,CandidateMedVersa2026,CandidateMedPLIBjournalextension2026,Bai2024M3DA3,luo2025vividmed,wu2026unibiomed}; however, for 3D scans, existing models route high-resolution spatial features only to the mask decoder rather than the language model, limiting the use of fine-grained spatial information during text generation~\citep{Bai2024M3DA3,luo2025vividmed,wu2026unibiomed}.
Moreover, despite advances in voxel-level visual encoders~\citep{CandidateMedVAE2026}, many of these models still rely on 2D slice encoders or coarsely downsampled 3D inputs at the expense of cross-slice continuity or fine spatial detail~\citep{wu2026unibiomed,Bai2024M3DA3}.
Whether clinical understanding and spatial perception can be learned jointly within a single foundation model and mutually reinforce one another therefore remains an open question.

Here we introduce HoloCT, a generalist 3D foundation model that unifies CT interpretation and 3D segmentation by jointly modelling whole-volume context and fine-grained anatomical detail within a unified vision--language representation (\figref{fig:main}).
Specifically, HoloCT couples a dual-scale visual encoding module, a unified multimodal large language model, and a mask decoder.
First, each scan is encoded through two parallel branches: a global encoder for whole-volume context and a detailed encoder that preserves fine-grained anatomical detail at standardized isotropic spacing.
Second, global tokens and fine-grained spatial tokens from the two encoders are supplied directly to the language model, enabling clinical reasoning to draw on fine-grained spatial features.
Third, segmentation tokens emitted by the language model guide the mask decoder for voxel-level delineation.
Accordingly, a single HoloCT model performs language-guided segmentation, lesion screening, visual question answering (VQA), and radiology report generation.
For grounded interpretation, HoloCT can pair its textual outputs with 3D masks of the corresponding anatomical regions or findings, providing spatial evidence that readers can inspect.

To develop and evaluate HoloCT at scale, we curated HoloDATA by converting segmentation annotations, question--answer pairs, and radiology reports across datasets into a unified instruction--response format, with 3D masks incorporated where available.
HoloDATA comprises 195,921 CT scans totalling more than 61 million slices from 73 public datasets and six clinical cohorts at five hospitals, and contains 6.2 million instruction dialogues and 2.8 million CT--mask pairs across 246 anatomical structures and 50 lesion categories.

Evaluated on 346 tasks across 11 task formulations on 61 internal and 12 external datasets, a single HoloCT-8B model outperformed 82 dataset-specific nnU-Net models on internal segmentation benchmarks and surpassed segmentation foundation models in zero-shot external transfer across both seen and unseen categories of anatomical structures and lesions.
HoloCT also exceeded task-specific classifiers as well as medical and general-purpose VLMs in lesion screening, VQA, and report generation.
In multicentre clinical validation across 11,455 CT scans from six cohorts at five hospitals, HoloCT surpassed competing models in the segmentation of five lesion categories, including laryngeal lesions, lung cancer and thyroid tumours, as well as in zero-shot abdominal and chest CT report generation.
In retrospective reader studies of histopathology-confirmed thyroid tumours and colon lesions, HoloCT-8B achieved higher sensitivity and accuracy than all four participating radiologists, including two senior readers.

Beyond benchmark performance, our analyses revealed bidirectional synergy between clinical understanding and spatial perception.
During training, adding text-only dialogues improved 3D segmentation accuracy, whereas incorporating mask-paired dialogues improved lesion screening, VQA, and report generation.
At inference, language-guided segmentation delineated previously unseen anatomical structures and lesions as well as diffuse free-text findings, outperforming the leading approach by 22.1 points; conversely, prompting HoloCT to generate corresponding 3D masks alongside its textual outputs significantly improved performance across these interpretation tasks while maintaining mask quality.
Together, these results demonstrate that unifying clinical understanding and spatial perception within a single foundation model strengthens both capabilities, aligning AI with how radiologists link diagnostic characterization to 3D anatomy when interpreting CT scans.

\section{Results}\label{sec2}\label{sec:results}

\subsection{Overview of HoloCT and HoloDATA}\label{sec:results-overview}

\begin{figure}[p]
    \NaturalPageCenteredGraphic{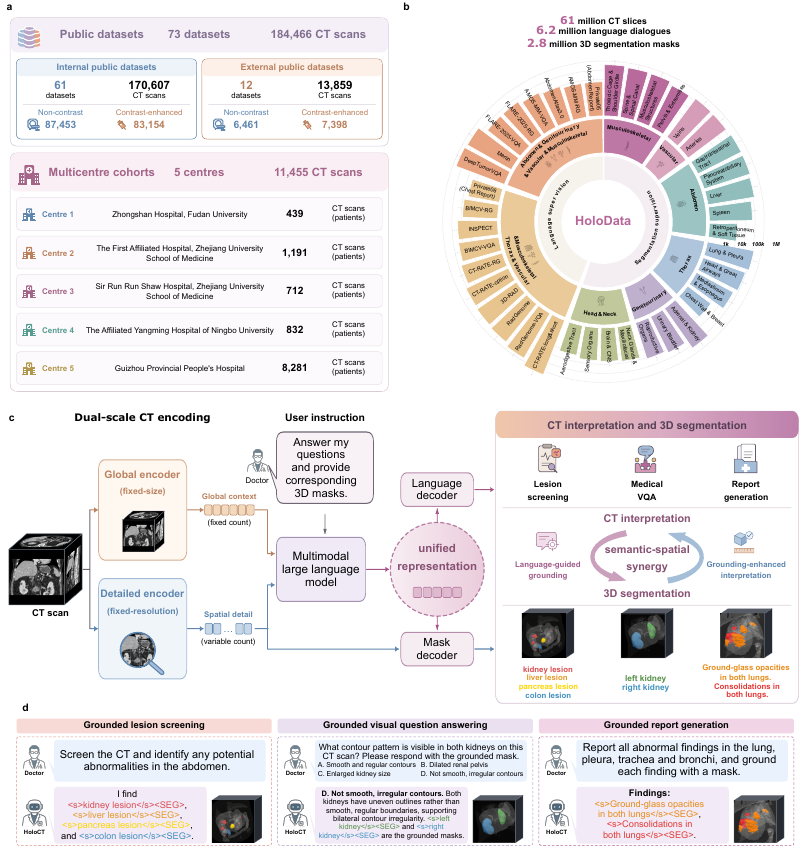}
\caption{\textbf{HoloDATA and HoloCT for unified CT interpretation and segmentation.}
\textbf{a}, Public datasets and multicentre cohorts comprising 195,921 CT scans, grouped by source and contrast status.
\textbf{b}, Language and segmentation annotations across datasets and anatomical systems.
\textbf{c}, Dual-scale encoding combines global context and high-resolution spatial tokens in a multimodal language model for text generation and 3D mask decoding.
\textbf{d}, Representative dialogues and masks for simultaneous multi-target lesion screening, grounded visual question answering (VQA) and finding-level grounded report generation.}
\label{fig:main}
\end{figure}

HoloCT is a generalist 3D foundation model developed and evaluated on the HoloDATA corpus to unify CT interpretation and voxel-level segmentation across volumetric CT tasks (\figref{fig:main}).
HoloDATA comprises 195,921 volumetric CT scans with more than 61 million slices, integrating 184,466 scans from 73 public datasets with 11,455 scans from six clinical cohorts at five hospitals (\figref[a]{fig:main}).
Of the 73 public datasets, 61 internal datasets supported model development and held-out testing, whereas 12 external datasets were held out entirely for zero-shot and fine-tuned transfer.
By standardizing segmentation, question--answer, and radiology report annotations across datasets into a unified dialogue format with 3D masks without manual relabelling, HoloDATA provides 6.2 million instruction dialogues and 2.8 million CT--mask pairs from non-contrast and contrast-enhanced scans, covering 246 anatomical structures and 50 lesion categories across six major anatomical systems (\figref[b]{fig:main}).

Architecturally, HoloCT encodes each CT scan through two parallel branches (\figref[c]{fig:main}): a global encoder for whole-volume context and a detailed encoder that preserves fine-grained anatomical detail at standardized isotropic spacing.
Features from both encoders are projected into whole-volume global tokens and fine-grained spatial tokens and supplied directly to a unified multimodal large language model instantiated as HoloCT-4B and HoloCT-8B, integrating whole-volume and fine-grained features within a shared vision--language representation.
When spatial localization is requested, segmentation tokens emitted by the language model interact with multiscale skip features from the detailed encoder in a mask decoder, pairing diagnostic text with 3D masks.

\subsection{Evaluation settings}\label{sec:results-evaluation-settings}

We evaluated HoloCT on 346 tasks spanning four task families and 11 formulations (\figref[d]{fig:main}): referring segmentation (target-level and finding-level), lesion screening (single-target and multi-target), VQA (closed-ended, open-ended and grounded) and report generation (narrative, structured, anatomy-grounded and finding-grounded).
Evaluations were organized into three tiers without scan overlap: held-out test splits of 61 internal public datasets, 12 external public datasets excluded from training, covering seen and unseen categories as well as VQA and report generation, and six clinical cohorts for lesion segmentation (centres~1--3), histopathology-confirmed lesion screening (centres~3 and~4) and report generation (centre~5).
More details can be found in \supprefrange{1}{5}.

A single pretrained HoloCT model was evaluated without task-specific fine-tuning on all internal benchmarks and in zero-shot external and clinical evaluations.
For supervised comparisons in external lesion screening and segmentation, we additionally fine-tuned all unified models per task on the pooled training splits.

HoloCT was compared with three families of state-of-the-art models: specialist networks trained per benchmark, including nnU-Net~\citep{Isensee2021nnUNet} for segmentation and 3D ResNet-18~\citep{He2016ResNet} for screening; segmentation foundation models VISTA3D~\citep{He2024VISTA3DAU}, SAT-Pro~\citep{Zhao2023LargevocabularySF} and VoxTell~\citep{Rokuss2025VoxTellFP}; and vision--language models Qwen3-VL-32B~\citep{Qwen3VL2025}, MedGemma~1.5-4B~\citep{Sellergren2026MedGemma15}, RadFM-14B~\citep{Wu2025TowardsGF}, HuluMed-32B~\citep{Jiang2025HuluMed} and M3D-Phi3-4B (M3D)~\citep{Bai2024M3DA3}. Details of these methods can be found in Supplementary Methods.
As few existing models support finding-grounded report generation, we additionally constructed a two-stage pipeline using HuluMed-32B and VoxTell for comparison.
For fair comparison, pretrained baselines were evaluated using released weights, whereas specialist and fine-tuned models used the corresponding training splits. In clinical screening, HoloCT was also compared with four radiologists in a blinded reader study.

Segmentation was assessed using Dice and HD95~\citep{Taha2015SegmentationMetrics}, with hit rates at Dice thresholds of 0.05 and 0.10 ($\mathrm{HIT}_{5\%}$ and $\mathrm{HIT}_{10\%}$) for finding-level referring segmentation.
Screening was assessed using F1, sensitivity, specificity and accuracy, together with conditional Dice on true-positive detections; closed-ended VQA was assessed by accuracy, and grounded outputs additionally by match rate and Dice.
Free-text outputs were scored with lexical metrics (BLEU-4~\citep{Papineni2002BLEU}, ROUGE-L~\citep{Lin2004ROUGE} and METEOR~\citep{Banerjee2005METEOR}) and clinical metrics (LlamaScore, evaluated using Llama-3.1-70B-Instruct~\citep{Dubey2024Llama3}, plus RaTEScore~\citep{Zhao2024RaTEScore} and GREEN~\citep{Ostmeier2024GREEN} for reports).
Building on prior work emphasizing both linguistic and clinical measures of report quality~\citep{CandidateRadCliQ2023,li2025towards}, we summarized free-text quality using the Language--Clinical Score (LCS), the equal-weight mean of linguistic similarity and clinical agreement (Methods).

\subsection{Open-vocabulary 3D segmentation}\label{sec:results-segmentation}

\begin{figure}[p]
    \NaturalPageCenteredGraphic{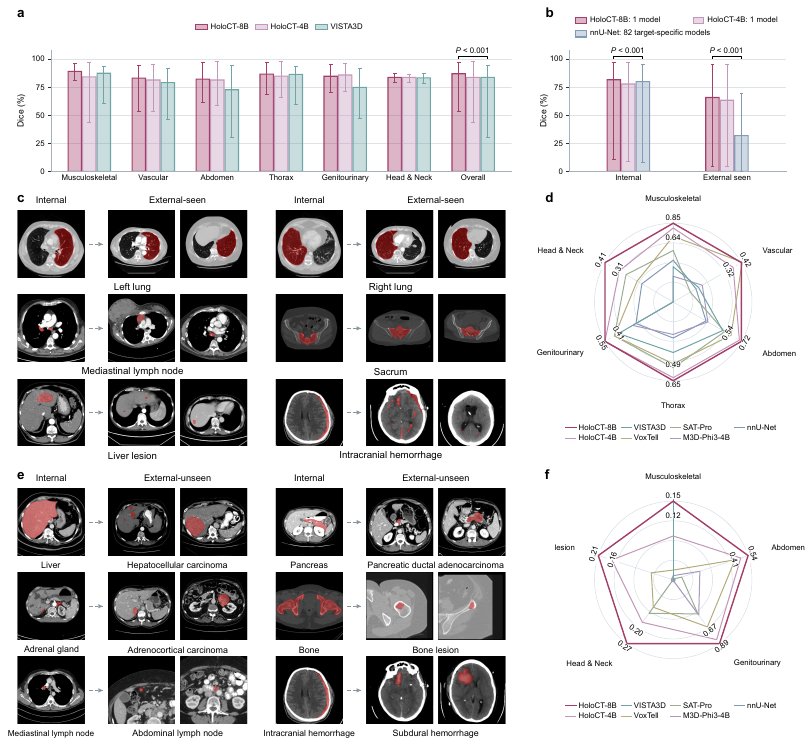}
\caption{\textbf{Language-guided segmentation across internal and external CT datasets.}
\textbf{a}, Internal comparison with VISTA3D across 128 shared targets.
\textbf{b}, Comparison with specialist nnU-Net models across 272 internal and 37 seen external targets.
\textbf{c,e}, Internal reference examples alongside external-seen and external-unseen targets, respectively.
\textbf{d,f}, External zero-shot Dice by anatomical and lesion group for 37 seen and 14 unseen targets, respectively.
In \textbf{a,b}, bars show means across targets; whiskers span target minima--maxima.
Brackets indicate unadjusted $P<0.001$ from two-sided paired $t$-tests across targets.
Radar axes in \textbf{d,f} are normalized to HoloCT-8B, with native Dice values on labelled rings.}
\label{fig:seg}
\end{figure}

We first evaluated HoloCT's foundational spatial perception through language-guided target-level referring segmentation across 272 internal and 51 external anatomical and lesion categories (\figref{fig:seg}, \suppref{8}).
On the 128 internal categories supported by the segmentation foundation model VISTA3D, HoloCT-8B increased mean Dice from 0.837 to 0.870 ($P<0.001$; \figref[a]{fig:seg}).
Across all 272 internal categories, a single HoloCT-8B model achieved a mean Dice of 0.816, outperforming the collection of 82 dedicated specialist nnU-Net models trained on individual datasets (0.799, $P<0.001$; \figref[b]{fig:seg}).

To assess zero-shot generalization beyond the training distribution, we next evaluated HoloCT without fine-tuning across 51 external categories comprising 37 seen and 14 unseen structures and lesions, where it consistently outperformed VISTA3D, SAT-Pro, M3D, and VoxTell (\figref[c--f]{fig:seg}).
Specifically, on the 37 seen categories, HoloCT-8B reached a mean Dice of 0.658, surpassing specialist nnU-Net models by 0.339 (0.319, $P<0.001$; \figref[b]{fig:seg}) and the strongest text-prompted baseline, VoxTell, by 0.108 ($P<0.001$).
It led five of the six anatomical and lesion groups, with VoxTell slightly higher on vascular structures (0.436 compared with 0.423; \figref[c,d]{fig:seg}).
On the 14 unseen categories, HoloCT-8B increased mean Dice over VoxTell from 0.340 to 0.414 ($P<0.001$) and led four of the five groups, with VISTA3D slightly higher on musculoskeletal bone lesions (0.165 compared with 0.154; \figref[e,f]{fig:seg}).
In summary, a single HoloCT surpassed both specialist and foundation models on in-distribution benchmarks and demonstrated strong generalization performance to unseen anatomy and lesions.

\subsection{Single- and multi-target grounded clinical screening}\label{sec:results-screening}

Beyond delineating known targets, clinical screening requires jointly determining whether an abnormality is present and localizing detected lesions on scans (\figref[a]{fig:grounded_screening}).
We evaluated single-target screening against ResNet-18, nnU-Net, and M3D across internal testing, external zero-shot transfer, and external fine-tuning (\figref[b,c]{fig:grounded_screening}), and multi-target screening against ResNet-18 and nnU-Net across internal testing and external fine-tuning (\figref[d,e]{fig:grounded_screening}).
In single-target screening, HoloCT-8B achieved the highest macro F1 in all three settings: it exceeded nnU-Net by 19.4 percentage points internally (0.651 and 0.458, respectively) and, when fine-tuned as a single pooled model across external screening datasets, surpassed dedicated specialist nnU-Net models by 9.2 percentage points (0.859 compared with 0.767, $P<0.001$; \figref[b,c]{fig:grounded_screening}).
Across nine external lesion types evaluated without fine-tuning (three seen and six unseen), HoloCT-8B reached a macro F1 of 0.397, substantially outperforming text-prompted M3D (0.037) as well as ResNet-18 (0.050) and nnU-Net (0.036), which are restricted to predefined label sets (\figref[c]{fig:grounded_screening}).

To reflect clinical workflows in which multiple abnormalities are assessed simultaneously, we next evaluated multi-target screening by querying and segmenting multiple lesion types in a single pass without per-lesion prompting (\figref[d]{fig:grounded_screening}).
HoloCT-8B raised macro F1 over nnU-Net from 0.368 to 0.718 internally and from 0.647 to 0.776 after external fine-tuning ($P<0.001$; \figref[e]{fig:grounded_screening}).
On detected positive scans, HoloCT's conditional Dice exceeded that of nnU-Net internally (0.679 compared with 0.337) and was lower after external fine-tuning (0.360 compared with 0.519), consistent with HoloCT returning positive masks on a larger and more difficult set of detected cases.
Across the 13 lesion types shared between single- and multi-target benchmarks, querying multiple targets jointly improved overall macro F1 over separate single-target queries for both HoloCT-8B (0.740 versus 0.727) and HoloCT-4B (0.729 versus 0.716; both $P<0.001$), although single-target queries remained higher on the externally fine-tuned subset (\suppreftwo{9}{10}).
By contrast, ResNet-18 and nnU-Net performed worse under multi-target prediction than with separate single-target models (\figref[f]{fig:grounded_screening}).

To test whether generating 3D masks aided lesion-presence classification, we removed the grounding request from the prompt and retained only the question (\figref[g]{fig:grounded_screening}).
Across 18 paired internal and external zero-shot lesion types, omitting mask generation reduced mean macro F1 from 0.562 to 0.505 for HoloCT-8B and from 0.531 to 0.504 for HoloCT-4B (both $P<0.001$; \figref[h]{fig:grounded_screening}).
Meanwhile, conditional Dice under grounded screening remained on par with standalone segmentation (\edfigref[a]{fig:extended_grounding}).

\subsection{Visual question answering and anatomy-grounded report generation}\label{sec:results-vqa-reporting}

\begin{figure}[p]
    \NaturalPageCenteredGraphic{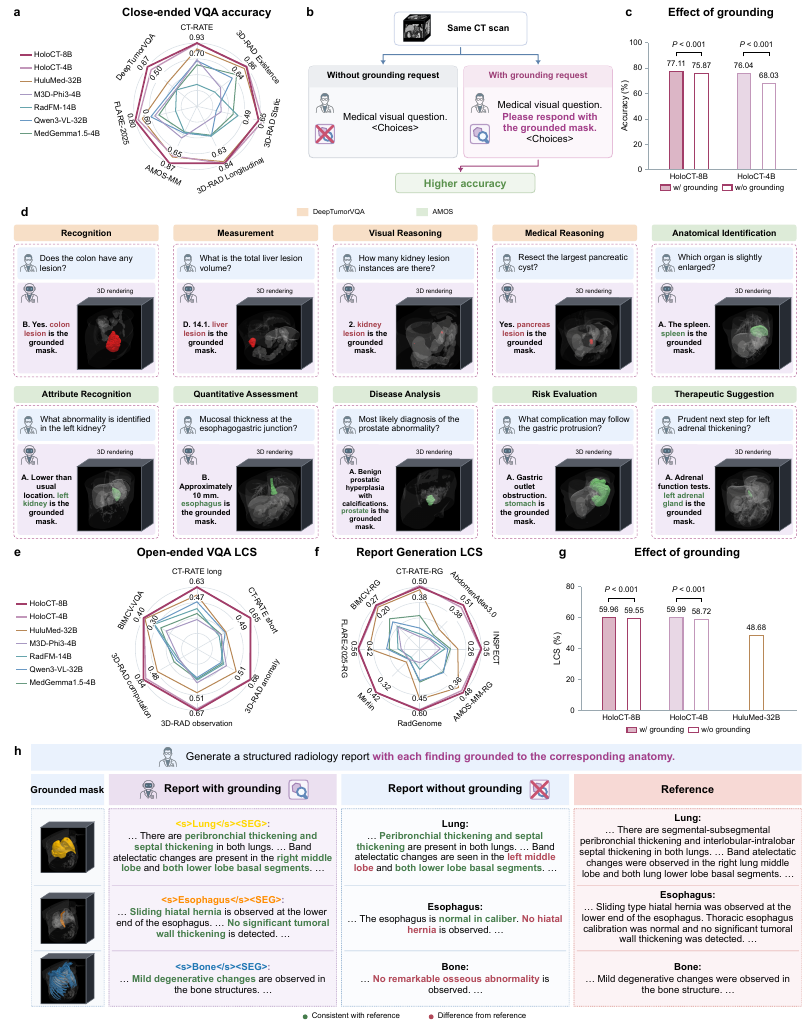}
\caption{\textbf{Visual question answering (VQA) and anatomy-grounded report generation.}
\textbf{a}, Closed-ended VQA accuracy.
\textbf{b}, VQA prompts with and without mask requests.
\textbf{c}, Mean overall accuracy across DeepTumorVQA and AMOS-MM-VQA.
\textbf{d}, Question--answer examples with supporting masks.
\textbf{e},\textbf{f}, Open-ended VQA and report-generation Language--Clinical Score (LCS), respectively, including external BIMCV.
\textbf{g}, RadGenome report LCS (\%) with and without grounding versus HuluMed-32B.
\textbf{h}, Example of anatomy-grounded report generation; green and red text indicate agreement and discrepancy with the reference.
In \textbf{a},\textbf{e},\textbf{f}, radar axes are normalized to HoloCT-8B, with native scores on labelled rings.
In \textbf{c},\textbf{g}, filled and open bars show means with and without grounding; brackets indicate unadjusted $P<0.001$ from two-sided scan-level paired permutation tests.}
\label{fig:vqa_rg}
\end{figure}

In addition to clinical screening, we evaluated HoloCT's broader clinical reasoning on closed-ended, open-ended, and grounded VQA, together with narrative, structured, and anatomy-grounded report generation (\figref{fig:vqa_rg}).
Across internal closed- and open-ended VQA benchmarks and external BIMCV-VQA~\citep{BIMCVCOVID19}, HoloCT outperformed competing general-purpose and medical VLMs (\figref[a,e]{fig:vqa_rg}).
In closed-ended VQA for CT-RATE, HoloCT-8B achieved an accuracy of 93.4\% (95\% confidence interval (CI), 92.9\%--94.0\%) compared with 84.7\% (95\% CI, 83.9\%--85.5\%) for the strongest baseline, HuluMed-32B (\suppref{11}).
On external BIMCV-VQA, HoloCT-8B reached an LCS of 0.401, compared with 0.390 for HuluMed-32B (\suppref{12}).
In grounded VQA across AMOS-MM-VQA~\citep{AMOSMM} and DeepTumorVQA~\citep{DeepTumorVQA}, generating supporting 3D masks alongside answers significantly improved mean accuracy from 75.9\% to 77.1\% for HoloCT-8B and from 68.0\% to 76.0\% for HoloCT-4B (both $P<0.001$; \figref[b--d]{fig:vqa_rg}); both models localized 94.0\% of reference target objects at Dice levels comparable to standalone segmentation (\edfigref[a]{fig:extended_grounding}, \suppreftwo{15}{16}).

Extending from targeted question answering to comprehensive scan interpretation, HoloCT also led all seven internal narrative and structured report-generation benchmarks as well as external BIMCV-RG (\figref[f]{fig:vqa_rg}).
On RadGenome and external BIMCV-RG, HoloCT-8B achieved LCS values of 0.595 and 0.268, exceeding HuluMed-32B (0.487 and 0.256, respectively).
In anatomy-grounded report generation on RadGenome, both HoloCT-4B and HoloCT-8B matched all reference thoracic regions (100\%) and raised report LCS to 0.600, outperforming text-only generation (0.587 and 0.595, respectively; both $P<0.001$) and HuluMed-32B (0.487; \figref[g]{fig:vqa_rg}, \suppreftwo{17}{18}).
In qualitative comparisons, pairing regional descriptions with 3D anatomical masks corrected a left-sided mislocalization of middle-lobe atelectasis and recovered a sliding hiatal hernia and mild bone degeneration missed by text-only report generation (\figref[h]{fig:vqa_rg}).

\subsection{Referring segmentation and finding-grounded report generation}\label{sec:results-finding-grounding}

Whereas anatomy-grounded reporting links text to predefined organs, clinical radiology reports describe diverse findings in free text, ranging from focal lesions such as pulmonary nodules or masses to diffuse, non-focal abnormalities such as bilateral ground-glass opacities (\figref[a,b]{fig:grounded_report_generation}).
To evaluate fine-grained localization from clinical descriptions, we first benchmarked referring segmentation on ReXGroundingCT~\citep{Baharoon2025ChestCTGrounding}, where HoloCT-8B outperformed VoxTell and SAT-Pro across focal Dice, non-focal Dice, and spatial hit rates: overall mean Dice reached 0.291 (95\% CI, 0.277--0.304), compared with 0.259 (95\% CI, 0.240--0.274) for VoxTell and 0.168 (95\% CI, 0.152--0.185) for SAT-Pro, and non-focal Dice reached 0.252 (95\% CI, 0.224--0.277), compared with 0.149 (95\% CI, 0.129--0.183) for VoxTell and 0.098 (95\% CI, 0.081--0.124) for SAT-Pro (\figref[c]{fig:grounded_report_generation}, \suppref{19}).
HoloCT thus localized both focal lesions and diffuse, non-focal chest CT findings more accurately than standalone text-prompted segmentation models from free-text clinical descriptions.

To unify free-text report generation with voxel-level finding localization, we next evaluated finding-grounded report generation, in which HoloCT autonomously identified abnormal findings across the chest CT scan and generated a dedicated 3D mask for each reported finding without pre-specified target prompts (\figref[b]{fig:grounded_report_generation}).
Compared with a two-stage pipeline combining HuluMed-32B for report generation with VoxTell for post-hoc finding segmentation, HoloCT-8B raised finding-detection F1 from 0.239 (95\% CI, 0.213--0.278) to 0.278 (95\% CI, 0.248--0.314) and mean conditional Dice from 0.186 (95\% CI, 0.165--0.213) to 0.234 (95\% CI, 0.209--0.258; \figref[d]{fig:grounded_report_generation}, \suppref{20}).
Removing the grounding request lowered all six linguistic and clinical report scores for HoloCT-8B (\figref[e]{fig:grounded_report_generation}); in qualitative review, grounded generation detected and localized bilateral consolidation missed by text-only report generation (\figref[f]{fig:grounded_report_generation}).

\subsection{Multicentre clinical validation and reader study}\label{sec:results-clinical}

\begin{figure}[p]
    \NaturalPageCenteredGraphic{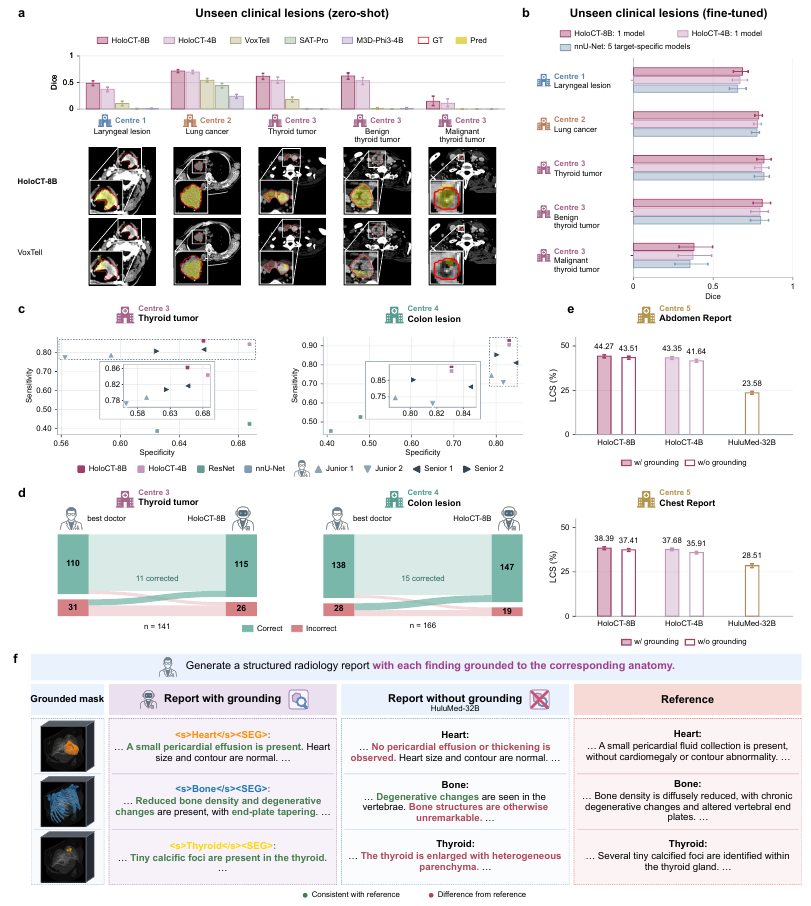}
\caption{\textbf{Multicentre clinical evaluation of segmentation, screening, and report generation.}
\textbf{a},\textbf{b}, Zero-shot and fine-tuned segmentation across five clinical tumour categories at centres 1--3; bars show mean Dice with scan-level bootstrap 95\% confidence intervals, and red outlines and yellow overlays indicate reference and predicted masks.
\textbf{c}, Sensitivity and specificity of fine-tuned models (squares) and two junior and two senior radiologists (triangles) on held-out thyroid ($n=141$) and colon ($n=166$) scans.
\textbf{d}, Case-level classification comparison between HoloCT-8B and the most accurate senior radiologist on the same cohorts; green and red indicate correct and incorrect diagnoses, respectively.
\textbf{e}, Zero-shot report Language--Clinical Score (LCS, \%) across 5,307 abdominal and 2,974 chest CT scans. Filled and open bars denote inference with and without grounding; whiskers show scan-level bootstrap 95\% confidence intervals.
\textbf{f}, Report excerpts and anatomical masks compared with HuluMed-32B and reference reports.}
\label{fig:real_world_clinical_evaluation}
\end{figure}

Routine hospital CT workflows span diverse anatomical sites, scanner protocols, and contrast phases, requiring models that generalize across centres without per-cohort re-engineering.
To evaluate clinical applicability across real-world diagnostic settings, we tested HoloCT on 11,455 CT scans from six retrospective cohorts at five hospitals, covering lesion delineation across five targets—laryngeal lesions on contrast-enhanced CT (centre~1), lung cancer on non-contrast CT (centre~2), and benign, malignant and combined thyroid tumours on non-contrast CT (centre~3)—as well as clinical screening of histopathology-confirmed thyroid tumours (centre~3) and colon lesions (centre~4), and high-volume non-contrast chest and abdominal report generation (centre~5; \figref{fig:real_world_clinical_evaluation}).
In zero-shot segmentation across the five clinical lesion categories at centres~1--3, HoloCT-4B and HoloCT-8B substantially outperformed M3D, SAT-Pro, and VoxTell; HoloCT-8B increased mean Dice over VoxTell from 0.107 (95\% CI, 0.066--0.150) to 0.488 (95\% CI, 0.441--0.531) on laryngeal lesions and from 0.543 (95\% CI, 0.506--0.579) to 0.715 (95\% CI, 0.687--0.741) on lung cancer (\figref[a]{fig:real_world_clinical_evaluation}, \suppref{8}).
Fine-tuned as a single model on the pooled training splits of all five clinical lesion categories, HoloCT-8B achieved a mean Dice of 0.696 compared with 0.681 across five dedicated specialist nnU-Net models, leading four of the five categories (\figref[b]{fig:real_world_clinical_evaluation}).
Specifically, HoloCT-8B achieved a Dice of 0.684 (95\% CI, 0.627--0.720) compared with 0.655 (95\% CI, 0.602--0.707) on laryngeal lesions and 0.786 (95\% CI, 0.762--0.809) compared with 0.776 (95\% CI, 0.742--0.790) on lung cancer, while matching nnU-Net on combined thyroid tumours at 0.819 (95\% CI, 0.774--0.864) compared with 0.819 (95\% CI, 0.759--0.853). Averaged over the five clinical lesion categories and both evaluation settings (zero-shot versus VoxTell; fine-tuned versus nnU-Net), HoloCT-8B improved Dice by 18.2 points (Methods).

We next evaluated diagnostic performance in two clinical scenarios prone to missed findings in routine CT interpretation: thyroid tumours on non-contrast CT (centre~3), where subtle low-attenuation lesions within iodine-rich thyroid parenchyma are difficult to discern without ultrasound or intravenous contrast, and colon lesions on routine contrast-enhanced CT (centre~4), where focal wall thickening or intraluminal masses can mimic peristalsis or luminal contents.
A single HoloCT model was fine-tuned on the merged training splits of both histopathology-confirmed cohorts, pairing classification with 3D tumour segmentation for thyroid scans and text-only classification for colon scans, and evaluated on held-out cases in the matching task format ($n=141$ thyroid and $n=166$ colon scans; \figref[c,d]{fig:real_world_clinical_evaluation}, \edfigref[b]{fig:extended_grounding}).
In blinded reader studies against two junior and two senior radiologists per cohort, HoloCT-8B surpassed every reader in sensitivity and overall accuracy: on thyroid tumours, it achieved 81.6\% accuracy (radiologists, 72.3\%--78.0\%), with 86.2\% sensitivity (95\% CI, 79.1\%--92.4\%), 65.6\% specificity (95\% CI, 50.0\%--82.4\%), and an F1 score of 0.879 (95\% CI, 0.829--0.917; \suppref{9}); on colon lesions, it achieved 88.6\% accuracy (radiologists, 75.3\%--83.1\%), with 92.6\% sensitivity (95\% CI, 86.9\%--97.1\%), 83.1\% specificity (95\% CI, 73.7\%--90.7\%), and an F1 score of 0.903 (95\% CI, 0.853--0.939).
Compared case by case with the most accurate senior radiologist in each cohort, HoloCT-8B correctly identified 11 thyroid scans and 15 colon scans missed by the reader while misclassifying six scans per cohort that the reader judged correctly, increasing total correct diagnoses from 110 to 115 of 141 thyroid cases and from 138 to 147 of 166 colon cases (\figref[d]{fig:real_world_clinical_evaluation}).

Finally, to assess automated report drafting across high-volume routine workloads without site-specific adaptation, we evaluated zero-shot anatomy-grounded report generation across 5,307 non-contrast abdominal and 2,974 non-contrast chest CT examinations from centre~5.
HoloCT-8B achieved LCS values of 0.443 (95\% CI, 0.434--0.451) on abdominal CT and 0.384 (95\% CI, 0.376--0.392) on chest CT, substantially outperforming HuluMed-32B at 0.236 (95\% CI, 0.228--0.243) and 0.285 (95\% CI, 0.275--0.296), respectively (\figref[e]{fig:real_world_clinical_evaluation}).
In qualitative review, HoloCT captured subtle multi-organ abnormalities---including small pericardial effusion, reduced bone density with end-plate changes, and tiny thyroid calcifications---that HuluMed-32B omitted or misdescribed, while linking each regional statement to a verifiable 3D anatomical mask (\figref[f]{fig:real_world_clinical_evaluation}).

\subsection{Bidirectional synergy between clinical understanding and spatial perception}\label{sec:results-ablations}

Having demonstrated strong performance across public benchmarks and clinical cohorts, we next investigated the underlying mechanism of semantic--spatial synergy during training by comparing six configurations varying task supervision and the integration of whole-volume and fine-grained spatial features (\figref[a--c]{fig:training_ablation}, \edfigref[c]{fig:extended_architecture}).
With only fine-grained tokens supplied to the language model, adding language supervision to segmentation-only training improved Dice from 0.601 to 0.613, whereas adding segmentation supervision to language-only training increased screening macro F1 from 0.256 to 0.351, alongside gains in VQA and report generation (\figref[c]{fig:training_ablation}).
Adding the global encoder further improved all four task families, while directly supplying both whole-volume and fine-grained tokens to the language model outperformed separate pathways and early token fusion (\figref[b,c]{fig:training_ablation}).

To further determine at inference whether clinical interpretation relied directly on fine-grained 3D spatial features, we masked language-model attention to fine-grained tokens in HoloCT-8B while leaving skip connections from the detailed encoder to the mask decoder intact (\figref[d]{fig:training_ablation}).
Blocking this attention significantly degraded performance across all four tasks ($P<0.001$; \figref[e]{fig:training_ablation}): segmentation Dice fell from 0.840 to 0.811, screening macro F1 dropped from 0.678 to 0.061, VQA accuracy declined from 77.1\% to 70.9\%, and report LCS decreased from 0.507 to 0.429.
Crossing spatial-token masking with retention or removal of the grounding request further showed a significant interaction ($P<0.001$; \figref[f,g]{fig:training_ablation}): masking fine-grained tokens reduced the gain from mask generation from 1.47 to 0.53 percentage points in screening F1, from 1.24 to 1.06 percentage points in VQA accuracy, and from 1.19 to 0.24 points in report LCS (0–100 scale). These results indicate that grounded interpretation depends on the fine-grained spatial tokens shared with segmentation, although masking constitutes an out-of-distribution intervention.

\section{Discussion}\label{sec:discussion}

HoloCT integrates whole-volume and fine-grained 3D features within a unified multimodal large language model, uniting CT interpretation and voxel-level segmentation across tasks largely addressed separately by segmentation models~\citep{Isensee2021nnUNet,He2024VISTA3DAU,Zhao2023LargevocabularySF,Rokuss2025VoxTellFP} and medical VLMs~\citep{blankemeier2026merlin,Hamamci2024GeneralistFM,Wu2025TowardsGF}, or connected only at the decoder output level~\citep{Bai2024M3DA3,luo2025vividmed,wu2026unibiomed}.
Built upon the large-scale HoloDATA corpus, HoloCT outperformed specialist and foundation segmentation models, task-specific classifiers, and VLMs across 346 benchmark tasks and six clinical cohorts at five hospitals.

A central finding of this work is the bidirectional synergy between clinical understanding and 3D spatial perception in volumetric CT.
Prior studies on 2D radiographs aligned report text with global and local image features~\citep{CandidateGLoRIA2021,CandidateCheXzero2022,CandidateKAD2023}, and recent medical visual tokenizers encode anatomy and semantics within unified representations~\citep{CandidateMedITok2025,CandidateMedVTok2026}.
Extending this joint representation design to volumetric CT, HoloCT supplies whole-volume global tokens and fine-grained spatial tokens directly to a unified language--segmentation model.
During training, adding text-only dialogues improved 3D segmentation accuracy, whereas incorporating mask-paired dialogues improved lesion screening, VQA, and radiology report generation.
At inference, this synergy operated in both directions: semantic representations enabled language-guided segmentation to generalize zero-shot to unseen anatomy, novel lesion categories, and diffuse free-text findings, while retaining the grounding request to generate 3D masks alongside text significantly improved screening, VQA, and report generation without compromising mask quality.
Crucially, masking language-model attention to fine-grained spatial tokens markedly attenuated the gains from mask generation, indicating that grounded clinical reasoning draws directly on the fine-grained spatial features shared with voxel-level segmentation.
Moreover, although HoloCT-8B achieved stronger overall performance, explicit grounding requests in VQA and report generation yielded larger gains for HoloCT-4B, suggesting that larger models may partially internalize spatial--semantic alignment, whereas smaller models benefit more from explicit spatial grounding during clinical interpretation.

Beyond single-task performance, unifying semantic reasoning and segmentation consolidates multi-step clinical workflows within a single model~\citep{CandidateGMAI2023}.
Whereas interactive and text-prompted segmentation models require users to specify each target in advance~\citep{ma2024segment}, HoloCT screened multiple candidate lesions in a single pass, improving overall detection accuracy over separate single-target queries while producing verifiable 3D masks.
In addition, in finding-grounded report generation, HoloCT detected and localized chest CT abnormalities simultaneously, outperforming a two-stage pipeline combining HuluMed-32B and VoxTell.

In clinical practice, pairing diagnostic text with voxel-level masks directly addresses the verifiability bottleneck of medical VLMs.
Although recent VLMs achieve strong aggregate scores in report generation~\citep{CandidateLLaVARad2025}, fine-grained perceptual errors and anatomical mislocalizations often remain hidden in free text~\citep{CandidateMMBU2026} and are labour-intensive to audit across CT volumes spanning hundreds of slices.
Whereas concept-based explanations support auditing only at the whole-image level~\citep{CandidateMONET2024,CandidateConceptCLIP2026}, grounding classifications, question answers, and report statements with 3D masks allows clinicians to inspect the volumetric evidence underlying each textual output.
Alongside spatial verifiability, multicentre clinical validation demonstrated strong real-world applicability: HoloCT-8B surpassed human radiologists in sensitivity and overall accuracy on histopathology-confirmed thyroid tumours and colon lesions and generalized zero-shot to high-volume chest and abdominal report generation, highlighting its potential for subtle-lesion triage and automated report drafting.

Several dataset and modelling limitations warrant future investigation.
First, although HoloDATA spans six major anatomical systems for segmentation, grounded report generation remains concentrated in thoracic and abdominal CT, with finding-level 3D masks currently limited to chest CT; extending fine-grained finding annotations to abdominal, head-and-neck, neuroimaging, and musculoskeletal CT will broaden grounded reporting.
Second, despite substantial gains over existing foundation models, segmentation accuracy on musculoskeletal bone lesions and diffuse, non-focal chest findings remains limited; adaptive high-resolution cropping and iterative mask refinement may further improve delineation of these challenging targets.
Third, HoloCT currently interprets single CT volumes independently, whereas routine radiological workflows rely on longitudinal comparisons with prior scans, multi-phase contrast studies, and integration with electronic health records or complementary modalities such as MRI and PET--CT; extending the unified architecture to these longitudinal and multimodal settings is an important next step.

From a clinical translation perspective, our hospital evaluation was retrospective and conducted within a single country, motivating prospective validation across international cohorts and diverse scanner protocols.
Further reader-in-the-loop trials are needed to quantify how interactive 3D mask verification affects diagnostic accuracy, reading efficiency, and clinician trust in routine workflows~\citep{Tanno2024CollaborationBC,CandidateRadiologistAI2024}.
Finally, safe clinical deployment will require calibrated uncertainty estimation and automated quality control alongside expert radiological oversight.

In conclusion, HoloCT demonstrates that clinical understanding and 3D spatial perception can be learned synergistically within a single generalist foundation model, offering a unified path towards more accurate, generalizable, and spatially verifiable CT interpretation.

\section{Methods}\label{sec:methods}\hypertarget{main-methods}{}

\subsection{Data curation}\label{sec:holo-data-construction}

HoloDATA was assembled from 73 public datasets, including 62 segmentation datasets and 11 with language annotations, alongside six clinical cohorts from five hospitals.
Dataset sources, annotations and data splits are listed in \supprefrange{1}{5}.
Predefined splits were retained; for public datasets without predefined splits, 20\% of scans were randomly held out for testing.
For entries marked as aligned, scans shared across datasets and their associated annotations were assigned to the same split. We verified that none of the 12 external datasets, the clinical cohorts or the internal test scans were included in the training data of VISTA3D or Med3DVLM, according to their published data descriptions. The 6.2 million instruction dialogues were counted as single-turn dialogues, with each turn of a multi-turn dialogue counted separately.
The 2.8 million 3D segmentation masks were counted as unique CT--mask pairs, excluding repeated uses of the same mask across tasks.

\phantomsection\label{sec:task-construction}
We extended the SAT label vocabulary~\citep{Zhao2023LargevocabularySF} into a shared target vocabulary of 246 anatomical structures and 50 lesion categories (\suppref{7}).
For target-level referring segmentation, prompts specified target names; for finding-level referring segmentation, prompts contained free-text findings from ReXGroundingCT~\citep{Baharoon2025ChestCTGrounding}.
Single-target and multi-target lesion screening required presence-or-absence decisions and masks for positive targets.
Missing annotations were not treated as negative labels.
Closed-ended VQA required binary or multiple-choice answers, whereas open-ended VQA required free-text answers.
Grounded VQA paired closed-ended answers with supporting segmentation masks.
Reference reports comprised findings or impressions for narrative report generation and region-specific findings for structured report generation.
Each dataset's original reporting format was preserved.
Anatomy-grounded report generation paired regional descriptions with anatomical masks, whereas finding-grounded report generation paired individual findings with their corresponding masks.

GPT-4o was used to generate report-question templates and to match questions from DeepTumorVQA~\citep{DeepTumorVQA} and AMOS-MM-VQA~\citep{AMOSMM} with supporting masks.
It was also used to clean translated BIMCV reports~\citep{BIMCVCOVID19} and to generate open-ended questions from them.
BIMCV image--text pairs were manually reviewed before external evaluation.
Dataset-specific processing is described in Supplementary Methods, and prompts are provided in \supprefrange{24}{29}.

\subsection{Model architecture}\label{sec:design-training}

\phantomsection\label{sec:ct-preprocessing}
Each CT scan was converted into a global input and a detailed input.
Scans were first reoriented to superior--right--anterior axes using MONAI~\citep{Cardoso2022MONAI}.
The global input was resized to $128\times256\times256$ voxels by linear interpolation.
The detailed input was cropped to the bounding box of voxels above $-1000$ Hounsfield units (HU), linearly resampled to $1.5\times1.5\times1.5$~mm and symmetrically padded with $-1000$~HU so that each dimension was divisible by 32.
Intensities in both inputs were clipped to $[-1000,1000]$~HU and scaled to $[0,1]$.

\phantomsection\label{sec:holo-ct-architecture}
HoloCT combined a Med3DVLM-derived global encoder~\citep{Xin2025Med3DVLM}, a VISTA3D-derived detailed encoder~\citep{He2024VISTA3DAU} and a Qwen3-VL language backbone~\citep{Qwen3VL2025}.
HoloCT-4B and HoloCT-8B used Qwen3-VL-4B and Qwen3-VL-8B, respectively.
A global projector produced 256 whole-volume global tokens from the global input, and a detailed projector produced a fine-grained spatial token grid with one token per $32^3$ voxels of the detailed input.
Whole-volume tokens preceded fine-grained tokens in the language-model input, so that fine-grained tokens could attend to whole-volume tokens through causal self-attention.

Qwen3-VL's multimodal rotary position embedding has three positional components (\edfigref[f]{fig:extended_components}).
Each whole-volume or text token received the same sequential index for all three components.
The fine-grained token at row $i$ and column $j$ of depth plane $k$ received indices $(b_k,b_k+i,b_k+j)$, where $b_k$ is the starting index of plane $k$.
A text depth marker for each plane encoded the position of the plane centre on the training grid:
\begin{equation}
    r_k=\left(32k+\frac{31}{2}\right)\times1.5~\mathrm{mm},
    \label{eq:nominal-depth-marker}
\end{equation}
where $k$ is the zero-based plane index, 32 is the voxel stride and 1.5~mm is the resampled spacing.

Each target description was followed by a \textless{}SEG\textgreater{} token, whose final hidden state was projected to a 48-dimensional target representation $q_k$.
A mask decoder with multiscale skip connections from the detailed encoder produced shared 48-channel voxel features $F(u)$ at each voxel $u$.
Target-specific logits and probabilities were obtained from their inner product:
\begin{equation}
    \begin{aligned}
        z_k(u)&=q_k^{\mathsf T}F(u),\\
        p_k(u)&=\operatorname{sigmoid}\!\left(z_k(u)\right).
    \end{aligned}
    \label{eq:language-conditioned-mask}
\end{equation}
Mask channels followed the order of the generated \textless{}SEG\textgreater{} tokens.
Predictions were restored to the original scan geometry and binarized at 0.5.
Training augmentation, layer specifications and implementation details are provided in Supplementary Methods.

\subsection{Training and ablations}\label{sec:training-procedure}
Training comprised three stages: vision--language alignment (Stage~1), multitask pre-training (Stage~2) and high-quality multitask tuning (Stage~3) (\suppref{6}).
The encoders and language backbone were initialized from their pretrained weights.
Stage~1 trained only the global and detailed projectors on approximately 0.68 million report-generation samples.

Stage~2 used approximately 2.70 million language-supervised and 3.50 million segmentation-supervised samples, including single-region report-generation samples in which each region of a structured report formed a separate sample.
The language backbone was trained with low-rank adaptation (LoRA)~\citep{Hu2022LoRA}, and all other modules were fully trainable.
Stage~3 used approximately 1.69 million language-supervised and 1.15 million segmentation-supervised samples with the same trainable modules.
Stage~3 excluded single-region report-generation samples, so that all report-generation tasks used complete reports rather than individual regional descriptions.
Stage~3 excluded RadGenome VQA, because of its limited question diversity and disproportionate sample size, and PASTA-Gen-TotalSeg, because of its low-quality TotalSegmentator pseudo-labels.
In Stages~2 and~3, segmentation-supervised and language-supervised updates alternated.

Language supervision used next-token cross-entropy over assistant-response tokens:
\begin{equation}
    \mathcal{L}_{\mathrm{text}}=-\frac{1}{|\mathcal{A}|}
    \sum_{t\in\mathcal{A}}\log P_{\theta}\!\left(y_t\mid X,y_{<t}\right),
    \label{eq:holo-text-loss}
\end{equation}
where $\theta$ denotes the model parameters, $\mathcal{A}$ the set of supervised response positions, $y_t$ the token at position $t$ and $X$ the visual and conversational context.
For segmentation-supervised samples, \textless{}SEG\textgreater{} hidden states were obtained by teacher forcing with the reference response, and each predicted mask was supervised with binary cross-entropy (BCE) and Dice losses:
\begin{align}
    \mathcal{L}_{\mathrm{BCE},k}
    &=-\frac{1}{V_k}\sum_{u=1}^{V_k}
    \left[m_k(u)\log p_k(u)+(1-m_k(u))\log(1-p_k(u))\right],
    \label{eq:holo-bce-loss}\\
    \mathcal{L}_{\mathrm{Dice},k}
    &=1-\frac{2\sum_u p_k(u)m_k(u)+\epsilon_{\mathrm{n}}}
    {\sum_u p_k(u)^2+\sum_u m_k(u)^2+\epsilon_{\mathrm{d}}},
    \label{eq:holo-dice-loss}
\end{align}
where $p_k(u)$ is the predicted foreground probability of target $k$ at voxel $u$, $m_k(u)$ is the binary reference label and $V_k$ is the number of voxels.
Both smoothing constants, $\epsilon_{\mathrm{n}}$ and $\epsilon_{\mathrm{d}}$, were $10^{-5}$.
The segmentation loss was averaged over the $K$ supervised masks in each minibatch:
\begin{equation}
    \mathcal{L}_{\mathrm{seg}}=\frac{1}{K}\sum_{k=1}^{K}
    \left(0.3\mathcal{L}_{\mathrm{BCE},k}+0.7\mathcal{L}_{\mathrm{Dice},k}\right).
    \label{eq:holo-seg-loss}
\end{equation}
The total loss was $\mathcal{L}_{\mathrm{text}}+\mathcal{L}_{\mathrm{seg}}$ for segmentation-supervised samples and $\mathcal{L}_{\mathrm{text}}$ otherwise.

Both model sizes were trained with an effective batch size of 64 in bfloat16 precision.
AdamW~\citep{Loshchilov2019AdamW} used learning rates of $5\times10^{-4}$ in Stage~1 and $5\times10^{-5}$ in Stages~2 and~3 with cosine decay, $(\beta_1,\beta_2)=(0.9,0.999)$, $\epsilon=10^{-8}$, zero weight decay and gradient clipping at a norm of 1.0.
LoRA used rank $r=128$, scaling factor $\alpha=256$ and dropout 0.05.

Downstream fine-tuning started from the Stage~3 checkpoints, ran for five epochs and otherwise used Stage~3 settings.
For external lesion screening, each HoloCT model was fine-tuned once on the pooled training data for single-target and multi-target lesion screening.
For clinical segmentation, training data for all five targets from centres~1--3 were pooled.
For clinical screening, thyroid and colon training cases from centres~3 and~4 were pooled; thyroid cases received screening and segmentation supervision, and colon cases received screening supervision only.
Evaluation prompts requested the outputs of each cohort's training task.

All ablations used HoloCT-8B.
Training ablations S1--S6 were initialized from the same pretrained weights as Stage~1 and trained for five epochs with Stage~3 settings on a random 10\% of Stage~3 data plus the training splits of the ablation benchmarks.
S1--S3 supplied only fine-grained tokens to the language model and differed in supervision: S1 used language supervision, S2 used segmentation supervision with text and mask losses, and S3 combined both.
S4--S6 combined both forms of supervision and differed in how whole-volume and fine-grained tokens were integrated.
S4 supplied only whole-volume tokens to the language model and used fine-grained features solely for mask decoding.
S5 fused projected whole-volume and fine-grained tokens before the language model using two-way attention inspired by the Segment Anything Model (SAM)~\citep{Kirillov2023SAM}, supplying the updated whole-volume tokens to the language model and passing the updated fine-grained tokens to the mask decoder.
S6 supplied both token sets directly to the language model and corresponded to the HoloCT architecture.
Inference ablations evaluated a single HoloCT-8B checkpoint on the training-ablation benchmarks in a $2\times2$ design: fine-grained tokens were either visible or masked in causal attention, and grounding requests were either retained in or removed from the prompts.
Masking preserved the position indices of fine-grained tokens, and the mask decoder still received fine-grained features.
Initialization of the remaining modules, ablation architectures and benchmarks are detailed in Supplementary Methods, and results are shown in \figref{fig:training_ablation}.

\subsection{Compared methods}\label{sec:compared-methods}
Segmentation baselines were nnU-Net v2, VISTA3D, SAT-Pro, VoxTell and M3D-Phi3-4B (hereafter M3D)~\citep{Isensee2021nnUNet,He2024VISTA3DAU,Zhao2023LargevocabularySF,Rokuss2025VoxTellFP,Bai2024M3DA3}, and lesion-screening baselines additionally included 3D ResNet-18~\citep{He2016ResNet}.
Text-generation baselines were M3D, HuluMed-32B, RadFM-14B, MedGemma~1.5-4B and Qwen3-VL-32B~\citep{Jiang2025HuluMed,Wu2025TowardsGF,Sellergren2026MedGemma15,Qwen3VL2025}.
Released weights were used unless stated otherwise.

We trained nnU-Net from scratch on each dataset.
For single-target lesion screening, we trained a separate ResNet-18 for each dataset and lesion category; for multi-target lesion screening, ResNet-18 used a separate binary head for each target.
We fine-tuned M3D on pooled lesion-screening training data and HuluMed-32B only on ReXGroundingCT training data, using LoRA for the language backbone with all other modules fully trainable.
We fully fine-tuned SAT-Pro on ReXGroundingCT for finding-level referring segmentation.
Training configurations are detailed in Supplementary Methods.

The finding-grounded report generation baseline was a pipeline in which released VoxTell segmented each finding generated by fine-tuned HuluMed-32B.
VoxTell's training annotations overlap with the ReXGroundingCT test set, so its standalone and pipeline results are affected by label leakage.

\subsection{Benchmark evaluation}\label{sec:benchmark-evaluation}
Public benchmarks were used to assess performance on internal test splits, zero-shot performance on external datasets and performance after task-specific fine-tuning.
Test scans were excluded from training and checkpoint selection.

Target-level referring segmentation was evaluated using Dice and the 95th-percentile Hausdorff distance (HD95, in millimetres)~\citep{Taha2015SegmentationMetrics}:
\begin{equation}
    D(P,G)=\frac{2|P\cap G|}{|P|+|G|},
    \label{eq:evaluation-dice}
\end{equation}
\begin{equation}
    \mathrm{HD95}=
    \max\!\left\{Q^{A}_{0.95}(\mathcal{D}_{P,G}),Q^{A}_{0.95}(\mathcal{D}_{G,P})\right\},
    \label{eq:evaluation-hd95}
\end{equation}
where $P$ and $G$ are the predicted and reference foreground voxel sets, $\mathcal{D}_{P,G}$ and $\mathcal{D}_{G,P}$ contain directed nearest-surface distances in physical space, and $Q^{A}_{0.95}$ is the surface-area-weighted 95th percentile.

Lesion screening was evaluated from counts of true positives (TP), true negatives (TN), false positives (FP) and false negatives (FN):
\begin{equation}
    \begin{aligned}
        \mathrm{Accuracy}&=\frac{\mathrm{TP}+\mathrm{TN}}{\mathrm{TP}+\mathrm{TN}+\mathrm{FP}+\mathrm{FN}},\\
        \mathrm{F1}&=\frac{2\mathrm{TP}}{2\mathrm{TP}+\mathrm{FP}+\mathrm{FN}},\\
        \mathrm{Sensitivity}&=\frac{\mathrm{TP}}{\mathrm{TP}+\mathrm{FN}},\\
        \mathrm{Specificity}&=\frac{\mathrm{TN}}{\mathrm{TN}+\mathrm{FP}}.
    \end{aligned}
    \label{eq:evaluation-classification}
\end{equation}
Responses from HoloCT and generative baselines were classified as positive or negative by case-insensitive string matching, ignoring surrounding whitespace.
When prompts included grounding requests, a positive prediction additionally required a nonempty target mask.
An nnU-Net prediction was positive whenever its target mask contained foreground, irrespective of overlap with the reference.
Conditional Dice was the mean Dice score over true-positive CT--target pairs $\mathcal{T}$:
\begin{equation}
    D_{\mathrm{cond}}=\frac{1}{|\mathcal{T}|}\sum_{i\in\mathcal{T}}D(P_i,G_i).
    \label{eq:evaluation-conditional-dice}
\end{equation}

Closed-ended VQA answers (yes or no, or option labels A--D) were parsed using the same string-matching rule, and invalid or incorrect answers counted as errors.
DeepTumorVQA and AMOS-MM-VQA accuracies were also reported by question category (\suppref{11}).
For grounded VQA, predicted and reference objects were matched by category; for anatomy-grounded report generation, regions were matched by anatomical name (\suppreftwo{16}{18}).
For both tasks, match rate was the number of matched references divided by the total number of references, pooled across scans and categories or regions.
Dice and HD95 were computed for matched objects or regions only.

Open-ended VQA and report generation were evaluated using BLEU-4~\citep{Papineni2002BLEU}, ROUGE-L~\citep{Lin2004ROUGE}, METEOR~\citep{Banerjee2005METEOR} and LlamaScore, an integer clinical-agreement score from 0 to 10 assigned by Llama-3.1-70B-Instruct~\citep{Dubey2024Llama3} as a large language model (LLM) judge.
Report generation was additionally evaluated using RaTEScore~\citep{Zhao2024RaTEScore} and GREEN~\citep{Ostmeier2024GREEN}.
We defined the Language--Clinical Score (LCS) as the equally weighted mean of linguistic similarity $N$ and clinical agreement $C$, each on a 0--1 scale:
\begin{equation}
    \begin{aligned}
        N&=\frac{\mathrm{BLEU\mbox{-}4}+\mathrm{ROUGE\mbox{-}L}+\mathrm{METEOR}}{3},\\
        C_{\mathrm{VQA}}&=\mathrm{LlamaScore}/10,\\
        C_{\mathrm{RG}}&=\frac{\mathrm{RaTEScore}+\mathrm{LlamaScore}/10+\mathrm{GREEN}}{3},
    \end{aligned}
    \label{eq:clinical-mean}
\end{equation}
\begin{equation}
    \mathrm{LCS}_{\mathrm{VQA}}=\frac{N+C_{\mathrm{VQA}}}{2},\qquad
    \mathrm{LCS}_{\mathrm{RG}}=\frac{N+C_{\mathrm{RG}}}{2}.
    \label{eq:text-composite}
\end{equation}

\phantomsection\label{sec:finding-grounding-evaluation}
For finding-level referring segmentation, each ReXGroundingCT~\citep{Baharoon2025ChestCTGrounding} finding was supplied as a prompt, and the predicted mask was compared with its reference mask using Dice and hit rates.
For $M$ findings with Dice scores $D_j$, hit rates were calculated as
\begin{equation}
    \mathrm{HIT}_{\tau}=\frac{1}{M}\sum_{j=1}^{M}\mathbf{1}[D_j>\tau],\qquad \tau\in\{0.05,0.10\},
    \label{eq:evaluation-hit}
\end{equation}
and denoted $\mathrm{HIT}_{5\%}$ and $\mathrm{HIT}_{10\%}$, respectively.
For finding-grounded report generation, the LLM judge assessed whether each generated finding matched a reference finding from the same scan, requiring agreement in pathology, morphology and location without contradictory clinical attributes.
Each reference finding could be matched to at most one prediction, and only such matches counted as TP; when several predictions matched one reference, the prediction with the highest Dice was retained.
Finding detection was evaluated using F1, with
\begin{equation}
    \mathrm{FP}=n_{\mathrm{pred}}-\mathrm{TP},\qquad
    \mathrm{FN}=n_{\mathrm{ref}}-\mathrm{TP},
    \label{eq:evaluation-finding-counts}
\end{equation}
where $n_{\mathrm{pred}}$ and $n_{\mathrm{ref}}$ are the numbers of predicted and reference findings.
Conditional Dice and hit rates were computed over true-positive findings.
For both finding-level tasks, localization metrics were reported overall and separately for focal and non-focal findings.
Evaluation implementation is detailed in Supplementary Methods.

\subsection{Clinical evaluation}\label{sec:study-design}\label{sec:clinical-validation}

Clinical evaluation used six retrospective cohorts from five centres, with one CT scan per patient.
The research ethics committees of all five participating centres (Zhongshan Hospital, Fudan University; the First Affiliated Hospital, Zhejiang University School of Medicine; Sir Run Run Shaw Hospital, Zhejiang University School of Medicine; the Affiliated Yangming Hospital of Ningbo University; and Guizhou Provincial People's Hospital) approved the study and waived informed consent; images and records were de-identified.
Sample sizes were determined by the number of eligible cases, without an a priori power calculation.
Scans from centres~1 and~4 were contrast-enhanced, and the remaining scans were non-contrast.
Centres~1 and~2 included only lesion-positive cases, whereas centres~3 and~4 included both positive and negative cases.
Four physicians delineated lesion masks for centres~1--3 using clinical and pathological findings.
Histopathology provided reference diagnoses for thyroid (centre~3) and colon (centre~4) lesion screening; colon cases had no reference masks.
The cohort design is shown in \figref[a]{fig:main}.

Lesion segmentation was evaluated at centres~1--3, and lesion screening at centres~3 and~4.
Each cohort was split at the patient level into 80\% training and 20\% testing.
Models were evaluated zero-shot and after fine-tuning on the training split, using the same test patients.
Two junior radiologists (3 and 5 years of experience) and two senior radiologists (13 and 17 years of experience) independently reviewed the original CT volumes of the test patients in shuffled order, without access to reference diagnoses or model outputs.
Screening performance of models and readers was measured by F1, sensitivity, specificity and accuracy.
Diagnostic correctness of each model was compared with that of the highest-accuracy senior reader in each cohort.
Reader comparisons are shown in \figref[c,d]{fig:real_world_clinical_evaluation}.

Abdominal and chest reporting cohorts from centre~5 were reserved for external zero-shot evaluation.
GPT-4o cleaned the Chinese reports, translated them into English and organized them by the Merlin abdominal and RadGenome thoracic region schemes.
HoloCT performed anatomy-grounded report generation, and its report LCS was compared with that of HuluMed-32B.
Report processing is detailed in \suppref{29}.

\subsection{Statistical analysis}\label{sec:statistical-analysis}

Segmentation scores were averaged across cases within each dataset and target, then with equal weight across datasets for each target:
\begin{equation}
    \bar{s}_{\ell}=\frac{1}{|\mathcal{D}_{\ell}|}
    \sum_{d\in\mathcal{D}_{\ell}}
    \left(\frac{1}{n_{d\ell}}\sum_{i=1}^{n_{d\ell}}s_{id\ell}\right).
    \label{eq:target-macro-dice}
\end{equation}
Here, $s_{id\ell}$ is the score for case $i$, dataset $d$ and target $\ell$; $n_{d\ell}$ is the number of evaluated cases, and $\mathcal{D}_{\ell}$ is the set of datasets containing target $\ell$.
Category and overall summaries weighted target means equally.
Lesion-screening F1 was computed from counts pooled within each dataset and target, followed by the same hierarchical averaging.
The external zero-shot screening comparison used the same nine targets for every model, assigning a score of zero to the six targets not supported by nnU-Net and 3D ResNet-18.
VQA scores were averaged over questions and report LCS over CT scans; finding-level F1 was computed from counts pooled across scans.
Figure-specific aggregations and error-bar definitions are provided in Supplementary Methods.

Confidence intervals were estimated from 1,000 percentile bootstrap resamples of CT scans within datasets (seed~42), keeping all observations from each scan together.
Paired conditions used identical resamples, and all aggregation steps were repeated for each resample.
The 2.5th and 97.5th percentiles defined 95\% confidence intervals.

All hypothesis tests were two-sided, and $P$ values were not adjusted for multiple comparisons.
Segmentation comparisons used paired $t$-tests of target-level scores $\bar{s}_{\ell}$.
Screening, VQA and report comparisons, including model--reader comparisons, used paired permutation tests with 10,000 permutations, in which the two conditions' complete predictions for each CT scan were exchanged with probability 0.5 before the metric was recomputed.
Score drops after grounding-request removal were compared between intact and masked attention using paired $t$-tests, with pairing by target for screening and by CT scan for VQA and report generation.

Abstract summary values for HoloCT-8B were computed as follows. (1) The 18.2-point clinical segmentation improvement is the macro average over the five clinical lesion categories of, per category, the equally weighted mean of two Dice differences: zero-shot HoloCT-8B minus zero-shot VoxTell, and fine-tuned HoloCT-8B minus specialist nnU-Net. (2) The 15.3-point report-generation improvement is the equally weighted mean, over the abdominal and chest cohorts of centre 5, of the LCS difference between grounded HoloCT-8B and HuluMed-32B. (3) The 85.1\% screening accuracy is the equally weighted mean of per-cohort accuracies (115/141 thyroid; 147/166 colon). (4) The 22.1-point open-vocabulary margin is the difference in macro Dice between HoloCT-8B and VoxTell over 16 equally weighted items: 10 external unseen targets (PANORAMA targets were excluded because they were used in VoxTell training), the five clinical categories evaluated zero-shot, and overall finding-level Dice on ReXGroundingCT.

\subsection{Computational hardware and software}\label{sec:computational-hardware-software}

All models were trained on NVIDIA H100 GPUs.
HoloCT-4B and HoloCT-8B each used 16 GPUs across two nodes.
nnU-Net and 3D ResNet-18 each used two GPUs, whereas M3D, HuluMed-32B and SAT-Pro each used eight.
HoloCT training used gradient checkpointing and DeepSpeed ZeRO-3, with optimizer states offloaded to CPU memory~\citep{Rajbhandari2020ZeRO}.
Training required approximately 3,500 GPU hours for HoloCT-4B and 4,500 GPU hours for HoloCT-8B.

The HoloCT software environment used Python 3.10.16, PyTorch 2.6.0 and CUDA runtime 12.4.127.
Core libraries included Transformers 4.57.0, MONAI 1.5.0, DeepSpeed 0.17.5, PEFT 0.17.1 and Accelerate 1.10.0.
Image processing and evaluation used NiBabel 5.2.1, SimpleITK 2.5.2, NumPy 1.26.4, SciPy 1.14.0, scikit-learn 1.7.1 and surface-distance 0.1.

\section*{Acknowledgements}

We thank the researchers and institutions that made the public CT datasets used in this study available. This work was supported by Alibaba DAMO Academy.

\bibliographystyle{sn-nature}
\bibliography{sn-bibliography}
\BeginNaturalFigurePages
\EndNaturalFigurePages

\BeginNaturalFigurePages
\EndNaturalFigurePages

\BeginNaturalFigurePages
\begin{figure}[p]
    \NaturalPageCenteredGraphic{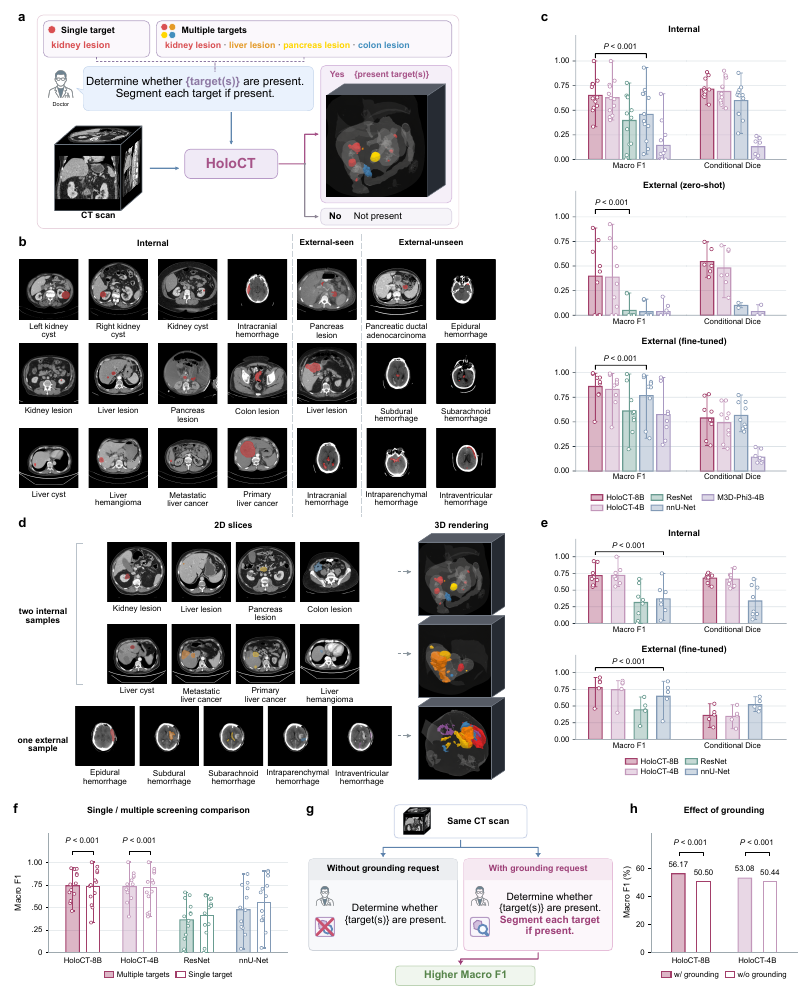}
\caption{\textbf{Single-target and multi-target lesion screening.}
\textbf{a}, Screening workflow with lesion classification and 3D masks.
\textbf{b},\textbf{d}, Single- and multi-target examples, respectively.
\textbf{c},\textbf{e}, Macro F1 and conditional Dice on true-positive cases for single- and multi-target screening, respectively.
\textbf{f}, Single- versus multi-target macro F1 across 13 matched public lesions.
\textbf{g}, Prompts with and without mask requests.
\textbf{h}, Grounding comparison averaged across 18 paired internal and external zero-shot lesions.
In \textbf{c},\textbf{e},\textbf{f}, bars show lesion means, circles denote lesions, and whiskers span minima--maxima.
Filled and open bars in \textbf{h} show means with and without grounding.
Brackets indicate unadjusted $P<0.001$ from two-sided scan-level paired permutation tests.}
\label{fig:grounded_screening}
\end{figure}
\EndNaturalFigurePages

\BeginNaturalFigurePages
\EndNaturalFigurePages

\BeginNaturalFigurePages
\begin{figure}[p]
    \NaturalPageCenteredGraphic{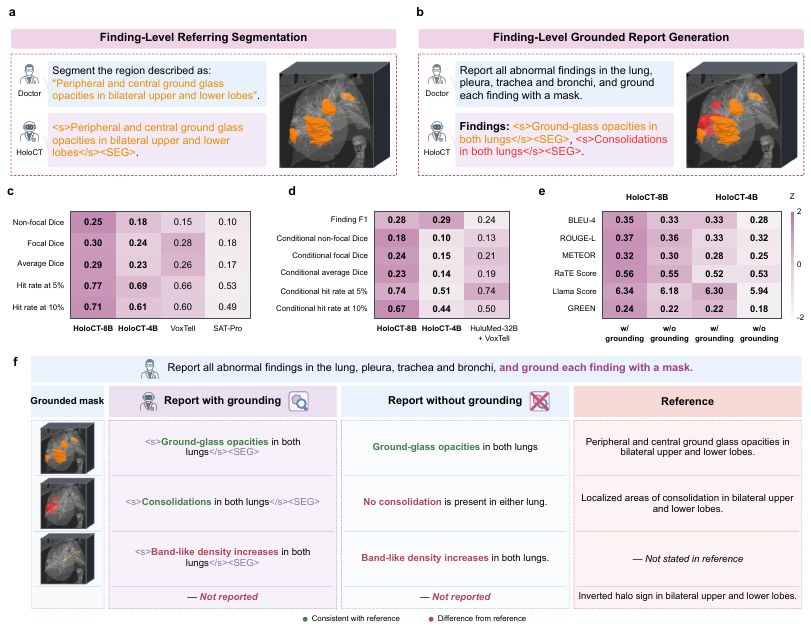}
\caption{\textbf{Referring segmentation and finding-grounded report generation.}
\textbf{a},\textbf{b}, Task illustrations for finding-level referring segmentation and finding-grounded report generation, respectively.
\textbf{c}, ReXGroundingCT referring-segmentation Dice and spatial hit rates; $\mathrm{HIT}_{5\%}$ and $\mathrm{HIT}_{10\%}$ denote the fractions of findings with Dice above 0.05 and 0.10.
\textbf{d}, Finding-detection F1 and conditional localization compared with a two-stage pipeline pairing HuluMed-32B with VoxTell.
\textbf{e}, Report-quality metrics with and without 3D mask requests; heatmap values show raw means and colours indicate within-metric standardized scores.
\textbf{f}, Representative grounded and text-only reports; green and red text indicate agreement and discrepancy with the reference.}
\label{fig:grounded_report_generation}
\end{figure}
\EndNaturalFigurePages

\BeginNaturalFigurePages
\EndNaturalFigurePages

\BeginNaturalFigurePages
\begin{figure}[p]
    \NaturalPageCenteredGraphic{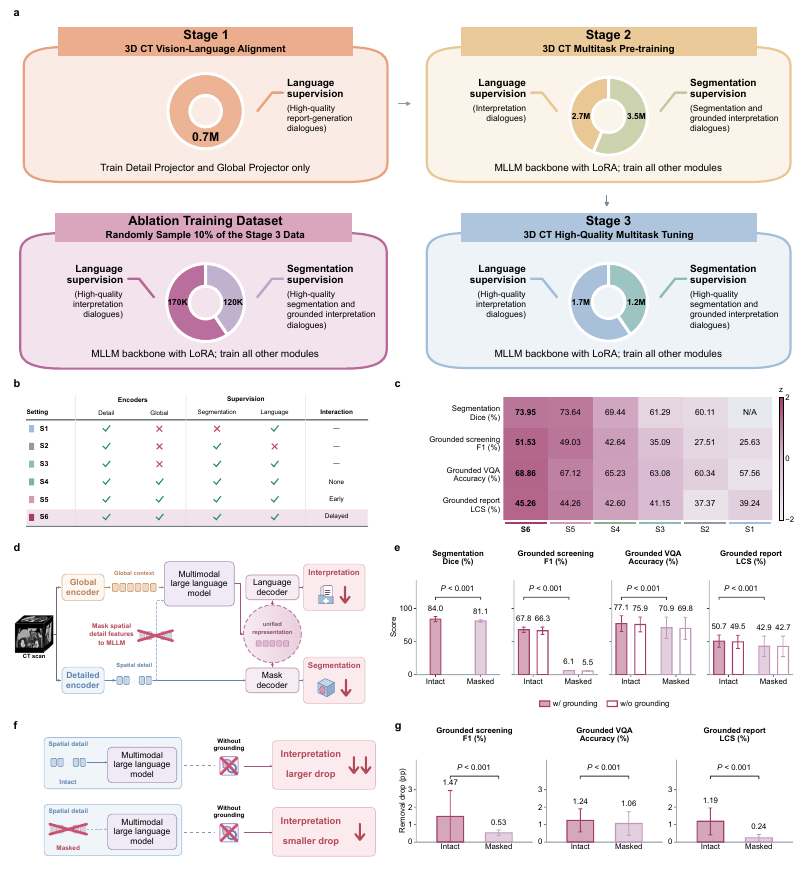}
\caption{\textbf{Training, dual-scale fusion, and spatial-attention ablations.}
\textbf{a}, Three-stage training curriculum and the 10\% Stage 3 subset combined with the complete training splits of the evaluation datasets for S1--S6.
\textbf{b}, Ablation settings S1--S6; S5 exchanges projected tokens prior to the language model, and S6 feeds both visual streams directly into it.
\textbf{c}, Task scores (\%), with colours indicating within-task standardized scores; segmentation is not applicable (N/A) to the text-only configuration S1.
\textbf{d}, Masking language-model attention to high-resolution spatial tokens with skip connections from the encoder to the mask decoder kept intact.
\textbf{e}, Performance under intact or masked spatial attention with mask requests retained (filled bars) or removed (open bars).
\textbf{f}, Design for removing the 3D mask request.
\textbf{g}, Score reductions (in percentage points) upon removing the mask request under intact versus masked spatial attention.
In \textbf{e},\textbf{g}, bars show means across the two evaluation datasets per task and whiskers span the two dataset-level values.
Brackets indicate unadjusted $P<0.001$: two-sided paired $t$-tests for segmentation in \textbf{e} and score reductions in \textbf{g}, and scan-level paired permutation tests for other tasks in \textbf{e}.}
\label{fig:training_ablation}
\end{figure}
\EndNaturalFigurePages

\BeginNaturalFigurePages
\setcounter{figure}{0}
\renewcommand{\figurename}{Extended Data Fig.}
\renewcommand{\theHfigure}{ED\arabic{figure}}
\begin{figure}[p]
    \NaturalPageCenteredGraphic{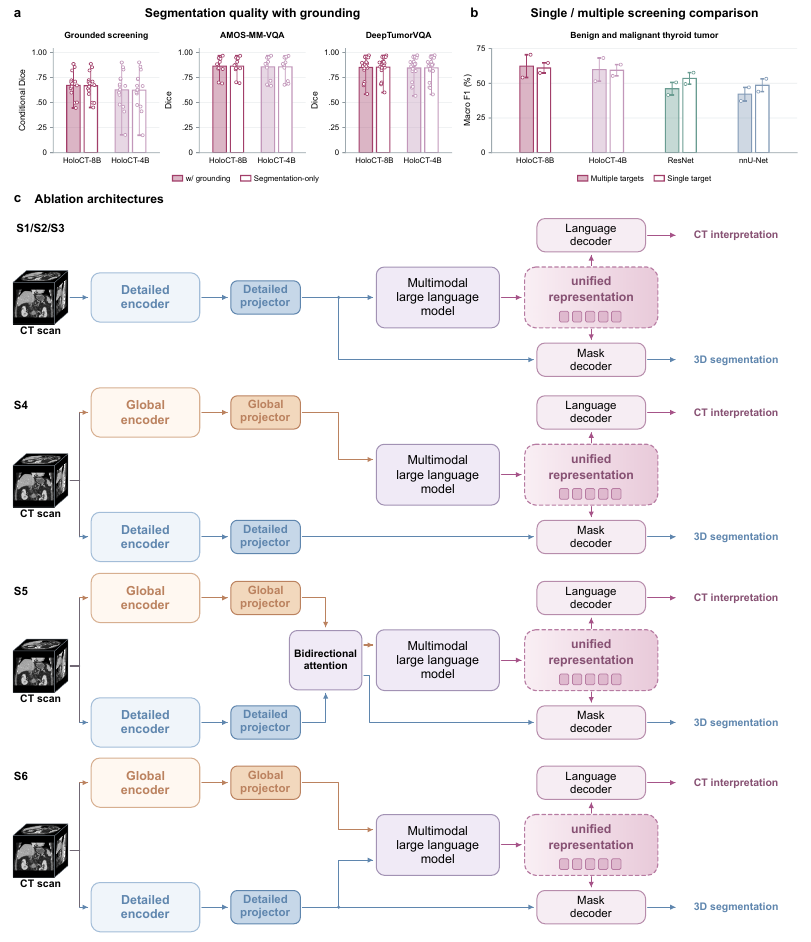}
\caption{\textbf{Segmentation quality, clinical screening, and ablation configurations.}
\textbf{a}, Grounded versus segmentation-only requests: conditional Dice across 15 screening lesions and Dice across 12 AMOS-MM-VQA and 20 DeepTumorVQA categories.
\textbf{b}, Fine-tuned single- versus multi-target thyroid screening ($n=141$ scans), with benign and malignant tumours shown separately.
In \textbf{a},\textbf{b}, bars show category means, circles denote categories, and whiskers span minima--maxima.
\textbf{c}, S1--S3 share a high-resolution encoder across different training tasks; S4 separates global language and high-resolution mask pathways; S5 adds bidirectional feature exchange after projection; S6 supplies both projected visual streams to the language model. High-resolution modules are labelled `Detailed' in \textbf{c}.}
\label{fig:extended_grounding}
\label{fig:extended_architecture}
\label{fig:extended_ablation_architectures}
\label{fig:extended_single_encoder}
\label{fig:extended_interaction}
\end{figure}
\clearpage

\begin{figure}[p]
    \NaturalPageCenteredGraphic{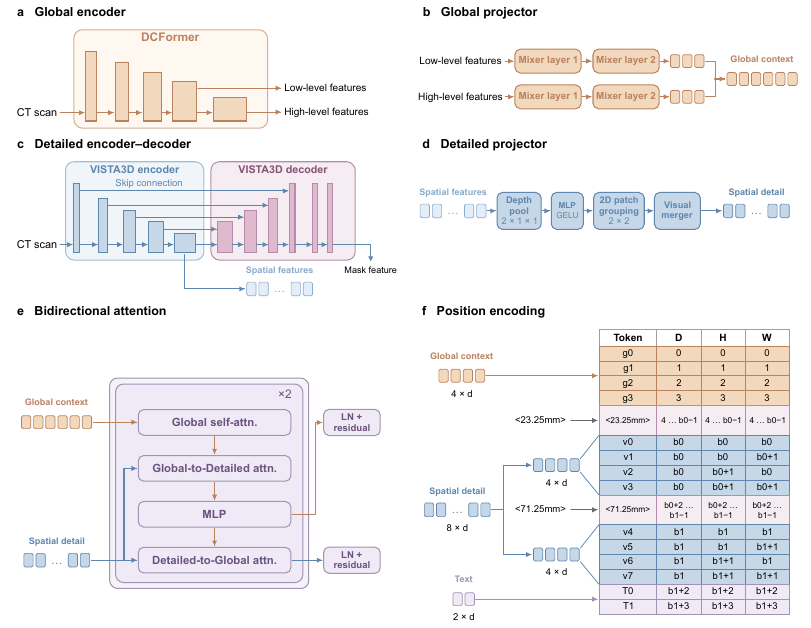}
\caption{\textbf{HoloCT encoder, projection, and position-encoding components.}
\textbf{a}, DCFormer global encoder.
\textbf{b}, Global projector with parallel Mixer streams.
\textbf{c}, VISTA3D high-resolution encoder and mask decoder with multiscale skip connections.
\textbf{d}, High-resolution projector with depth pooling, feature projection, patch grouping, and visual merging.
\textbf{e}, Bidirectional attention after visual projection in ablation configuration S5.
\textbf{f}, Three-axis position-index example for global, high-resolution, and language tokens. High-resolution modules are labelled `Detailed' in \textbf{c}--\textbf{e}.}
\label{fig:extended_components}
\end{figure}
\clearpage

\EndNaturalFigurePages

\normalfont\normalsize
\setlength{\parindent}{1.5em}
\setlength{\parskip}{0pt}
\let\textbf\HoloRomanTextbf
\captionsetup{format=plain}
\newgeometry{margin=1in}
\savegeometry{supplement}
\newcommand{\datasetlink}[1]{\href{#1}{Download}}
\providecommand{\bibcommenthead}{}

\pagestyle{fancy}
\fancyhf{}
\fancyfoot[C]{\thepage}
\renewcommand{\headrulewidth}{0pt}
\renewcommand{\footrulewidth}{0pt}
\setlength{\headwidth}{\textwidth}
\fancyhfoffset[L,R]{0pt}

\fancypagestyle{plain}{
    \fancyhf{}
    \fancyfoot[C]{\thepage}
    \renewcommand{\headrulewidth}{0pt}
    \renewcommand{\footrulewidth}{0pt}
}

\definecolor{leftblue}{RGB}{232,242,248}
\definecolor{headerblue}{RGB}{218,233,242}

\captionsetup[table]{
    name=Supplementary Table,
    labelfont=bf,
    font=small,
    skip=3pt,
    justification=centering,
    singlelinecheck=false
}

\renewcommand{\thetable}{\arabic{table}}

\renewcommand{\listtablename}{Table of contents}

\renewcommand{\cfttabpresnum}{Supplementary Table~}
\renewcommand{\cfttabaftersnum}{\enspace|\enspace}
\setlength{\cfttabnumwidth}{12em}
\renewcommand{\cfttableader}{\cftdotfill{\cftdotsep}}
\renewcommand{\cfttabfont}{\normalfont\raggedright}
\renewcommand{\cfttabpagefont}{\normalfont}
\setlength{\cftbeforetabskip}{2pt}
\setlength{\cfttabindent}{0pt}

\setlength{\cftsecnumwidth}{0pt}
\renewcommand{\cftsecleader}{\cftdotfill{\cftdotsep}}

\setlength{\tabcolsep}{4pt}
\renewcommand{\arraystretch}{1.20}

\newcolumntype{L}[1]{%
    >{\raggedright\arraybackslash}p{#1}%
}

\newcolumntype{C}[1]{%
    >{\centering\arraybackslash}p{#1}%
}

\newcommand{\CI}{95\% CI}
\newcommand{\placeholder}{--}
\newcommand{\metriccell}{\makecell[c]{--\\(--, --)}}

\providecommand{\softcline}[1]{}

\renewcommand{\softcline}[1]{%
    \arrayrulecolor{black!38}%
    \cline{#1}%
    \arrayrulecolor{black}%
}

\newenvironment{ImportedPortraitTable}{%
    \clearpage
    \begingroup

    \renewcommand{\baselinestretch}{1}\selectfont

    \newgeometry{
        top=0.40in,
        bottom=0.60in,
        left=0.14in,
        right=0.14in,
        footskip=0.26in
    }%

    \setlength{\headwidth}{\textwidth}%
    \fancyhfoffset[L,R]{0pt}%
    \pagestyle{fancy}%
    \thispagestyle{fancy}%

    \captionsetup[table]{
        name=Supplementary Table,
        font=importedportraitcaption,
        labelfont=bf,
        textfont=normalfont,
        labelsep=tablepipe,
        skip=3pt,
        justification=centering,
        singlelinecheck=false
    }%

    \setlength{\LTcapwidth}{18.60cm}%
    \setlength{\LTleft}{\fill}%
    \setlength{\LTright}{\fill}%

    \setlength{\LTpre}{0pt}%
    \setlength{\LTpost}{0pt}%

    \setlength{\tabcolsep}{0.45pt}%
    \renewcommand{\arraystretch}{1.00}%
    \setlength{\extrarowheight}{0pt}%
    \setlength{\arrayrulewidth}{0.48pt}%

    \arrayrulecolor{black}%

    \renewcommand{\softcline}[1]{%
        \arrayrulecolor{black!35}%
        \cline{##1}%
        \arrayrulecolor{black}%
    }%

    \sloppy
    \emergencystretch=2em

    \fontsize{5.30pt}{5.90pt}\selectfont
}{%
    \arrayrulecolor{black}%

    \loadgeometry{supplement}

    \setlength{\headwidth}{\textwidth}%
    \fancyhfoffset[L,R]{0pt}%

    \endgroup
    \clearpage
}

\newenvironment{ImportedDatasetTables}{%
    \clearpage
    \begingroup

    \renewcommand{\baselinestretch}{1}\selectfont

    \newgeometry{
        top=0.40in,
        bottom=0.60in,
        left=0.14in,
        right=0.14in,
        footskip=0.26in
    }%

    \setlength{\headwidth}{\textwidth}%
    \fancyhfoffset[L,R]{0pt}%
    \pagestyle{fancy}%
    \thispagestyle{fancy}%

    \captionsetup[table]{
        name=Supplementary Table,
        font=importedportraitcaption,
        labelfont=bf,
        textfont=normalfont,
        labelsep=tablepipe,
        skip=3pt,
        justification=centering,
        singlelinecheck=false
    }%

    \setlength{\LTcapwidth}{18.60cm}%
    \setlength{\LTleft}{\fill}%
    \setlength{\LTright}{\fill}%
    \setlength{\LTpre}{5pt}%
    \setlength{\LTpost}{7pt}%

    \setlength{\tabcolsep}{0.45pt}%
    \renewcommand{\arraystretch}{3.00}%
    \setlength{\extrarowheight}{0pt}%
    \setlength{\arrayrulewidth}{0.48pt}%

    \arrayrulecolor{black}%

    \renewcommand{\softcline}[1]{%
        \arrayrulecolor{black!35}%
        \cline{##1}%
        \arrayrulecolor{black}%
    }%

    \sloppy
    \emergencystretch=2em

    \fontsize{5.30pt}{5.90pt}\selectfont
}{%
    \arrayrulecolor{black}%

    \loadgeometry{supplement}

    \setlength{\headwidth}{\textwidth}%
    \fancyhfoffset[L,R]{0pt}%

    \endgroup
    \clearpage
}

\newenvironment{ImportedLandscapeTable}{%
    \clearpage
    \begingroup

    \renewcommand{\baselinestretch}{1}\selectfont

    \newgeometry{
        top=0.16in,
        bottom=0.56in,
        left=0.16in,
        right=0.16in,
        footskip=0.24in
    }%

    \begin{landscape}%

    \setlength{\headwidth}{\textwidth}%
    \fancyhfoffset[L,R]{0pt}%
    \pagestyle{fancy}%
    \thispagestyle{fancy}%

    \captionsetup[table]{
        name=Supplementary Table,
        font=importedlandscapecaption,
        labelfont=bf,
        textfont=normalfont,
        labelsep=tablepipe,
        skip=2pt,
        justification=centering,
        singlelinecheck=false,
        width=0.98\linewidth
    }%

    \setlength{\tabcolsep}{0.35pt}%
    \renewcommand{\arraystretch}{0.98}%
    \setlength{\extrarowheight}{0.10pt}%
    \setlength{\arrayrulewidth}{0.50pt}%

    \arrayrulecolor{black}%

    \renewcommand{\softcline}[1]{%
        \arrayrulecolor{black!38}%
        \cline{##1}%
        \arrayrulecolor{black}%
    }%

    \sloppy
    \emergencystretch=2em

    \fontsize{4.55pt}{5.15pt}\selectfont
}{%
    \arrayrulecolor{black}%

    \end{landscape}%
    \loadgeometry{supplement}

    \setlength{\headwidth}{\textwidth}%
    \fancyhfoffset[L,R]{0pt}%

    \endgroup
    \clearpage
}

\newcommand{\InputImportedPortraitTableScoped}[2]{%
    \begin{ImportedPortraitTable}
        \begingroup
            \def\HoloTablePrefix{#1}%
            \let\HoloOriginalLabel\label
            \let\HoloOriginalRef\ref

            \renewcommand{\label}[1]{%
                \HoloOriginalLabel{\HoloTablePrefix:##1}%
            }%

            \renewcommand{\ref}[1]{%
                \HoloOriginalRef{\HoloTablePrefix:##1}%
            }%

            \input{#2}%
        \endgroup
    \end{ImportedPortraitTable}
}

\newcommand{\InputSplitImportedPortraitTableScoped}[3]{%
    \begin{ImportedPortraitTable}
        \begingroup
            \def\HoloTablePrefix{#1}%
            \let\HoloOriginalLabel\label
            \let\HoloOriginalRef\ref

            \renewcommand{\label}[1]{%
                \HoloOriginalLabel{\HoloTablePrefix:##1}%
            }%

            \renewcommand{\ref}[1]{%
                \HoloOriginalRef{\HoloTablePrefix:##1}%
            }%

            \input{#2}%
            \input{#3}%
        \endgroup
    \end{ImportedPortraitTable}
}

\newcommand{\InputImportedLandscapeTableScoped}[2]{%
    \begin{ImportedLandscapeTable}
        \begingroup
            \def\HoloTablePrefix{#1}%
            \let\HoloOriginalLabel\label
            \let\HoloOriginalRef\ref

            \renewcommand{\label}[1]{%
                \HoloOriginalLabel{\HoloTablePrefix:##1}%
            }%

            \renewcommand{\ref}[1]{%
                \HoloOriginalRef{\HoloTablePrefix:##1}%
            }%

            \input{#2}%
        \endgroup
    \end{ImportedLandscapeTable}
}

\newcommand{\InputImportedPortraitTable}[1]{%
    \begin{ImportedPortraitTable}
        \input{#1}%
    \end{ImportedPortraitTable}
}

\newcommand{\InputSplitImportedPortraitTable}[2]{%
    \begin{ImportedPortraitTable}
        \input{#1}%
        \input{#2}%
    \end{ImportedPortraitTable}
}

\newcommand{\InputImportedLandscapeTable}[1]{%
    \begin{ImportedLandscapeTable}
        \input{#1}%
    \end{ImportedLandscapeTable}
}

\newcommand{\InputStandardTable}[1]{%
    \clearpage
    \begingroup
        \input{#1}%
    \endgroup
    \clearpage
}

\phantomsection
\label{sec:supplement-start}
\pdfbookmark[0]{Supplementary Information}{supplement}

\begin{center}
{\Large\textbf{Supplementary Information}}
\end{center}
\vspace{0.5em}

\hypertarget{supp-contents}{}
\makeatletter
\let\HoloSavedBibContext\the@ipfilectr
\listoftables
\global\let\the@ipfilectr\HoloSavedBibContext
\makeatother
\clearpage

\begingroup
\providecommand{\figref}[2][]{Fig.~\ifcsname r@#2\endcsname\ref{#2}\else\ref{main-#2}\fi#1}
\providecommand{\edfigref}[2][]{Extended Data Fig.~\ifcsname r@#2\endcsname\ref{#2}\else\ref{main-#2}\fi#1}
\emergencystretch=3em
\section*{Supplementary Methods}
\hypertarget{supp-methods}{}
\addcontentsline{lot}{section}{Supplementary Methods}
\fontsize{10.5pt}{12.5pt}\selectfont
\setlength{\parindent}{1.5em}
\setlength{\parskip}{0pt}
\clubpenalty=10000
\widowpenalty=10000

\subsection*{Data processing and task construction}
For training, each multi-turn dialogue was used as a single sample.
Only the CT-RATE~\cite{hamamci2026generalist} long- and short-answer data were organized as multi-turn dialogues; all other instruction data were single-turn dialogues.
Stage-specific inventories count training samples, including samples reused across stages, and task-specific mask inventories include repeated uses of shared masks.
Target-level referring segmentation used positive cases only, whereas lesion screening used explicitly annotated positive and negative cases.
Dataset and stage inventories are listed in Supplementary Tables~\ref{tab:dataset_inventory_segmentation}--\ref{tab:training_stage_composition}.

Report-question templates were derived from each source dataset's description and 50 sample reports.
Grounding requests were added to the prompts for grounded tasks.
For anatomy-grounded report generation, thoracic reports were organized into ten RadGenome~\citep{zhang2025development} regions, and abdominal reports were organized according to the Merlin~\citep{blankemeier2026merlin} abdominal region scheme.
Task prompts are listed in Supplementary Tables~\ref{tab:segmentation_prompts}--\ref{tab:grounded_report_prompts}, and data-construction prompts in Supplementary Table~\ref{tab:preprocessing_prompts}.

\noindent\textbf{CT-RATE-Option preprocessing.}\enspace
Image markers and task-specific tags were removed.
We excluded questions without recognizable answer options and responses that did not contain exactly one answer label (A--D).
The remaining questions, options and answers were converted into a standardized format.

\noindent\textbf{Grounded VQA construction.}\enspace
For DeepTumorVQA~\citep{DeepTumorVQA} and AMOS-MM-VQA~\citep{AMOSMM}, masks were selected according to anatomical specificity, laterality and lesion relevance.
Questions about organ measurements required organ masks, and questions about lesion presence, count, size or volume required lesion masks rather than parent-organ masks.
Multiple masks were retained only for bilateral findings or multi-object comparisons.
DeepTumorVQA questions assigned three or more masks were reviewed to retain only the masks needed to support the answer.
Questions without suitable masks received no grounding annotation.

\noindent\textbf{BIMCV report and VQA construction.}~\citep{BIMCVCOVID19}\enspace
We excluded empty reports, reports shorter than 10 or longer than 512 words, and duplicate reports.
Cleaning of the Spanish-to-English translations retained current imaging findings and removed non-imaging content and references to prior studies, without adding unsupported facts.
Cleaned narratives formed BIMCV-RG, and BIMCV-VQA contained up to five open-ended single-turn dialogues per CT scan.
Only non-redundant questions supported by the cleaned report were retained.

\subsection*{Model implementation}

\noindent\textbf{CT preprocessing and augmentation.}\enspace
Masks underwent the same cropping, resampling and padding as the detailed input, using nearest-neighbour interpolation and zero padding.
Spatial augmentation applied identical random transformations to the global input, the detailed input and the masks.
Each spatial axis was flipped independently with probability 0.2, and rotations of $90^{\circ}$, $180^{\circ}$ or $270^{\circ}$ in the plane of the last two spatial axes were applied with probability 0.2.
For intensity augmentation, scaled intensities were multiplied by a factor sampled uniformly from $[0.9,1.1]$ with probability 0.1 and shifted by an offset in $[-0.1,0.1]$ with probability 0.5.

\noindent\textbf{Network architecture.}\enspace
The global encoder was adapted from the Med3DVLM DCFormer~\citep{Xin2025Med3DVLM}, with five stages of 64, 96, 192, 384 and 768 channels containing 1, 2, 3, 6 and 2 blocks, respectively.
Its decomposed 3D convolutions summed the outputs of three axis-aligned branches.
The penultimate feature map had 256 spatial locations and 384 channels, and the final map had 32 locations and 768 channels.
The global projector passed each map through two Mixer layers, each using two-layer GELU multilayer perceptrons (MLPs) for token and channel mixing, and concatenated the outputs into 256 whole-volume tokens.

The detailed encoder had five resolution levels with 48, 96, 192, 384 and 768 channels containing 1, 2, 2, 4 and 4 residual blocks, respectively.
Four strided operations downsampled the input by a factor of 16 along each axis.
The detailed projector applied average pooling with a kernel and stride of 2 along the depth axis and 1 along the other axes, followed by a GELU MLP and the Qwen3-VL visual merger~\citep{Qwen3VL2025}, which combined each $2\times2$ in-plane neighbourhood into one fine-grained token of the language model's hidden dimension.
These operations gave an overall stride of 32 along each axis.
The encoders and projectors are shown in \edfigref[a--d]{fig:extended_components}.

\noindent\textbf{Visual tokens and position embedding.}\enspace
Whole-volume tokens followed \mbox{\textless{}global\_feature\textgreater{}}, and each plane of fine-grained tokens followed \mbox{\textless{}local\_feature\textgreater{}}.
The chat template combined these blocks with the query using the Qwen3-VL vision-start and vision-end delimiters.
Position indices advanced sequentially through text and delimiters, and each visual block started after the largest preceding index, which defined $b_k$.
Depth markers were tokenized as ordinary text and could span multiple positions.

\noindent\textbf{Target representation and mask decoding.}\enspace
Each target description was enclosed by \textless{}s\textgreater{} and \textless{}/s\textgreater{}.
These two tokens, \textless{}SEG\textgreater{}, \mbox{\textless{}global\_feature\textgreater{}} and \mbox{\textless{}local\_feature\textgreater{}} were added to the tokenizer vocabulary, and the embedding matrix was expanded accordingly.
The projection from the \textless{}SEG\textgreater{} hidden state to $q_k$ was a GELU MLP.
The mask decoder upsampled the bottleneck features of the detailed encoder through four stages with skip connections, followed by two residual 3D convolutional blocks.
Output restoration removed padding and reversed resampling, cropping and reorientation, preserving the original affine transformation.

\noindent\textbf{Alternative interaction pathways.}\enspace
In variants S4 and S5, the mask decoder received fine-grained features from the detailed encoder and \textless{}SEG\textgreater{} hidden states from the language model.
S5 additionally applied two successive two-way attention blocks~\citep{Kirillov2023SAM} to the projected whole-volume and fine-grained tokens before the language model.
Each block had 16 attention heads and residual connections and applied, in order, self-attention among whole-volume tokens, cross-attention from whole-volume queries to fine-grained keys and values, an MLP, and reverse cross-attention from fine-grained queries to whole-volume keys and values.
Queries and keys used 3D rotary position embeddings.
After both blocks, each token stream was added to its projected input features.
The updated whole-volume tokens entered the language model, and the updated fine-grained tokens were passed to the mask decoder.
These pathways are shown in \edfigref[c]{fig:extended_architecture}, and the two-way attention block in \edfigref[e]{fig:extended_components}.

\subsection*{Training implementation and ablation benchmarks}
\noindent\textbf{Initialization.}\enspace
The visual merger in the detailed projector inherited Qwen3-VL weights, and mask-decoder layers retained from VISTA3D inherited VISTA3D weights~\citep{He2024VISTA3DAU}.
The remaining projector layers, the \textless{}SEG\textgreater{} projection, the embeddings of newly added tokens and the S5 two-way attention blocks were randomly initialized.

\noindent\textbf{Sampling and loss computation.}\enspace
Language-supervised and segmentation-supervised samples were drawn from separately shuffled pools, and samples were repeated as needed to complete each distributed sampling cycle.
During training and validation, each GPU processed one example per minibatch, and gradients were accumulated over four minibatches per optimizer update.
Supervised response positions included the response-end marker and excluded system and user tokens, visual placeholders and padding.
Mask losses were computed in float32, and BCE was computed from logits for numerical stability.

\noindent\textbf{Ablation benchmarks.}\enspace
Training and inference ablations used the same benchmarks, preprocessing and prompts.
Target-level referring segmentation was evaluated on TotalSegmentatorV2~\citep{wasserthal2023totalsegmentator} and AbdomenAtlas3.0~\citep{RadGPT} using macro Dice, and single-target lesion screening was evaluated on AbdomenAtlas3.0 and LiMT~\citep{LiMT} using macro F1.
Grounded VQA was evaluated on AMOS-MM-VQA and DeepTumorVQA using accuracy.
Anatomy-grounded and finding-grounded report generation were evaluated on RadGenome and ReXGroundingCT~\citep{Baharoon2025ChestCTGrounding}, respectively, using report LCS.
Prompts for textual outputs included grounding requests, except in the inference settings in which these requests were removed.
Segmentation was not evaluated for S1, which received no segmentation supervision.
Results are shown in \figref{fig:training_ablation} and \edfigref[c]{fig:extended_architecture}, and scores are listed in Supplementary Tables~\ref{ST23:tab:ablation_study_detailed_scores} and~\ref{ST24:tab:ablation_study_detailed_scores}.

\subsection*{Baseline training}

\noindent\textbf{nnU-Net.}\enspace
nnU-Net~\citep{Isensee2021nnUNet} used its automatically planned 3D full-resolution configuration and batch size.
Following SAT~\citep{Zhao2023LargevocabularySF}, TotalSegmentator labels were divided into six non-overlapping subsets, each assigned a separate model, whereas a single model was trained for each other dataset.
For datasets with overlapping annotations, labels were converted to mutually exclusive classes for training only, with higher-index labels taking precedence.
Models were trained for 1,000 epochs of 250 iterations each, using SGD with an initial learning rate of 0.01 to minimize the sum of Dice and cross-entropy losses.

\noindent\textbf{3D ResNet-18.}\citep{He2016ResNet}\enspace
Inputs were resized to $128^3$ voxels, with intensities clipped to $[-1000,1000]$~HU.
Models were trained for 100 epochs with a batch size of 100 in bfloat16 precision, using AdamW~\citep{Loshchilov2019AdamW} with a fixed learning rate of $10^{-4}$ and weight decay of $10^{-4}$ to minimize class-weighted cross-entropy loss.

\noindent\textbf{M3D and HuluMed-32B.}\enspace
M3D~\citep{Bai2024M3DA3} inputs were min--max normalized and resized to $32\times256\times256$ voxels.
HuluMed-32B~\citep{Jiang2025HuluMed} processed 180 axial slices with a spatial merge size of 2 and token compression.
Both models were trained for three epochs with a batch size of 64, using AdamW with a learning rate of $5\times10^{-5}$, 3\% warmup followed by cosine decay, and weight decay of 0.01; HuluMed-32B was trained in bfloat16 precision.
LoRA~\citep{Hu2022LoRA} used $r=16$, $\alpha=32$ and dropout 0.05 for M3D, and $r=64$, $\alpha=128$ and dropout 0.05 for HuluMed-32B.

\noindent\textbf{SAT-Pro.}\enspace
The image encoder, BioLORD-based text encoder and projector were updated.
Inputs were reoriented to RAS, resampled to $1\times1\times3$~mm spacing, clipped to $[-500,1000]$~HU and z-score normalized per scan.
Training crops contained $288\times288\times64$ voxels, with a patch size of $32^3$ voxels.
The model was trained for 25 epochs with a batch size of 16 in bfloat16 precision, using AdamW with a learning rate of $10^{-4}$ and 2,000 warmup steps.

\subsection*{Evaluation implementation}
Evaluation prompts and target-name mappings are listed in Supplementary Tables~\ref{tab:segmentation_prompts}--\ref{tab:grounded_report_prompts} and~\ref{tab:segmentation_label_hierarchy}, and benchmark scores in Supplementary Tables~\ref{ST5:tab:id_ood_segmentation_scores}--\ref{ST22:tab:conditional_dice_zeroshot_comparison}.

\noindent\textbf{Segmentation and lesion screening.}\enspace
Predicted probability maps were binarized at 0.5, and predicted and reference masks were scored on the reference voxel grid without further resampling.
Foreground comprised nonzero voxels in binary masks or the target label in multiclass masks; reference labels came from the original annotations, and each target was scored independently.
Pairs in which both masks were empty were excluded, and pairs with one empty mask received Dice~0.
HD95~\citep{Taha2015SegmentationMetrics} was computed using the DeepMind surface-distance implementation.
When HoloCT generated a mask for the same target more than once, only the first mask channel was evaluated.

\noindent\textbf{Segmentation baselines.}\enspace
VISTA3D was prompted with class labels only, without points or boxes, and masks of multiple classes mapped to one target were merged by union.
Text-prompted baselines received each model's own target names when available and HoloDATA target names otherwise.
VoxTell~\citep{Rokuss2025VoxTellFP} predictions were restored from the cropped region to native space.
Targets unsupported by nnU-Net or VISTA3D received Dice~0.

\noindent\textbf{Closed-ended and grounded VQA.}\enspace
Questions with unparseable reference answers were excluded.
Visual grounding was scored independently of answer accuracy.
Object-set exact match required the predicted object set to equal the reference set, with no unknown or duplicate objects.

\noindent\textbf{Open-ended VQA and report generation.}\enspace
BLEU-4~\citep{Papineni2002BLEU} was computed with smoothing and ROUGE-L~\citep{Lin2004ROUGE} with stemming.
When several references were available, the first was used; empty predictions received a score of zero.
The LLM judge~\citep{Dubey2024Llama3} used temperature~0 throughout, with separate evaluation prompts for VQA and report generation.
For VQA, the judge received the question, question type, reference and prediction, and assessed accuracy, relevance, completeness and clinical soundness.
RaTEScore~\citep{Zhao2024RaTEScore} used RaTE-NER-DeBERTa for entity extraction and BioLORD-2023-C for semantic comparison.

\noindent\textbf{Anatomy-grounded report generation.}\enspace
Grounded and ungrounded report generation were compared on the same scans, with the same reference reports and evaluation prompts.
Before text scoring, grounding tokens were removed from generated reports, leaving clinical wording, anatomical labels and their order unchanged.

\noindent\textbf{Finding-grounded report generation.}\enspace
Predicted findings were extracted from finding tags in HoloCT reports and, for the pipeline baseline, by splitting HuluMed-32B reports at commas; both sets were then evaluated identically.
Contradictory clinical attributes assessed by the LLM judge included laterality, negation, distribution and spatial relationships.
Ties between highest-Dice predictions were resolved in favour of the earliest prediction.

\subsection*{Metric aggregation and figure summaries}

\noindent\textbf{Lesion screening.}\enspace
For each of the 18 targets in \figref[h]{fig:grounded_screening}, scores were averaged over all dataset--task settings, covering single-target and multi-target lesion screening in internal and external zero-shot evaluations.
Each model's summary score was the unweighted mean of its 18 target scores.

\noindent\textbf{VQA.}\enspace
In \figref[a]{fig:vqa_rg}, AMOS-MM-VQA and DeepTumorVQA scores were unweighted means of question-category accuracies, whereas \figref[c]{fig:vqa_rg} shows the mean of the two datasets' question-level accuracies.

\noindent\textbf{Report generation.}\enspace
Segmentation scores for anatomy-grounded report generation were averaged with equal weight across anatomical regions.

\noindent\textbf{Ablation.}\enspace
In \figref[c,e]{fig:training_ablation}, ablation scores were averaged within each dataset and then with equal weight across datasets.
The score drop after grounding-request removal was the score with grounding requests minus the score without them, computed from the values in \figref[e]{fig:training_ablation}.

\noindent\textbf{Radar plots.}\enspace
Radar plots in \figref[d,f]{fig:seg} and \figref[a,e,f]{fig:vqa_rg} normalized each axis to the HoloCT-8B score for visualization only; statistical tests used unnormalized scores.

\noindent\textbf{Error bars.}\enspace
Whiskers in \figref[a,b,e]{fig:real_world_clinical_evaluation} show CT-level bootstrap 95\% confidence intervals for clinical lesion segmentation and report LCS.
Whiskers in \figref[a,b]{fig:seg}, \figref[c,e,f]{fig:grounded_screening} and \edfigref[a,b]{fig:extended_grounding} span the minimum and maximum target scores.
In \figref[e]{fig:training_ablation}, whiskers span the two dataset-level scores, and in \figref[g]{fig:training_ablation}, the two dataset-level score drops.
\par
\endgroup

\clearpage

\setcounter{table}{0}

\begin{ImportedDatasetTables}
\setlength{\arrayrulewidth}{0.48pt}
\newcommand{\InventoryNameRule}[1]{%
    \noalign{\vskip-0.48pt\global\arrayrulewidth=0.85pt}%
    \hhline{#1}%
    \noalign{\global\arrayrulewidth=0.48pt}%
}
\begin{longtable}{|>{\columncolor{leftblue}\centering\arraybackslash}m{0.95cm}|>{\columncolor{leftblue}\centering\arraybackslash}m{3.15cm}|>{\centering\arraybackslash}m{1.00cm}|>{\centering\arraybackslash}m{0.95cm}|>{\centering\arraybackslash}m{1.15cm}|>{\centering\arraybackslash}m{2.80cm}|>{\centering\arraybackslash}m{3.20cm}|>{\centering\arraybackslash}m{1.75cm}|>{\centering\arraybackslash}m{2.20cm}|}
\caption[Dataset details for segmentation]{\protect\hypertarget{supp-table-1}{}\textbf{Dataset details for segmentation.} Dataset Type: Internal, External or Private. Here, aligned means that a dataset follows the target dataset's split to prevent label leakage.}%
\label{tab:dataset_inventory_segmentation}\\[-0.15ex]
\hline
\rowcolor{headerblue}
\textbf{ID} & \textbf{Dataset Name} & \textbf{3D CTs} & \textbf{Classes} & \textbf{Masks} & \textbf{Body Parts} & \textbf{Split Rule} & \textbf{Dataset Type} & \textbf{Download Links} \\
\hline
\endfirsthead
\caption*{\textbf{Supplementary Table~\ref{tab:dataset_inventory_segmentation} (continued)\enspace|\enspace Dataset details for segmentation.} Dataset Type: Internal, External or Private. Here, aligned means that a dataset follows the target dataset's split to prevent label leakage.}\\[-0.15ex]
\hline
\rowcolor{headerblue}
\textbf{ID} & \textbf{Dataset Name} & \textbf{3D CTs} & \textbf{Classes} & \textbf{Masks} & \textbf{Body Parts} & \textbf{Split Rule} & \textbf{Dataset Type} & \textbf{Download Links} \\
\hline
\endhead
\endfoot
\endlastfoot
001 & AbdomenCT-1k~\cite{AbdomenCT1K} & 957 & 4 & 3,606 & Abdomen, Genitourinary & VISTA3D & Internal & \datasetlink{https://github.com/JunMa11/AbdomenCT-1K} \\
\softcline{1-9}\InventoryNameRule{|-|-|~|~|~|~|~|~|~|}
002 & AMOS22~\cite{AMOS22} & 588 & 15 & 5,167 & Abdomen, Thorax & VISTA3D & Internal & \datasetlink{https://zenodo.org/records/7262581} \\
\softcline{1-9}\InventoryNameRule{|-|-|~|~|~|~|~|~|~|}
003 & MSD-colon~\cite{MSD} & 190 & 1 & 126 & Abdomen & VISTA3D & Internal & \datasetlink{http://medicaldecathlon.com/} \\
\softcline{1-9}\InventoryNameRule{|-|-|~|~|~|~|~|~|~|}
004 & MSD-hepatic-vessel~\cite{MSD} & 443 & 2 & 606 & Abdomen, Vascular & VISTA3D & Internal & \datasetlink{http://medicaldecathlon.com/} \\
\softcline{1-9}\InventoryNameRule{|-|-|~|~|~|~|~|~|~|}
005 & MSD-liver~\cite{MSD} & 201 & 2 & 249 & Abdomen & VISTA3D & Internal & \datasetlink{http://medicaldecathlon.com/} \\
\softcline{1-9}\InventoryNameRule{|-|-|~|~|~|~|~|~|~|}
006 & MSD-lung~\cite{MSD} & 95 & 1 & 63 & Thorax & VISTA3D & Internal & \datasetlink{http://medicaldecathlon.com/} \\
\softcline{1-9}\InventoryNameRule{|-|-|~|~|~|~|~|~|~|}
007 & MSD-pancreas~\cite{MSD} & 420 & 2 & 562 & Abdomen & VISTA3D & Internal & \datasetlink{http://medicaldecathlon.com/} \\
\softcline{1-9}\InventoryNameRule{|-|-|~|~|~|~|~|~|~|}
008 & MSD-spleen~\cite{MSD} & 61 & 1 & 41 & Abdomen & VISTA3D & Internal & \datasetlink{http://medicaldecathlon.com/} \\
\hline
\end{longtable}
\clearpage
\begin{longtable}{|>{\columncolor{leftblue}\centering\arraybackslash}m{0.65cm}|>{\centering\arraybackslash}m{3.25cm}|>{\centering\arraybackslash}m{1cm}|>{\centering\arraybackslash}m{2.85cm}|>{\centering\arraybackslash}m{1cm}|>{\centering\arraybackslash}m{1.1cm}|>{\centering\arraybackslash}m{1.15cm}|>{\centering\arraybackslash}m{3.2cm}|>{\centering\arraybackslash}m{1.75cm}|>{\centering\arraybackslash}m{2.2cm}|}
\caption[Dataset details for grounded lesion screening]{\protect\hypertarget{supp-table-classification}{}\textbf{Dataset details for grounded lesion screening.} Dataset Type: Internal, External or Private. Here, aligned means that a dataset follows the target dataset's split to prevent label leakage. Masks equal the number of positive CTs for mask-annotated labels; Private04 has classification labels only and no masks. Positive and Negative are full-cohort label-specific counts; NR denotes not reported. The Private04 held-out test set contains 166 CTs (95 positive, 71 negative); these test counts are not full-cohort counts. Total CTs count each listed dataset once across its labels; overlapping dataset variants are not deduplicated.}%
\label{tab:dataset_inventory_classification}\\[-0.15ex]
\hline
\rowcolor{headerblue}
\textbf{ID} & \textbf{Dataset Name} & \textbf{3D CTs} & \textbf{Label Name} & \textbf{Masks} & \textbf{Positive} & \textbf{Negative} & \textbf{Split Rule} & \textbf{Dataset Type} & \textbf{Download Links} \\
\hline
\endfirsthead
\caption*{\textbf{Supplementary Table~\ref{tab:dataset_inventory_classification} (continued)\enspace|\enspace}\textbf{Dataset details for grounded lesion screening.} Dataset Type: Internal, External or Private. Here, aligned means that a dataset follows the target dataset's split to prevent label leakage. Masks equal the number of positive CTs for mask-annotated labels; Private04 has classification labels only and no masks. Positive and Negative are full-cohort label-specific counts; NR denotes not reported. The Private04 held-out test set contains 166 CTs (95 positive, 71 negative); these test counts are not full-cohort counts. Total CTs count each listed dataset once across its labels; overlapping dataset variants are not deduplicated.}\\[-0.15ex]
\hline
\rowcolor{headerblue}
\textbf{ID} & \textbf{Dataset Name} & \textbf{3D CTs} & \textbf{Label Name} & \textbf{Masks} & \textbf{Positive} & \textbf{Negative} & \textbf{Split Rule} & \textbf{Dataset Type} & \textbf{Download Links} \\
\hline
\endhead
\endfoot
\endlastfoot
001 &  &  & left kidney cyst & 30 & 30 & 1,198 &  &  &  \\*
\softcline{4-7}\InventoryNameRule{|-|~|~|~|~|~|~|~|~|~|}
002 & \multirow{-2}{=}{\centering Total-SegmentatorV2~\cite{wasserthal2023totalsegmentator}} & \multirow{-2}{=}{\centering 1,228} & right kidney cyst & 25 & 25 & 1,203 & \multirow{-2}{=}{\centering VISTA3D} & \multirow{-2}{=}{\centering Internal} & \multirow{-2}{=}{\centering \datasetlink{https://zenodo.org/records/6802614}} \\
\softcline{1-10}\InventoryNameRule{|-|-|~|~|~|~|~|~|~|~|}
003 & KiTS23~\cite{KiTS23} & 489 & kidney cyst & 248 & 248 & 241 & SAT & Internal & \datasetlink{https://github.com/neheller/kits23} \\
\softcline{1-10}\InventoryNameRule{|-|-|~|~|~|~|~|~|~|~|}
004 & CT-ICH~\cite{CTICH} & 75 & intracranial hemorrhage & 36 & 36 & 39 & random & Internal & \datasetlink{https://physionet.org/content/ct-ich/1.3.0/} \\
\softcline{1-10}\InventoryNameRule{|-|-|~|~|~|~|~|~|~|~|}
005 & PanTS~\cite{PanTS} & 9,901 & pancreas lesion & 1,032 & 1,032 & 8,869 & random & Internal & \datasetlink{https://github.com/MrGiovanni/PanTS} \\
\softcline{1-10}\InventoryNameRule{|-|-|~|~|~|~|~|~|~|~|}
006 &  &  & kidney lesion & 2,222 & 2,222 & 7,040 &  &  &  \\*
\softcline{4-7}\InventoryNameRule{|-|~|~|~|~|~|~|~|~|~|}
007 &  &  & liver lesion & 1,508 & 1,508 & 7,754 &  &  &  \\*
\softcline{4-7}\InventoryNameRule{|-|~|~|~|~|~|~|~|~|~|}
008 &  &  & pancreas lesion & 1,113 & 1,113 & 8,149 &  &  &  \\
\hline
\end{longtable}
\clearpage
\begin{longtable}{|>{\columncolor{leftblue}\centering\arraybackslash}m{3cm}|>{\centering\arraybackslash}m{1.1cm}|>{\centering\arraybackslash}m{3.75cm}|>{\centering\arraybackslash}m{3.00cm}|>{\centering\arraybackslash}m{2.15cm}|>{\centering\arraybackslash}m{1.75cm}|>{\centering\arraybackslash}m{2.2cm}|}
\caption[Dataset details for visual question answering]{\protect\hypertarget{supp-table-2}{}\textbf{Dataset details for visual question answering.} Dataset Type: Internal, External or Private. Here, aligned means that a dataset follows the target dataset's split to prevent label leakage.}%
\label{tab:dataset_inventory_rg_vqa}\\[-0.15ex]
\hline
\rowcolor{headerblue}
\textbf{Dataset Name} & \textbf{3D CTs} & \textbf{Single-turn QAs} & \textbf{Body Parts} & \textbf{Split Rule} & \textbf{Dataset Type} & \textbf{Download Links} \\
\hline
\endfirsthead
\caption*{\textbf{Supplementary Table~\ref{tab:dataset_inventory_rg_vqa} (continued)\enspace|\enspace}\textbf{Dataset details for visual question answering.} Dataset Type: Internal, External or Private. Here, aligned means that a dataset follows the target dataset's split to prevent label leakage.}\\[-0.15ex]
\hline
\rowcolor{headerblue}
\textbf{Dataset Name} & \textbf{3D CTs} & \textbf{Single-turn QAs} & \textbf{Body Parts} & \textbf{Split Rule} & \textbf{Dataset Type} & \textbf{Download Links} \\
\hline
\endhead
\endfoot
\endlastfoot
CT-RATE-option~\cite{hamamci2026generalist} & 46,609 & 135,384 & Thorax, Vascular, Musculoskeletal & official & Internal & \datasetlink{https://huggingface.co/datasets/ibrahimhamamci/CT-RATE} \\
\softcline{1-7}\InventoryNameRule{|-|~|~|~|~|~|~|}
CT-RATE-long\&short~\cite{hamamci2026generalist} & 50,175 & \makecell[c]{2,472,480\\(within 669,798 multi-turn QAs)} & Thorax, Vascular, Musculoskeletal & official & Internal & \datasetlink{https://huggingface.co/datasets/ibrahimhamamci/CT-RATE} \\
\softcline{1-7}\InventoryNameRule{|-|~|~|~|~|~|~|}
3D-RAD~\cite{ThreeDRAD} & 16,188 & 170,105 & Thorax, Vascular, Musculoskeletal & official & Internal & \datasetlink{https://huggingface.co/datasets/Tang-xiaoxiao/3D-RAD} \\
\softcline{1-7}\InventoryNameRule{|-|~|~|~|~|~|~|}
RadGenome-VQA~\cite{zhang2025development} & 25,692 & 481,752 & Thorax, Vascular, Musculoskeletal & official & Internal & \datasetlink{https://huggingface.co/datasets/RadGenome/RadGenome-ChestCT} \\
\softcline{1-7}\InventoryNameRule{|-|~|~|~|~|~|~|}
AMOS-MM-VQA~\cite{AMOSMM} & 1,687 & 16,538 & Abdomen, Genitourinary, Vascular, Musculoskeletal & AMOS-22-aligned & Internal & \datasetlink{https://era-ai-biomed.github.io/amos/dataset.html} \\
\softcline{1-7}\InventoryNameRule{|-|~|~|~|~|~|~|}
FLARE-2025-VQA~\cite{FLARE2025} & 3,917 & 37,795 & Abdomen, Genitourinary, Vascular, Musculoskeletal & AMOS-MM-aligned & Internal & \datasetlink{https://huggingface.co/datasets/FLARE-MedFM/FLARE-Task5-MLLM-3D} \\
\softcline{1-7}\InventoryNameRule{|-|~|~|~|~|~|~|}
DeepTumorVQA~\cite{DeepTumorVQA} & 9,255 & 439,973 & Abdomen, Genitourinary, Vascular, Musculoskeletal & AbdomenAtlas3.0-aligned & Internal & \datasetlink{https://github.com/Schuture/DeepTumorVQA} \\
\softcline{1-7}\InventoryNameRule{|-|~|~|~|~|~|~|}
BIMCV-VQA~\cite{BIMCVCOVID19} & 6,991 & 26,434 & Thorax, Vascular, Musculoskeletal & random & External & \datasetlink{https://bimcv.cipf.es/bimcv-projects/bimcv-covid19/} \\
\hline
\end{longtable}
\begin{longtable}{|>{\columncolor{leftblue}\centering\arraybackslash}m{3cm}|>{\centering\arraybackslash}m{1.1cm}|>{\centering\arraybackslash}m{2.15cm}|>{\centering\arraybackslash}m{3.55cm}|>{\centering\arraybackslash}m{2.8cm}|>{\centering\arraybackslash}m{1.75cm}|>{\centering\arraybackslash}m{2.2cm}|}
\caption[Dataset details for report generation]{\protect\hypertarget{supp-table-rg}{}\textbf{Dataset details for report generation.} Dataset Type: Internal, External or Private. Here, aligned means that a dataset follows the target dataset's split to prevent label leakage.}%
\label{tab:dataset_inventory_rg}\\[-0.15ex]
\hline
\rowcolor{headerblue}
\textbf{Dataset Name} & \textbf{3D CTs} & \textbf{Single-turn QAs} & \textbf{Body Parts} & \textbf{Split Rule} & \textbf{Dataset Type} & \textbf{Download Links} \\
\hline
\endfirsthead
\caption*{\textbf{Supplementary Table~\ref{tab:dataset_inventory_rg} (continued)\enspace|\enspace}\textbf{Dataset details for report generation.} Dataset Type: Internal, External or Private. Here, aligned means that a dataset follows the target dataset's split to prevent label leakage.}\\[-0.15ex]
\hline
\rowcolor{headerblue}
\textbf{Dataset Name} & \textbf{3D CTs} & \textbf{Single-turn QAs} & \textbf{Body Parts} & \textbf{Split Rule} & \textbf{Dataset Type} & \textbf{Download Links} \\
\hline
\endhead
\endfoot
\endlastfoot
CT-RATE-RG~\cite{hamamci2026generalist} & 50,175 & 50,175 & Thorax, Vascular, Musculoskeletal & official & Internal & \datasetlink{https://huggingface.co/datasets/ibrahimhamamci/CT-RATE} \\
\softcline{1-7}\InventoryNameRule{|-|~|~|~|~|~|~|}
RadGenome~\cite{zhang2025development} & 25,692 & 223,643 & Thorax, Vascular, Musculoskeletal & official & Internal & \datasetlink{https://huggingface.co/datasets/RadGenome/RadGenome-ChestCT} \\
\softcline{1-7}\InventoryNameRule{|-|~|~|~|~|~|~|}
INSPECT~\cite{INSPECT} & 23,226 & 23,226 & Thorax, Vascular, Musculoskeletal & random & Internal & \datasetlink{https://aimi.stanford.edu/datasets/inspect-Multimodal-Dataset-for-Pulmonary-Embolism-Diagnosis-and-Prognosis} \\
\softcline{1-7}\InventoryNameRule{|-|~|~|~|~|~|~|}
Merlin~\cite{blankemeier2026merlin} & 25,372 & 328,318 & Abdomen, Genitourinary, Vascular, Musculoskeletal & random & Internal & \datasetlink{https://stanfordaimi.azurewebsites.net/datasets/60b9c7ff-877b-48ce-96c3-0194c8205c40} \\
\softcline{1-7}\InventoryNameRule{|-|~|~|~|~|~|~|}
AMOS-MM-RG~\cite{AMOSMM} & 1,687 & 5,350 & Abdomen, Genitourinary, Vascular, Musculoskeletal & AMOS-22-aligned & Internal & \datasetlink{https://era-ai-biomed.github.io/amos/dataset.html} \\
\softcline{1-7}\InventoryNameRule{|-|~|~|~|~|~|~|}
FLARE-2025-RG~\cite{FLARE2025} & 3,917 & 35,295 & Abdomen, Genitourinary, Vascular, Musculoskeletal & AMOS-MM-aligned & Internal & \datasetlink{https://huggingface.co/datasets/FLARE-MedFM/FLARE-Task5-MLLM-3D} \\
\softcline{1-7}\InventoryNameRule{|-|~|~|~|~|~|~|}
AbdomenAtlas3.0~\cite{RadGPT} & 9,262 & 9,262 & Abdomen, Genitourinary, Vascular, Musculoskeletal & Several-dataset-aligned & Internal & \datasetlink{https://github.com/MrGiovanni/RadGPT} \\
\softcline{1-7}\InventoryNameRule{|-|~|~|~|~|~|~|}
BIMCV-RG~\cite{BIMCVCOVID19} & 6,991 & 6,991 & Thorax, Vascular, Musculoskeletal & random & External & \datasetlink{https://bimcv.cipf.es/bimcv-projects/bimcv-covid19/} \\
\hline
\end{longtable}
\clearpage
\begin{longtable}{|>{\columncolor{leftblue}\centering\arraybackslash}m{3.1cm}|>{\centering\arraybackslash}m{1cm}|>{\centering\arraybackslash}m{1.35cm}|>{\centering\arraybackslash}m{1.8cm}|>{\centering\arraybackslash}m{3.2cm}|>{\centering\arraybackslash}m{2.85cm}|>{\centering\arraybackslash}m{1.75cm}|>{\centering\arraybackslash}m{2.2cm}|}
\caption[Dataset details for grounded closed-ended visual question answering and grounded report generation]{\protect\hypertarget{supp-table-3}{}\textbf{Dataset details for grounded closed-ended visual question answering and grounded report generation.} Dataset Type: Internal, External or Private. Here, aligned means that a dataset follows the target dataset's split to prevent label leakage. This table lists public grounding-supervision sources. Private report-evaluation cohorts are listed in Supplementary Table~\ref{tab:dataset_inventory_rg}; their text results appear in Supplementary Table~17. RadGenome-grounded contains 192,242 + 25,690 = 217,932 single-turn QAs.}%
\label{tab:dataset_inventory_grounded_vqa}\\[-0.15ex]
\hline
\rowcolor{headerblue}
\textbf{Dataset Name} & \textbf{3D CTs} & \textbf{Masks} & \textbf{Single-turn QAs} & \textbf{Body Parts} & \textbf{Split Rule} & \textbf{Dataset Type} & \textbf{Download Links} \\
\hline
\endfirsthead
\caption*{\textbf{Supplementary Table~\ref{tab:dataset_inventory_grounded_vqa} (continued)\enspace|\enspace}\textbf{Dataset details for grounded closed-ended visual question answering and grounded report generation.} Dataset Type: Internal, External or Private. Here, aligned means that a dataset follows the target dataset's split to prevent label leakage. This table lists public grounding-supervision sources. Private report-evaluation cohorts are listed in Supplementary Table~\ref{tab:dataset_inventory_rg}; their text results appear in Supplementary Table~17. RadGenome-grounded contains 192,242 + 25,690 = 217,932 single-turn QAs.}\\[-0.15ex]
\hline
\rowcolor{headerblue}
\textbf{Dataset Name} & \textbf{3D CTs} & \textbf{Masks} & \textbf{Single-turn QAs} & \textbf{Body Parts} & \textbf{Split Rule} & \textbf{Dataset Type} & \textbf{Download Links} \\
\hline
\endhead
\endfoot
\endlastfoot
AMOS-MM-VQA-grounded~\cite{AMOSMM} & 207 & 1,642 & 1,438 & Abdomen, Genitourinary, Vascular, Musculoskeletal & AMOS-22-aligned & Internal & \datasetlink{https://era-ai-biomed.github.io/amos/dataset.html} \\
\softcline{1-8}\InventoryNameRule{|-|~|~|~|~|~|~|~|}
DeepTumorVQA-grounded~\cite{DeepTumorVQA} & 9,169 & 329,929 & 218,226 & Abdomen, Genitourinary, Vascular, Musculoskeletal & AbdomenAtlas3.0-aligned & Internal & \datasetlink{https://github.com/Schuture/DeepTumorVQA} \\
\softcline{1-8}\InventoryNameRule{|-|~|~|~|~|~|~|~|}
RadGenome-VQA-grounded~\cite{zhang2025development} & 25,687 & 162,480 & 468,643 & Thorax, Vascular, Musculoskeletal & CT-RATE-aligned & Internal & \datasetlink{https://huggingface.co/datasets/RadGenome/RadGenome-ChestCT} \\
\softcline{1-8}\InventoryNameRule{|-|~|~|~|~|~|~|~|}
RadGenome-grounded~\cite{zhang2025development} & 25,690 & 192,242 & 217,932 & Thorax, Vascular, Musculoskeletal & CT-RATE-aligned & Internal & \datasetlink{https://huggingface.co/datasets/RadGenome/RadGenome-ChestCT} \\
\softcline{1-8}\InventoryNameRule{|-|~|~|~|~|~|~|~|}
ReXGroundingCT~\cite{Baharoon2025ChestCTGrounding} & 3,042 & 7,802 & 10,844 & Thorax, Vascular, Musculoskeletal & CT-RATE-aligned & Internal & \datasetlink{https://huggingface.co/datasets/rajpurkarlab/ReXGroundingCT} \\
\softcline{1-8}
\hline
\rowcolor{headerblue}
\textbf{TOTAL} & \textbf{63,795} & \textbf{694,095} & \textbf{917,083} & \textbf{} & \textbf{} & \textbf{} & \textbf{} \\
\hline
\end{longtable}
\setlength{\arrayrulewidth}{0.48pt}
\end{ImportedDatasetTables}

\setcounter{table}{5}
\begin{ImportedDatasetTables}
\fontsize{7pt}{8.8pt}\selectfont
\renewcommand{\arraystretch}{1.65}
\setlength{\tabcolsep}{3pt}
\begin{longtable}{|>{\raggedright\arraybackslash}m{2.7cm}|>{\centering\arraybackslash}m{1.7cm}|>{\raggedright\arraybackslash}m{2.2cm}|>{\raggedright\arraybackslash}m{4.0cm}|>{\raggedright\arraybackslash}m{6.0cm}|>{\centering\arraybackslash}m{1.5cm}|}
\caption[Dataset composition across training stages]{\protect\hypertarget{supp-table-training}{}\textbf{Dataset composition across training stages.} Each dialogue represents one training sample: a single-turn dialogue and a multi-turn dialogue each count as one sample. Counts are stage-specific and expressed in millions (M); they are not sums of the single-turn QA counts in the source inventories.}%
\label{tab:training_stage_composition}\\
\hline
\rowcolor{headerblue}
\textbf{Training stage} & \textbf{Supervision stream} & \textbf{Source data} & \textbf{Dialogue types included} & \textbf{Stage-specific selection} & \textbf{No. of dialogues} \\
\hline
\endfirsthead
\caption*{\textbf{Supplementary Table~\ref{tab:training_stage_composition} (continued)\enspace|\enspace Dataset composition across training stages.}}\\
\hline
\rowcolor{headerblue}
\textbf{Training stage} & \textbf{Supervision stream} & \textbf{Source data} & \textbf{Dialogue types included} & \textbf{Stage-specific selection} & \textbf{No. of dialogues} \\
\hline
\endhead
\endfoot
\endlastfoot
\textbf{Stage 1: 3D CT vision--language alignment} & Language & Supplementary Table~\ref{tab:dataset_inventory_rg} & Report generation & High-quality report-generation dialogues & \textbf{0.68M} \\
\hline

\multirow{2}{=}{\textbf{Stage 2: 3D CT multitask pre-training}} & Language & Supplementary Tables~\ref{tab:dataset_inventory_classification}--\ref{tab:dataset_inventory_rg} & Single-target and multi-target lesion screening without grounding; closed-ended and open-ended VQA; narrative and structured report generation & All eligible training dialogues from the corresponding training splits & \textbf{2.70M} \\*
\cline{2-6}
 & Segmentation & Supplementary Tables~\ref{tab:dataset_inventory_segmentation}, \ref{tab:dataset_inventory_classification} and~\ref{tab:dataset_inventory_grounded_vqa} & Target-level and finding-level referring segmentation; single-target and multi-target lesion screening with grounding; grounded VQA; anatomy-grounded and finding-grounded report generation & All eligible training dialogues from the corresponding training splits & \textbf{3.50M} \\
\hline
\multirow{2}{=}{\textbf{Stage 3: 3D CT high-quality multitask tuning}} & Language & Supplementary Tables~\ref{tab:dataset_inventory_classification}--\ref{tab:dataset_inventory_rg} & Single-target and multi-target lesion screening without grounding; closed-ended and open-ended VQA; narrative and structured report generation & Single-region report-generation dialogues were excluded; RadGenome-VQA dialogues were excluded & \textbf{1.69M} \\
\hline
\hline
\end{longtable}
\end{ImportedDatasetTables}

\setcounter{table}{6}

\begin{ImportedDatasetTables}
\renewcommand{\arraystretch}{1.65}
\begin{longtable}{|>{\columncolor{leftblue}\centering\arraybackslash}m{2.65cm}|>{\centering\arraybackslash}m{3.75cm}|>{\centering\arraybackslash}m{9.40cm}|>{\centering\arraybackslash}m{2.55cm}|}
\caption[Segmentation label hierarchy and counts]{\protect\hypertarget{supp-table-4}{}\textbf{Segmentation label hierarchy and counts.}}%
\label{tab:segmentation_label_hierarchy}\\[-0.15ex]
\hline
\rowcolor{headerblue}
& & \multicolumn{2}{c|}{\textbf{Standard Terms and Counts}} \\
\cline{3-4}
\rowcolor{headerblue}
\multirow{-2}{2.65cm}{\centering\textbf{Primary Category}} & \multirow{-2}{3.75cm}{\centering\textbf{Secondary Category}} & \textbf{ID} & \textbf{OOD} \\
\hline
\endfirsthead
\caption*{\textbf{Supplementary Table~\ref{tab:segmentation_label_hierarchy} (continued)\enspace|\enspace Segmentation label hierarchy and counts.}}\\[-0.15ex]
\hline
\rowcolor{headerblue}
& & \multicolumn{2}{c|}{\textbf{Standard Terms and Counts}} \\
\cline{3-4}
\rowcolor{headerblue}
\multirow{-2}{2.65cm}{\centering\textbf{Primary Category}} & \multirow{-2}{3.75cm}{\centering\textbf{Secondary Category}} & \textbf{ID} & \textbf{OOD} \\
\hline
\endhead
\endfoot
\endlastfoot
& Brain \& CNS & brain(962), brainstem(210), optic chiasm(210), pituitary gland(162), left temporal lobe(120), right temporal lobe(120), left hippocampus(120), right hippocampus(120), intracranial hemorrhage(357) & epidural hemorrhage(32), subdural hemorrhage(158), subarachnoid hemorrhage(142), intraparenchymal hemorrhage(142), intraventricular hemorrhage(120) \\
\softcline{2-4}
& Sensory Organs & left cochlea(162), right cochlea(162), left anterior eyeball(42), right anterior eyeball(42), left posterior eyeball(42), right posterior eyeball(42), left lacrimal gland(42), right lacrimal gland(42), left optic nerve(210), right optic nerve(210), left eyeball(652), right eyeball(648), left lens(120), right lens(120), left middle ear(120), right middle ear(120), left internal auditory canal(120), right internal auditory canal(120), left tympanic cavity(120), right tympanic cavity(120), left vestibule semicircular canal(120), right vestibule semicircular canal(120), left eustachian tube bone(120), right eustachian tube bone(120), left mastoid process(120), right mastoid process(120) & -- \\
\softcline{2-4}
& Upper Aerodigestive Tract & arytenoid(42), buccal mucosa(42), oral cavity(162), cricopharyngeal inlet(42), cervical esophagus(42), larynx glottis(162), larynx supraglottis(162), lips(42), nasal cavity(533), larynx(120), pharynx constrictor muscle(120), nasopharyngeal tumor(120) & laryngeal cancer(439) \\
\hline
\end{longtable}
\end{ImportedDatasetTables}

\setcounter{table}{7}
\InputSplitImportedPortraitTableScoped
    {ST5}
    {table1_subtask1_part1.tex}%
    {table1_subtask1_part2.tex}

\setcounter{table}{8}
\InputImportedPortraitTableScoped
    {ST6}
    {table2_subtask2.tex}

\setcounter{table}{9}
\InputImportedPortraitTableScoped
    {ST7}
    {table3_subtask3.tex}

\setcounter{table}{10}
\InputImportedPortraitTableScoped
    {ST8}
    {table4_close_ended_vqa.tex}

\setcounter{table}{11}
\InputImportedPortraitTableScoped
    {ST9}
    {table5_open_ended_vqa.tex}

\setcounter{table}{12}
\InputImportedPortraitTableScoped
    {ST10}
    {table6_narrative_report_generation.tex}

\setcounter{table}{13}
\InputImportedPortraitTableScoped
    {ST14}
    {table7_structured_report_generation.tex}

\setcounter{table}{14}
\InputImportedPortraitTableScoped
    {ST15}
    {table8_grounded_vqa_text_quality.tex}

\setcounter{table}{15}
\InputImportedPortraitTableScoped
    {ST16}
    {table9_grounded_vqa_segmentation_quality.tex}

\setcounter{table}{16}
\InputImportedPortraitTableScoped
    {ST17}
    {table10_grounded_report_generation_text_quality.tex}

\setcounter{table}{17}
\InputImportedPortraitTableScoped
    {ST18}
    {table11_grounded_report_generation_radgenome_segmentation.tex}

\setcounter{table}{18}
\InputImportedPortraitTableScoped
    {ST19}
    {table12_finding_level_referring_segmentation.tex}

\setcounter{table}{19}
\InputImportedPortraitTableScoped
    {ST20}
    {table13_grounded_report_generation_rexgrounding_segmentation.tex}

\setcounter{table}{20}
\InputImportedPortraitTableScoped
    {ST22}
    {table17_subtask2_seg.tex}

\setcounter{table}{21}
\InputImportedPortraitTableScoped
    {ST23}
    {table15_S1_S6.tex}

\setcounter{table}{22}
\InputImportedPortraitTableScoped
    {ST24}
    {table16_mask_remove.tex}

\loadgeometry{supplement}
\setlength{\headwidth}{\textwidth}
\fancyhfoffset[L,R]{0pt}
\begingroup
\interlinepenalty=10000
\urlstyle{same}
\bibliographystyle{sn-nature}
\bibliography{sn-bibliography}

\begingroup
\providecommand{\bibnumfmt}[1]{#1.}
\renewcommand{\bibnumfmt}[1]{#1.}
\expandafter\def\csname @biblabel\endcsname#1{#1.}
\begin{thebibliography}{10}
\expandafter\ifx\csname url\endcsname\relax
  \def\url#1{\burl{#1}}\fi
\expandafter\ifx\csname urlprefix\endcsname\relax\def\urlprefix{URL }\fi
\providecommand{\bibinfo}[2]{#2}
\providecommand{\eprint}[2][]{\url{#2}}
\providecommand{\doi}[1]{\url{https://doi.org/#1}}
\bibcommenthead

\bibitem{smithbindman2019trends}
\bibinfo{author}{Smith-Bindman, R.} et~al.
\newblock \bibinfo{title}{Trends in use of medical imaging in {US} health care
  systems and in {Ontario}, {Canada}, 2000-2016}.
\newblock \emph{\bibinfo{journal}{JAMA}} \textbf{\bibinfo{volume}{322}},
  \bibinfo{pages}{843--856} (\bibinfo{year}{2019}).

\bibitem{Isensee2021nnUNet}
\bibinfo{author}{Isensee, F.}, \bibinfo{author}{Jaeger, P.~F.},
  \bibinfo{author}{Kohl, S. A.~A.}, \bibinfo{author}{Petersen, J.} \&
  \bibinfo{author}{Maier-Hein, K.~H.}
\newblock \bibinfo{title}{{nnU-Net}: a self-configuring method for deep
  learning-based biomedical image segmentation}.
\newblock \emph{\bibinfo{journal}{Nat. Methods}} \textbf{\bibinfo{volume}{18}},
  \bibinfo{pages}{203--211} (\bibinfo{year}{2021}).

\bibitem{He2024VISTA3DAU}
\bibinfo{author}{He, Y.} et~al.
\newblock \bibinfo{title}{{VISTA3D}: a unified segmentation foundation model
  for {3D} medical imaging}.
\newblock In \emph{\bibinfo{booktitle}{Proc. IEEE/CVF Conference on Computer
  Vision and Pattern Recognition}}, pp. \bibinfo{pages}{20863--20873} (IEEE,
  \bibinfo{year}{2025}).

\bibitem{Zhao2023LargevocabularySF}
\bibinfo{author}{Zhao, Z.} et~al.
\newblock \bibinfo{title}{Large-vocabulary segmentation for medical images with
  text prompts}.
\newblock \emph{\bibinfo{journal}{npj Digit. Med.}}
  \textbf{\bibinfo{volume}{8}}, \bibinfo{pages}{566} (\bibinfo{year}{2025}).

\bibitem{Rokuss2025VoxTellFP}
\bibinfo{author}{Rokuss, M.} et~al.
\newblock \bibinfo{title}{{VoxTell}: free-text promptable universal {3D}
  medical image segmentation}.
\newblock In \emph{\bibinfo{booktitle}{Proc. IEEE/CVF Conference on Computer
  Vision and Pattern Recognition}}, pp. \bibinfo{pages}{37538--37557} (IEEE,
  \bibinfo{year}{2026}).

\bibitem{blankemeier2026merlin}
\bibinfo{author}{Blankemeier, L.} et~al.
\newblock \bibinfo{title}{{Merlin}: a computed tomography vision--language
  foundation model and dataset}.
\newblock \emph{\bibinfo{journal}{Nature}} \textbf{\bibinfo{volume}{652}},
  \bibinfo{pages}{1318--1328} (\bibinfo{year}{2026}).

\bibitem{Hamamci2024GeneralistFM}
\bibinfo{author}{Hamamci, I.~E.} et~al.
\newblock \bibinfo{title}{Generalist foundation models from a multimodal
  dataset for {3D} computed tomography}.
\newblock \emph{\bibinfo{journal}{Nat. Biomed. Eng.}}
  \textbf{\bibinfo{volume}{10}}, \bibinfo{pages}{1610--1628}
  (\bibinfo{year}{2026}).

\bibitem{Wu2025TowardsGF}
\bibinfo{author}{Wu, C.}, \bibinfo{author}{Zhang, X.}, \bibinfo{author}{Zhang,
  Y.}, \bibinfo{author}{Hui, H.}, \bibinfo{author}{Wang, Y.} \&
  \bibinfo{author}{Xie, W.}
\newblock \bibinfo{title}{Towards generalist foundation model for radiology by
  leveraging web-scale {2D}\&{3D} medical data}.
\newblock \emph{\bibinfo{journal}{Nat. Commun.}} \textbf{\bibinfo{volume}{16}},
  \bibinfo{pages}{7866} (\bibinfo{year}{2025}).

\bibitem{CandidateBiomedParse2025}
\bibinfo{author}{Zhao, T.} et~al.
\newblock \bibinfo{title}{A foundation model for joint segmentation, detection
  and recognition of biomedical objects across nine modalities}.
\newblock \emph{\bibinfo{journal}{Nat. Methods}} \textbf{\bibinfo{volume}{22}},
  \bibinfo{pages}{166--176} (\bibinfo{year}{2025}).

\bibitem{CandidateMedVersa2026}
\bibinfo{author}{Zhou, H.-Y.}, \bibinfo{author}{Acosta, J.~N.},
  \bibinfo{author}{Adithan, S.}, \bibinfo{author}{Datta, S.},
  \bibinfo{author}{Topol, E.~J.} \& \bibinfo{author}{Rajpurkar, P.}
\newblock \bibinfo{title}{{MedVersa}: a generalist foundation model for diverse
  medical imaging tasks}.
\newblock \emph{\bibinfo{journal}{NEJM AI}} \textbf{\bibinfo{volume}{3}},
  \bibinfo{pages}{AIoa2500595} (\bibinfo{year}{2026}).

\bibitem{CandidateMedPLIBjournalextension2026}
\bibinfo{author}{Shen, L.} et~al.
\newblock \bibinfo{title}{From image to pixels: towards fine-grained medical
  vision-language models}.
\newblock \emph{\bibinfo{journal}{IEEE Trans. Pattern Anal. Mach. Intell.}}
  \textbf{\bibinfo{volume}{48}}, \bibinfo{pages}{9982--9996}
  (\bibinfo{year}{2026}).

\bibitem{Bai2024M3DA3}
\bibinfo{author}{Bai, F.}, \bibinfo{author}{Du, Y.}, \bibinfo{author}{Huang,
  T.}, \bibinfo{author}{Meng, M. Q.-H.} \& \bibinfo{author}{Zhao, B.}
\newblock \bibinfo{title}{{M3D}: advancing {3D} medical image analysis with
  multi-modal large language models}.
\newblock Preprint at
  {\urlstyle{same}\url{https://doi.org/10.48550/arXiv.2404.00578}}
  (\bibinfo{year}{2024}).

\bibitem{luo2025vividmed}
\bibinfo{author}{Luo, L.}, \bibinfo{author}{Tang, B.}, \bibinfo{author}{Chen,
  X.}, \bibinfo{author}{Han, R.} \& \bibinfo{author}{Chen, T.}
\newblock \bibinfo{title}{{VividMed}: vision language model with versatile
  visual grounding for medicine}.
\newblock In \emph{\bibinfo{booktitle}{Proc. 2025 Conference of the Nations of
  the Americas Chapter of the Association for Computational Linguistics: Human
  Language Technologies (Volume 1: Long Papers)}}, pp.
  \bibinfo{pages}{1800--1821} (ACL, \bibinfo{year}{2025}).

\bibitem{wu2026unibiomed}
\bibinfo{author}{Wu, L.} et~al.
\newblock \bibinfo{title}{A universal foundation model for grounded biomedical
  image interpretation}.
\newblock \emph{\bibinfo{journal}{Nat. Commun.}} \textbf{\bibinfo{volume}{17}},
  \bibinfo{pages}{7173} (\bibinfo{year}{2026}).

\bibitem{CandidateMedVAE2026}
\bibinfo{author}{Varma, M.} et~al.
\newblock \bibinfo{title}{{MedVAE}: efficient automated interpretation of
  medical images with large-scale generalizable autoencoders}.
\newblock In \emph{\bibinfo{booktitle}{Proc. International Conference on
  Medical Imaging with Deep Learning}}, vol. 301, pp.
  \bibinfo{pages}{1597--1626} (PMLR, \bibinfo{year}{2026}).

\bibitem{He2016ResNet}
\bibinfo{author}{He, K.}, \bibinfo{author}{Zhang, X.}, \bibinfo{author}{Ren,
  S.} \& \bibinfo{author}{Sun, J.}
\newblock \bibinfo{title}{Deep residual learning for image recognition}.
\newblock In \emph{\bibinfo{booktitle}{Proc. IEEE Conference on Computer Vision
  and Pattern Recognition}}, pp. \bibinfo{pages}{770--778} (IEEE,
  \bibinfo{year}{2016}).

\bibitem{Qwen3VL2025}
\bibinfo{author}{Bai, S.} et~al.
\newblock \bibinfo{title}{{Qwen3-VL} technical report}.
\newblock Preprint at
  {\urlstyle{same}\url{https://doi.org/10.48550/arXiv.2511.21631}}
  (\bibinfo{year}{2025}).

\bibitem{Sellergren2026MedGemma15}
\bibinfo{author}{Sellergren, A.} et~al.
\newblock \bibinfo{title}{{MedGemma} 1.5 technical report}.
\newblock Preprint at
  {\urlstyle{same}\url{https://doi.org/10.48550/arXiv.2604.05081}}
  (\bibinfo{year}{2026}).

\bibitem{Jiang2025HuluMed}
\bibinfo{author}{Jiang, S.} et~al.
\newblock \bibinfo{title}{{Hulu-Med}: a transparent generalist model towards
  holistic medical vision-language understanding}.
\newblock Preprint at
  {\urlstyle{same}\url{https://doi.org/10.48550/arXiv.2510.08668}}
  (\bibinfo{year}{2025}).

\bibitem{Taha2015SegmentationMetrics}
\bibinfo{author}{Taha, A.~A.} \& \bibinfo{author}{Hanbury, A.}
\newblock \bibinfo{title}{Metrics for evaluating {3D} medical image
  segmentation: analysis, selection, and tool}.
\newblock \emph{\bibinfo{journal}{BMC Med. Imaging}}
  \textbf{\bibinfo{volume}{15}}, \bibinfo{pages}{29} (\bibinfo{year}{2015}).

\bibitem{Papineni2002BLEU}
\bibinfo{author}{Papineni, K.}, \bibinfo{author}{Roukos, S.},
  \bibinfo{author}{Ward, T.} \& \bibinfo{author}{Zhu, W.-J.}
\newblock \bibinfo{title}{{BLEU}: a method for automatic evaluation of machine
  translation}.
\newblock In \emph{\bibinfo{booktitle}{Proc. 40th Annual Meeting of the
  Association for Computational Linguistics}}, pp. \bibinfo{pages}{311--318}
  (ACL, \bibinfo{year}{2002}).

\bibitem{Lin2004ROUGE}
\bibinfo{author}{Lin, C.-Y.}
\newblock \bibinfo{title}{{ROUGE}: a package for automatic evaluation of
  summaries}.
\newblock In \emph{\bibinfo{booktitle}{Proc. Text Summarization Branches Out}},
  pp. \bibinfo{pages}{74--81} (ACL, \bibinfo{year}{2004}).

\bibitem{Banerjee2005METEOR}
\bibinfo{author}{Banerjee, S.} \& \bibinfo{author}{Lavie, A.}
\newblock \bibinfo{title}{{METEOR}: an automatic metric for {MT} evaluation
  with improved correlation with human judgments}.
\newblock In \emph{\bibinfo{booktitle}{Proc. ACL Workshop on Intrinsic and
  Extrinsic Evaluation Measures for Machine Translation and/or Summarization}},
  pp. \bibinfo{pages}{65--72} (ACL, \bibinfo{year}{2005}).

\bibitem{Dubey2024Llama3}
\bibinfo{author}{Grattafiori, A.} et~al.
\newblock \bibinfo{title}{The {Llama} 3 herd of models}.
\newblock Preprint at
  {\urlstyle{same}\url{https://doi.org/10.48550/arXiv.2407.21783}}
  (\bibinfo{year}{2024}).

\bibitem{Zhao2024RaTEScore}
\bibinfo{author}{Zhao, W.}, \bibinfo{author}{Wu, C.}, \bibinfo{author}{Zhang,
  X.}, \bibinfo{author}{Zhang, Y.}, \bibinfo{author}{Wang, Y.} \&
  \bibinfo{author}{Xie, W.}
\newblock \bibinfo{title}{{RaTEScore}: a metric for radiology report
  generation}.
\newblock In \emph{\bibinfo{booktitle}{Proc. 2024 Conference on Empirical
  Methods in Natural Language Processing}}, pp. \bibinfo{pages}{15004--15019}
  (ACL, \bibinfo{year}{2024}).

\bibitem{Ostmeier2024GREEN}
\bibinfo{author}{Ostmeier, S.} et~al.
\newblock \bibinfo{title}{{GREEN}: generative radiology report evaluation and
  error notation}.
\newblock In \emph{\bibinfo{booktitle}{Findings of the Association for
  Computational Linguistics: EMNLP 2024}}, pp. \bibinfo{pages}{374--390} (ACL,
  \bibinfo{year}{2024}).

\bibitem{CandidateRadCliQ2023}
\bibinfo{author}{Yu, F.} et~al.
\newblock \bibinfo{title}{Evaluating progress in automatic chest {X}-ray
  radiology report generation}.
\newblock \emph{\bibinfo{journal}{Patterns}} \textbf{\bibinfo{volume}{4}},
  \bibinfo{pages}{100802} (\bibinfo{year}{2023}).

\bibitem{li2025towards}
\bibinfo{author}{Li, C.-Y.} et~al.
\newblock \bibinfo{title}{Towards a holistic framework for multimodal {LLM} in
  {3D} brain {CT} radiology report generation}.
\newblock \emph{\bibinfo{journal}{Nat. Commun.}} \textbf{\bibinfo{volume}{16}},
  \bibinfo{pages}{2258} (\bibinfo{year}{2025}).

\bibitem{BIMCVCOVID19}
\bibinfo{author}{de~la Iglesia~Vay{\'a}, M.} et~al.
\newblock \bibinfo{title}{{BIMCV} {COVID-19}+: a large annotated dataset of
  {RX} and {CT} images from {COVID-19} patients}.
\newblock Preprint at
  {\urlstyle{same}\url{https://doi.org/10.48550/arXiv.2006.01174}}
  (\bibinfo{year}{2020}).

\bibitem{AMOSMM}
\bibinfo{author}{Ji, Y.}, \bibinfo{author}{Ge, C.}, \bibinfo{author}{Li, R.} \&
  \bibinfo{author}{Luo, P.}
\newblock \bibinfo{title}{{AMOS-MM}: abdominal multimodal analysis challenge:
  structured description of the challenge design}.
\newblock Zenodo {\urlstyle{same}\url{https://doi.org/10.5281/zenodo.10992155}}
  (\bibinfo{year}{2024}).

\bibitem{DeepTumorVQA}
\bibinfo{author}{Chen, Y.} et~al.
\newblock \bibinfo{title}{Are vision language models ready for clinical
  diagnosis? {A} {3D} medical benchmark for tumor-centric visual question
  answering}.
\newblock Preprint at
  {\urlstyle{same}\url{https://doi.org/10.48550/arXiv.2505.18915}}
  (\bibinfo{year}{2025}).

\bibitem{Baharoon2025ChestCTGrounding}
\bibinfo{author}{Baharoon, M.} et~al.
\newblock \bibinfo{title}{State-of-the-art text-prompted medical segmentation
  models struggle to ground chest {CT} findings}.
\newblock In \emph{\bibinfo{booktitle}{Proc. Machine Learning for Healthcare
  Conference}}, vol. 298, pp. \bibinfo{pages}{1--30} (PMLR,
  \bibinfo{year}{2025}).

\bibitem{CandidateGLoRIA2021}
\bibinfo{author}{Huang, S.-C.}, \bibinfo{author}{Shen, L.},
  \bibinfo{author}{Lungren, M.~P.} \& \bibinfo{author}{Yeung, S.}
\newblock \bibinfo{title}{{GLoRIA}: a multimodal global-local representation
  learning framework for label-efficient medical image recognition}.
\newblock In \emph{\bibinfo{booktitle}{Proc. IEEE/CVF International Conference
  on Computer Vision}}, pp. \bibinfo{pages}{3942--3951} (IEEE,
  \bibinfo{year}{2021}).

\bibitem{CandidateCheXzero2022}
\bibinfo{author}{Tiu, E.}, \bibinfo{author}{Talius, E.},
  \bibinfo{author}{Patel, P.}, \bibinfo{author}{Langlotz, C.~P.},
  \bibinfo{author}{Ng, A.~Y.} \& \bibinfo{author}{Rajpurkar, P.}
\newblock \bibinfo{title}{Expert-level detection of pathologies from
  unannotated chest {X-ray} images via self-supervised learning}.
\newblock \emph{\bibinfo{journal}{Nat. Biomed. Eng.}}
  \textbf{\bibinfo{volume}{6}}, \bibinfo{pages}{1399--1406}
  (\bibinfo{year}{2022}).

\bibitem{CandidateKAD2023}
\bibinfo{author}{Zhang, X.}, \bibinfo{author}{Wu, C.}, \bibinfo{author}{Zhang,
  Y.}, \bibinfo{author}{Xie, W.} \& \bibinfo{author}{Wang, Y.}
\newblock \bibinfo{title}{Knowledge-enhanced visual-language pre-training on
  chest radiology images}.
\newblock \emph{\bibinfo{journal}{Nat. Commun.}} \textbf{\bibinfo{volume}{14}},
  \bibinfo{pages}{4542} (\bibinfo{year}{2023}).

\bibitem{CandidateMedITok2025}
\bibinfo{author}{Ma, C.} et~al.
\newblock \bibinfo{title}{Unified medical image tokenizer for autoregressive
  synthesis and understanding}.
\newblock Preprint at
  {\urlstyle{same}\url{https://doi.org/10.48550/arXiv.2505.19225}}
  (\bibinfo{year}{2025}).

\bibitem{CandidateMedVTok2026}
\bibinfo{author}{Ma, C.} et~al.
\newblock \bibinfo{title}{{MedVTok}: a general-purpose medical visual
  tokenizer}.
\newblock In \emph{\bibinfo{booktitle}{Proc. Advances in Neural Information
  Processing Systems}} (\bibinfo{year}{2026}).
\newblock \bibinfo{note}{In press}.

\bibitem{CandidateGMAI2023}
\bibinfo{author}{Moor, M.} et~al.
\newblock \bibinfo{title}{Foundation models for generalist medical artificial
  intelligence}.
\newblock \emph{\bibinfo{journal}{Nature}} \textbf{\bibinfo{volume}{616}},
  \bibinfo{pages}{259--265} (\bibinfo{year}{2023}).

\bibitem{ma2024segment}
\bibinfo{author}{Ma, J.}, \bibinfo{author}{He, Y.}, \bibinfo{author}{Li, F.},
  \bibinfo{author}{Han, L.}, \bibinfo{author}{You, C.} \&
  \bibinfo{author}{Wang, B.}
\newblock \bibinfo{title}{Segment anything in medical images}.
\newblock \emph{\bibinfo{journal}{Nat. Commun.}} \textbf{\bibinfo{volume}{15}},
  \bibinfo{pages}{654} (\bibinfo{year}{2024}).

\bibitem{CandidateLLaVARad2025}
\bibinfo{author}{Zambrano~Chaves, J.~M.} et~al.
\newblock \bibinfo{title}{A clinically accessible small multimodal radiology
  model and evaluation metric for chest {X-ray} findings}.
\newblock \emph{\bibinfo{journal}{Nat. Commun.}} \textbf{\bibinfo{volume}{16}},
  \bibinfo{pages}{3108} (\bibinfo{year}{2025}).

\bibitem{CandidateMMBU2026}
\bibinfo{author}{D'Cunha, R.} et~al.
\newblock \bibinfo{title}{{MMBU}: a massive multi-modal biomedical
  understanding benchmark to probe the perception capabilities of
  vision-language models}.
\newblock In \emph{\bibinfo{booktitle}{Proc. European Conference on Computer
  Vision}}, pp. \bibinfo{pages}{588--604} (ECCV, \bibinfo{year}{2026}).

\bibitem{CandidateMONET2024}
\bibinfo{author}{Kim, C.} et~al.
\newblock \bibinfo{title}{Transparent medical image {AI} via an image--text
  foundation model grounded in medical literature}.
\newblock \emph{\bibinfo{journal}{Nat. Med.}} \textbf{\bibinfo{volume}{30}},
  \bibinfo{pages}{1154--1165} (\bibinfo{year}{2024}).

\bibitem{CandidateConceptCLIP2026}
\bibinfo{author}{Nie, Y.} et~al.
\newblock \bibinfo{title}{An explainable biomedical foundation model via
  large-scale concept-enhanced vision--language pretraining}.
\newblock \emph{\bibinfo{journal}{Nat. Biomed. Eng.}}
  {\urlstyle{same}\url{https://doi.org/10.1038/s41551-026-01764-x}}
  (\bibinfo{year}{2026}).

\bibitem{Tanno2024CollaborationBC}
\bibinfo{author}{Tanno, R.} et~al.
\newblock \bibinfo{title}{Collaboration between clinicians and vision--language
  models in radiology report generation}.
\newblock \emph{\bibinfo{journal}{Nat. Med.}} \textbf{\bibinfo{volume}{31}},
  \bibinfo{pages}{599--608} (\bibinfo{year}{2025}).

\bibitem{CandidateRadiologistAI2024}
\bibinfo{author}{Yu, F.}, \bibinfo{author}{Moehring, A.},
  \bibinfo{author}{Banerjee, O.}, \bibinfo{author}{Salz, T.},
  \bibinfo{author}{Agarwal, N.} \& \bibinfo{author}{Rajpurkar, P.}
\newblock \bibinfo{title}{Heterogeneity and predictors of the effects of {AI}
  assistance on radiologists}.
\newblock \emph{\bibinfo{journal}{Nat. Med.}} \textbf{\bibinfo{volume}{30}},
  \bibinfo{pages}{837--849} (\bibinfo{year}{2024}).

\bibitem{Cardoso2022MONAI}
\bibinfo{author}{Cardoso, M.~J.} et~al.
\newblock \bibinfo{title}{{MONAI}: an open-source framework for deep learning
  in healthcare}.
\newblock Preprint at
  {\urlstyle{same}\url{https://doi.org/10.48550/arXiv.2211.02701}}
  (\bibinfo{year}{2022}).

\bibitem{Xin2025Med3DVLM}
\bibinfo{author}{Xin, Y.}, \bibinfo{author}{Ates, G.~C.},
  \bibinfo{author}{Gong, K.} \& \bibinfo{author}{Shao, W.}
\newblock \bibinfo{title}{{Med3DVLM}: an efficient vision-language model for
  {3D} medical image analysis}.
\newblock \emph{\bibinfo{journal}{IEEE J. Biomed. Health Inform.}}
  \textbf{\bibinfo{volume}{30}}, \bibinfo{pages}{2524--2536}
  (\bibinfo{year}{2026}).

\bibitem{Hu2022LoRA}
\bibinfo{author}{Hu, E.~J.} et~al.
\newblock \bibinfo{title}{{LoRA}: low-rank adaptation of large language
  models}.
\newblock In \emph{\bibinfo{booktitle}{Proc. International Conference on
  Learning Representations}} (ICLR, \bibinfo{year}{2022}).

\bibitem{Loshchilov2019AdamW}
\bibinfo{author}{Loshchilov, I.} \& \bibinfo{author}{Hutter, F.}
\newblock \bibinfo{title}{Decoupled weight decay regularization}.
\newblock In \emph{\bibinfo{booktitle}{Proc. International Conference on
  Learning Representations}} (ICLR, \bibinfo{year}{2019}).

\bibitem{Kirillov2023SAM}
\bibinfo{author}{Kirillov, A.} et~al.
\newblock \bibinfo{title}{Segment anything}.
\newblock In \emph{\bibinfo{booktitle}{Proc. IEEE/CVF International Conference
  on Computer Vision}}, pp. \bibinfo{pages}{4015--4026} (IEEE,
  \bibinfo{year}{2023}).

\bibitem{Rajbhandari2020ZeRO}
\bibinfo{author}{Rajbhandari, S.}, \bibinfo{author}{Rasley, J.},
  \bibinfo{author}{Ruwase, O.} \& \bibinfo{author}{He, Y.}
\newblock \bibinfo{title}{{ZeRO}: memory optimizations toward training trillion
  parameter models}.
\newblock In \emph{\bibinfo{booktitle}{Proc. International Conference for High
  Performance Computing, Networking, Storage and Analysis}}, pp.
  \bibinfo{pages}{1--16} (IEEE, \bibinfo{year}{2020}).

\end{thebibliography}
\endgroup


\begingroup
\providecommand{\bibnumfmt}[1]{#1.}
\renewcommand{\bibnumfmt}[1]{#1.}
\expandafter\def\csname @biblabel\endcsname#1{#1.}
\begin{thebibliography}{10}
\expandafter\ifx\csname url\endcsname\relax
  \def\url#1{\burl{#1}}\fi
\expandafter\ifx\csname urlprefix\endcsname\relax\def\urlprefix{URL }\fi
\providecommand{\bibinfo}[2]{#2}
\providecommand{\eprint}[2][]{\url{#2}}
\providecommand{\doi}[1]{\url{https://doi.org/#1}}
\bibcommenthead

\bibitem{hamamci2026generalist}
\bibinfo{author}{Hamamci, I.~E.} et~al.
\newblock \bibinfo{title}{Generalist foundation models from a multimodal
  dataset for {3D} computed tomography}.
\newblock \emph{\bibinfo{journal}{Nat. Biomed. Eng.}}
  \textbf{\bibinfo{volume}{10}}, \bibinfo{pages}{1610--1628}
  (\bibinfo{year}{2026}).

\bibitem{zhang2025development}
\bibinfo{author}{Zhang, X.} et~al.
\newblock \bibinfo{title}{Development of a large-scale grounded vision language
  dataset for chest {CT} analysis}.
\newblock \emph{\bibinfo{journal}{Sci. Data}} \textbf{\bibinfo{volume}{12}},
  \bibinfo{pages}{1636} (\bibinfo{year}{2025}).

\bibitem{blankemeier2026merlin}
\bibinfo{author}{Blankemeier, L.} et~al.
\newblock \bibinfo{title}{{Merlin}: a computed tomography vision--language
  foundation model and dataset}.
\newblock \emph{\bibinfo{journal}{Nature}} \textbf{\bibinfo{volume}{652}},
  \bibinfo{pages}{1318--1328} (\bibinfo{year}{2026}).

\bibitem{DeepTumorVQA}
\bibinfo{author}{Chen, Y.} et~al.
\newblock \bibinfo{title}{Are vision language models ready for clinical
  diagnosis? {A} {3D} medical benchmark for tumor-centric visual question
  answering}.
\newblock Preprint at
  {\urlstyle{same}\url{https://doi.org/10.48550/arXiv.2505.18915}}
  (\bibinfo{year}{2025}).

\bibitem{AMOSMM}
\bibinfo{author}{Ji, Y.}, \bibinfo{author}{Ge, C.}, \bibinfo{author}{Li, R.} \&
  \bibinfo{author}{Luo, P.}
\newblock \bibinfo{title}{{AMOS-MM}: abdominal multimodal analysis challenge:
  structured description of the challenge design}.
\newblock Zenodo {\urlstyle{same}\url{https://doi.org/10.5281/zenodo.10992155}}
  (\bibinfo{year}{2024}).

\bibitem{BIMCVCOVID19}
\bibinfo{author}{de~la Iglesia~Vay{\'a}, M.} et~al.
\newblock \bibinfo{title}{{BIMCV} {COVID-19}+: a large annotated dataset of
  {RX} and {CT} images from {COVID-19} patients}.
\newblock Preprint at
  {\urlstyle{same}\url{https://doi.org/10.48550/arXiv.2006.01174}}
  (\bibinfo{year}{2020}).

\bibitem{Xin2025Med3DVLM}
\bibinfo{author}{Xin, Y.}, \bibinfo{author}{Ates, G.~C.},
  \bibinfo{author}{Gong, K.} \& \bibinfo{author}{Shao, W.}
\newblock \bibinfo{title}{{Med3DVLM}: an efficient vision-language model for
  {3D} medical image analysis}.
\newblock \emph{\bibinfo{journal}{IEEE J. Biomed. Health Inform.}}
  \textbf{\bibinfo{volume}{30}}, \bibinfo{pages}{2524--2536}
  (\bibinfo{year}{2026}).

\bibitem{Qwen3VL2025}
\bibinfo{author}{Bai, S.} et~al.
\newblock \bibinfo{title}{{Qwen3-VL} technical report}.
\newblock Preprint at
  {\urlstyle{same}\url{https://doi.org/10.48550/arXiv.2511.21631}}
  (\bibinfo{year}{2025}).

\bibitem{Kirillov2023SAM}
\bibinfo{author}{Kirillov, A.} et~al.
\newblock \bibinfo{title}{Segment anything}.
\newblock In \emph{\bibinfo{booktitle}{Proc. IEEE/CVF International Conference
  on Computer Vision}}, pp. \bibinfo{pages}{4015--4026} (IEEE,
  \bibinfo{year}{2023}).

\bibitem{He2024VISTA3DAU}
\bibinfo{author}{He, Y.} et~al.
\newblock \bibinfo{title}{{VISTA3D}: a unified segmentation foundation model
  for {3D} medical imaging}.
\newblock In \emph{\bibinfo{booktitle}{Proc. IEEE/CVF Conference on Computer
  Vision and Pattern Recognition}}, pp. \bibinfo{pages}{20863--20873} (IEEE,
  \bibinfo{year}{2025}).

\bibitem{wasserthal2023totalsegmentator}
\bibinfo{author}{Wasserthal, J.} et~al.
\newblock \bibinfo{title}{{TotalSegmentator}: robust segmentation of 104
  anatomic structures in {CT} images}.
\newblock \emph{\bibinfo{journal}{Radiol. Artif. Intell.}}
  \textbf{\bibinfo{volume}{5}}, \bibinfo{pages}{e230024}
  (\bibinfo{year}{2023}).

\bibitem{RadGPT}
\bibinfo{author}{Bassi, P. R. A.~S.} et~al.
\newblock \bibinfo{title}{{RadGPT}: constructing {3D} image-text tumor
  datasets}.
\newblock In \emph{\bibinfo{booktitle}{Proc. IEEE/CVF International Conference
  on Computer Vision}}, pp. \bibinfo{pages}{23720--23730} (IEEE,
  \bibinfo{year}{2025}).

\bibitem{LiMT}
\bibinfo{author}{Liu, Z.} et~al.
\newblock \bibinfo{title}{{LiMT}: a multi-task liver image benchmark dataset}.
\newblock \emph{\bibinfo{journal}{IEEE J. Biomed. Health Inform.}}
  \textbf{\bibinfo{volume}{30}}, \bibinfo{pages}{5088--5099}
  (\bibinfo{year}{2026}).

\bibitem{Baharoon2025ChestCTGrounding}
\bibinfo{author}{Baharoon, M.} et~al.
\newblock \bibinfo{title}{State-of-the-art text-prompted medical segmentation
  models struggle to ground chest {CT} findings}.
\newblock In \emph{\bibinfo{booktitle}{Proc. Machine Learning for Healthcare
  Conference}}, vol. 298, pp. \bibinfo{pages}{1--30} (PMLR,
  \bibinfo{year}{2025}).

\bibitem{Isensee2021nnUNet}
\bibinfo{author}{Isensee, F.}, \bibinfo{author}{Jaeger, P.~F.},
  \bibinfo{author}{Kohl, S. A.~A.}, \bibinfo{author}{Petersen, J.} \&
  \bibinfo{author}{Maier-Hein, K.~H.}
\newblock \bibinfo{title}{{nnU-Net}: a self-configuring method for deep
  learning-based biomedical image segmentation}.
\newblock \emph{\bibinfo{journal}{Nat. Methods}} \textbf{\bibinfo{volume}{18}},
  \bibinfo{pages}{203--211} (\bibinfo{year}{2021}).

\bibitem{Zhao2023LargevocabularySF}
\bibinfo{author}{Zhao, Z.} et~al.
\newblock \bibinfo{title}{Large-vocabulary segmentation for medical images with
  text prompts}.
\newblock \emph{\bibinfo{journal}{npj Digit. Med.}}
  \textbf{\bibinfo{volume}{8}}, \bibinfo{pages}{566} (\bibinfo{year}{2025}).

\bibitem{He2016ResNet}
\bibinfo{author}{He, K.}, \bibinfo{author}{Zhang, X.}, \bibinfo{author}{Ren,
  S.} \& \bibinfo{author}{Sun, J.}
\newblock \bibinfo{title}{Deep residual learning for image recognition}.
\newblock In \emph{\bibinfo{booktitle}{Proc. IEEE Conference on Computer Vision
  and Pattern Recognition}}, pp. \bibinfo{pages}{770--778} (IEEE,
  \bibinfo{year}{2016}).

\bibitem{Loshchilov2019AdamW}
\bibinfo{author}{Loshchilov, I.} \& \bibinfo{author}{Hutter, F.}
\newblock \bibinfo{title}{Decoupled weight decay regularization}.
\newblock In \emph{\bibinfo{booktitle}{Proc. International Conference on
  Learning Representations}} (ICLR, \bibinfo{year}{2019}).

\bibitem{Bai2024M3DA3}
\bibinfo{author}{Bai, F.}, \bibinfo{author}{Du, Y.}, \bibinfo{author}{Huang,
  T.}, \bibinfo{author}{Meng, M. Q.-H.} \& \bibinfo{author}{Zhao, B.}
\newblock \bibinfo{title}{{M3D}: advancing {3D} medical image analysis with
  multi-modal large language models}.
\newblock Preprint at
  {\urlstyle{same}\url{https://doi.org/10.48550/arXiv.2404.00578}}
  (\bibinfo{year}{2024}).

\bibitem{Jiang2025HuluMed}
\bibinfo{author}{Jiang, S.} et~al.
\newblock \bibinfo{title}{{Hulu-Med}: a transparent generalist model towards
  holistic medical vision-language understanding}.
\newblock Preprint at
  {\urlstyle{same}\url{https://doi.org/10.48550/arXiv.2510.08668}}
  (\bibinfo{year}{2025}).

\bibitem{Hu2022LoRA}
\bibinfo{author}{Hu, E.~J.} et~al.
\newblock \bibinfo{title}{{LoRA}: low-rank adaptation of large language
  models}.
\newblock In \emph{\bibinfo{booktitle}{Proc. International Conference on
  Learning Representations}} (ICLR, \bibinfo{year}{2022}).

\bibitem{Taha2015SegmentationMetrics}
\bibinfo{author}{Taha, A.~A.} \& \bibinfo{author}{Hanbury, A.}
\newblock \bibinfo{title}{Metrics for evaluating {3D} medical image
  segmentation: analysis, selection, and tool}.
\newblock \emph{\bibinfo{journal}{BMC Med. Imaging}}
  \textbf{\bibinfo{volume}{15}}, \bibinfo{pages}{29} (\bibinfo{year}{2015}).

\bibitem{Rokuss2025VoxTellFP}
\bibinfo{author}{Rokuss, M.} et~al.
\newblock \bibinfo{title}{{VoxTell}: free-text promptable universal {3D}
  medical image segmentation}.
\newblock In \emph{\bibinfo{booktitle}{Proc. IEEE/CVF Conference on Computer
  Vision and Pattern Recognition}}, pp. \bibinfo{pages}{37538--37557} (IEEE,
  \bibinfo{year}{2026}).

\bibitem{Papineni2002BLEU}
\bibinfo{author}{Papineni, K.}, \bibinfo{author}{Roukos, S.},
  \bibinfo{author}{Ward, T.} \& \bibinfo{author}{Zhu, W.-J.}
\newblock \bibinfo{title}{{BLEU}: a method for automatic evaluation of machine
  translation}.
\newblock In \emph{\bibinfo{booktitle}{Proc. 40th Annual Meeting of the
  Association for Computational Linguistics}}, pp. \bibinfo{pages}{311--318}
  (ACL, \bibinfo{year}{2002}).

\bibitem{Lin2004ROUGE}
\bibinfo{author}{Lin, C.-Y.}
\newblock \bibinfo{title}{{ROUGE}: a package for automatic evaluation of
  summaries}.
\newblock In \emph{\bibinfo{booktitle}{Proc. Text Summarization Branches Out}},
  pp. \bibinfo{pages}{74--81} (ACL, \bibinfo{year}{2004}).

\bibitem{Dubey2024Llama3}
\bibinfo{author}{Grattafiori, A.} et~al.
\newblock \bibinfo{title}{The {Llama} 3 herd of models}.
\newblock Preprint at
  {\urlstyle{same}\url{https://doi.org/10.48550/arXiv.2407.21783}}
  (\bibinfo{year}{2024}).

\bibitem{Zhao2024RaTEScore}
\bibinfo{author}{Zhao, W.}, \bibinfo{author}{Wu, C.}, \bibinfo{author}{Zhang,
  X.}, \bibinfo{author}{Zhang, Y.}, \bibinfo{author}{Wang, Y.} \&
  \bibinfo{author}{Xie, W.}
\newblock \bibinfo{title}{{RaTEScore}: a metric for radiology report
  generation}.
\newblock In \emph{\bibinfo{booktitle}{Proc. 2024 Conference on Empirical
  Methods in Natural Language Processing}}, pp. \bibinfo{pages}{15004--15019}
  (ACL, \bibinfo{year}{2024}).

\bibitem{AbdomenCT1K}
\bibinfo{author}{Ma, J.} et~al.
\newblock \bibinfo{title}{{AbdomenCT-1K}: is abdominal organ segmentation a
  solved problem?}
\newblock \emph{\bibinfo{journal}{IEEE Trans. Pattern Anal. Mach. Intell.}}
  \textbf{\bibinfo{volume}{44}}, \bibinfo{pages}{6695--6714}
  (\bibinfo{year}{2022}).

\bibitem{AMOS22}
\bibinfo{author}{Ji, Y.} et~al.
\newblock \bibinfo{title}{{AMOS}: a large-scale abdominal multi-organ benchmark
  for versatile medical image segmentation}.
\newblock In \emph{\bibinfo{booktitle}{Proc. Advances in Neural Information
  Processing Systems}}, vol. 35, pp. \bibinfo{pages}{36722--36732} (Curran
  Associates, \bibinfo{year}{2022}).

\bibitem{MSD}
\bibinfo{author}{Antonelli, M.} et~al.
\newblock \bibinfo{title}{The medical segmentation decathlon}.
\newblock \emph{\bibinfo{journal}{Nat. Commun.}} \textbf{\bibinfo{volume}{13}},
  \bibinfo{pages}{4128} (\bibinfo{year}{2022}).

\bibitem{AeroPath}
\bibinfo{author}{St{\o}verud, K.-H.} et~al.
\newblock \bibinfo{title}{{AeroPath}: an airway segmentation benchmark dataset
  with challenging pathology and baseline method}.
\newblock \emph{\bibinfo{journal}{PLoS ONE}} \textbf{\bibinfo{volume}{19}},
  \bibinfo{pages}{e0311416} (\bibinfo{year}{2024}).

\bibitem{VerSe}
\bibinfo{author}{Sekuboyina, A.} et~al.
\newblock \bibinfo{title}{{VerSe}: a vertebrae labelling and segmentation
  benchmark for multi-detector {CT} images}.
\newblock \emph{\bibinfo{journal}{Med. Image Anal.}}
  \textbf{\bibinfo{volume}{73}}, \bibinfo{pages}{102166}
  (\bibinfo{year}{2021}).

\bibitem{PancreasCT}
\bibinfo{author}{Roth, H.}, \bibinfo{author}{Farag, A.},
  \bibinfo{author}{Turkbey, E.~B.}, \bibinfo{author}{Lu, L.},
  \bibinfo{author}{Liu, J.} \& \bibinfo{author}{Summers, R.~M.}
\newblock \bibinfo{title}{Data from {Pancreas-CT}} (version 2).
\newblock \bibinfo{howpublished}{The Cancer Imaging Archive}
  {\urlstyle{same}\url{https://doi.org/10.7937/k9/tcia.2016.tnb1kqbu}}
  (\bibinfo{year}{2016}).

\bibitem{BTCV}
\bibinfo{editor}{Landman, B.}, \bibinfo{editor}{Xu, Z.},
  \bibinfo{editor}{Iglesias, J.~E.}, \bibinfo{editor}{Styner, M.},
  \bibinfo{editor}{Langerak, T.~R.} \& \bibinfo{editor}{Klein, A.} (eds).
\newblock \emph{\bibinfo{title}{2015 {MICCAI} multi-{Atlas} labeling beyond the
  cranial vault: workshop and challenge}}.
\newblock Synapse {\urlstyle{same}\url{https://doi.org/10.7303/syn3193805}}
  (\bibinfo{year}{2015}).

\bibitem{CRLMCT}
\bibinfo{author}{Simpson, A.~L.} et~al.
\newblock \bibinfo{title}{Preoperative {CT} and survival data for patients
  undergoing resection of colorectal liver metastases}.
\newblock \emph{\bibinfo{journal}{Sci. Data}} \textbf{\bibinfo{volume}{11}},
  \bibinfo{pages}{172} (\bibinfo{year}{2024}).

\bibitem{HANSeg}
\bibinfo{author}{Podobnik, G.}, \bibinfo{author}{Strojan, P.},
  \bibinfo{author}{Peterlin, P.}, \bibinfo{author}{Ibragimov, B.} \&
  \bibinfo{author}{Vrtovec, T.}
\newblock \bibinfo{title}{{HaN-Seg}: the head and neck organ-at-risk {CT} and
  {MR} segmentation dataset}.
\newblock \emph{\bibinfo{journal}{Med. Phys.}} \textbf{\bibinfo{volume}{50}},
  \bibinfo{pages}{1917--1927} (\bibinfo{year}{2023}).

\bibitem{KiTS23}
\bibinfo{author}{Heller, N.}, \bibinfo{author}{Isensee, F.},
  \bibinfo{author}{Tejpaul, R.}, \bibinfo{author}{Wood, A.},
  \bibinfo{author}{Papanikolopoulos, N.} \& \bibinfo{author}{Weight, C.}
\newblock \bibinfo{title}{2023 kidney and kidney tumor segmentation challenge}.
\newblock \bibinfo{howpublished}{Zenodo}
  {\urlstyle{same}\url{https://doi.org/10.5281/zenodo.7840133}}
  (\bibinfo{year}{2023}).

\bibitem{KiPA22}
\bibinfo{author}{He, Y.} et~al.
\newblock \bibinfo{title}{Meta grayscale adaptive network for {3D} integrated
  renal structures segmentation}.
\newblock \emph{\bibinfo{journal}{Med. Image Anal.}}
  \textbf{\bibinfo{volume}{71}}, \bibinfo{pages}{102055}
  (\bibinfo{year}{2021}).

\bibitem{CHAOS}
\bibinfo{author}{Kavur, A.~E.} et~al.
\newblock \bibinfo{title}{{CHAOS} challenge - combined ({CT}-{MR}) healthy
  abdominal organ segmentation}.
\newblock \emph{\bibinfo{journal}{Med. Image Anal.}}
  \textbf{\bibinfo{volume}{69}}, \bibinfo{pages}{101950}
  (\bibinfo{year}{2021}).

\bibitem{FLARE22}
\bibinfo{author}{Ma, J.} et~al.
\newblock \bibinfo{title}{Unleashing the strengths of unlabelled data in deep
  learning-assisted pan-cancer abdominal organ quantification: the {FLARE22}
  challenge}.
\newblock \emph{\bibinfo{journal}{Lancet Digit. Health}}
  \textbf{\bibinfo{volume}{6}}, \bibinfo{pages}{e815--e826}
  (\bibinfo{year}{2024}).

\bibitem{FLARE23}
\bibinfo{author}{Ma, J.} et~al.
\newblock \bibinfo{title}{Automatic organ and pan-cancer segmentation in
  abdomen {CT}: the {FLARE} 2023 challenge}.
\newblock Preprint at
  {\urlstyle{same}\url{https://doi.org/10.48550/arXiv.2408.12534}}
  (\bibinfo{year}{2024}).

\bibitem{CTORG}
\bibinfo{author}{Rister, B.}, \bibinfo{author}{Yi, D.},
  \bibinfo{author}{Shivakumar, K.}, \bibinfo{author}{Nobashi, T.} \&
  \bibinfo{author}{Rubin, D.~L.}
\newblock \bibinfo{title}{{CT-ORG}, a new dataset for multiple organ
  segmentation in computed tomography}.
\newblock \emph{\bibinfo{journal}{Sci. Data}} \textbf{\bibinfo{volume}{7}},
  \bibinfo{pages}{381} (\bibinfo{year}{2020}).

\bibitem{LIDCIDRI}
\bibinfo{author}{Armato~III, S.~G.} et~al.
\newblock \bibinfo{title}{The lung image database consortium ({LIDC}) and image
  database resource initiative ({IDRI}): a completed reference database of lung
  nodules on {CT} scans}.
\newblock \emph{\bibinfo{journal}{Med. Phys.}} \textbf{\bibinfo{volume}{38}},
  \bibinfo{pages}{915--931} (\bibinfo{year}{2011}).

\bibitem{SLIVER07}
\bibinfo{author}{Heimann, T.} et~al.
\newblock \bibinfo{title}{Comparison and evaluation of methods for liver
  segmentation from {CT} datasets}.
\newblock \emph{\bibinfo{journal}{IEEE Trans. Med. Imaging}}
  \textbf{\bibinfo{volume}{28}}, \bibinfo{pages}{1251--1265}
  (\bibinfo{year}{2009}).

\bibitem{LUNA16}
\bibinfo{author}{Setio, A. A.~A.} et~al.
\newblock \bibinfo{title}{Validation, comparison, and combination of algorithms
  for automatic detection of pulmonary nodules in computed tomography images:
  the {LUNA16} challenge}.
\newblock \emph{\bibinfo{journal}{Med. Image Anal.}}
  \textbf{\bibinfo{volume}{42}}, \bibinfo{pages}{1--13} (\bibinfo{year}{2017}).

\bibitem{COVID19Seg}
\bibinfo{author}{Ma, J.} et~al.
\newblock \bibinfo{title}{Toward data-efficient learning: a benchmark for
  {COVID-19} {CT} lung and infection segmentation}.
\newblock \emph{\bibinfo{journal}{Med. Phys.}} \textbf{\bibinfo{volume}{48}},
  \bibinfo{pages}{1197--1210} (\bibinfo{year}{2021}).

\bibitem{CTPelvic1K}
\bibinfo{author}{Liu, P.} et~al.
\newblock \bibinfo{title}{Deep learning to segment pelvic bones: large-scale
  {CT} datasets and baseline models}.
\newblock \emph{\bibinfo{journal}{Int. J. Comput. Assist. Radiol. Surg.}}
  \textbf{\bibinfo{volume}{16}}, \bibinfo{pages}{749--756}
  (\bibinfo{year}{2021}).

\bibitem{Couinaud}
\bibinfo{author}{Tian, J.}, \bibinfo{author}{Liu, L.}, \bibinfo{author}{Shi,
  Z.} \& \bibinfo{author}{Xu, F.}
\newblock \bibinfo{title}{Automatic {Couinaud} segmentation from {CT} volumes
  on liver using {GLC-UNet}}.
\newblock In \emph{\bibinfo{booktitle}{Proc. International Workshop on Machine
  Learning in Medical Imaging}}, vol. 11861, pp. \bibinfo{pages}{274--282}
  (Springer, \bibinfo{year}{2019}).

\bibitem{DAPAtlas}
\bibinfo{author}{Jaus, A.} et~al.
\newblock \bibinfo{title}{Towards unifying anatomy segmentation: automated
  generation of a full-body {CT} dataset}.
\newblock In \emph{\bibinfo{booktitle}{Proc. IEEE International Conference on
  Image Processing}}, pp. \bibinfo{pages}{41--47} (IEEE, \bibinfo{year}{2024}).

\bibitem{FUMPE}
\bibinfo{author}{Masoudi, M.}, \bibinfo{author}{Pourreza, H.-R.},
  \bibinfo{author}{Saadatmand-Tarzjan, M.}, \bibinfo{author}{Eftekhari, N.},
  \bibinfo{author}{Zargar, F.~S.} \& \bibinfo{author}{Rad, M.~P.}
\newblock \bibinfo{title}{A new dataset of computed-tomography angiography
  images for computer-aided detection of pulmonary embolism}.
\newblock \emph{\bibinfo{journal}{Sci. Data}} \textbf{\bibinfo{volume}{5}},
  \bibinfo{pages}{180180} (\bibinfo{year}{2018}).

\bibitem{LNDb}
\bibinfo{author}{Pedrosa, J.} et~al.
\newblock \bibinfo{title}{{LNDb} challenge on automatic lung cancer patient
  management}.
\newblock \emph{\bibinfo{journal}{Med. Image Anal.}}
  \textbf{\bibinfo{volume}{70}}, \bibinfo{pages}{102027}
  (\bibinfo{year}{2021}).

\bibitem{MMWHS}
\bibinfo{author}{Zhuang, X.} et~al.
\newblock \bibinfo{title}{Evaluation of algorithms for multi-modality whole
  heart segmentation: an open-access grand challenge}.
\newblock \emph{\bibinfo{journal}{Med. Image Anal.}}
  \textbf{\bibinfo{volume}{58}}, \bibinfo{pages}{101537}
  (\bibinfo{year}{2019}).

\bibitem{PDDCA}
\bibinfo{author}{Raudaschl, P.~F.} et~al.
\newblock \bibinfo{title}{Evaluation of segmentation methods on head and neck
  {CT}: auto-segmentation challenge 2015}.
\newblock \emph{\bibinfo{journal}{Med. Phys.}} \textbf{\bibinfo{volume}{44}},
  \bibinfo{pages}{2020--2036} (\bibinfo{year}{2017}).

\bibitem{PARSE2022}
\bibinfo{author}{Luo, G.} et~al.
\newblock \bibinfo{title}{Efficient automatic segmentation for multi-level
  pulmonary arteries: the {PARSE} challenge}.
\newblock Preprint at
  {\urlstyle{same}\url{https://doi.org/10.48550/arXiv.2304.03708}}
  (\bibinfo{year}{2023}).

\bibitem{SEGA}
\bibinfo{author}{Radl, L.} et~al.
\newblock \bibinfo{title}{{AVT}: multicenter aortic vessel tree {CTA} dataset
  collection with ground truth segmentation masks}.
\newblock \emph{\bibinfo{journal}{Data Brief}} \textbf{\bibinfo{volume}{40}},
  \bibinfo{pages}{107801} (\bibinfo{year}{2022}).

\bibitem{SegRap2023}
\bibinfo{author}{Luo, X.} et~al.
\newblock \bibinfo{title}{{SegRap2023}: a benchmark of organs-at-risk and gross
  tumor volume segmentation for radiotherapy planning of nasopharyngeal
  carcinoma}.
\newblock \emph{\bibinfo{journal}{Med. Image Anal.}}
  \textbf{\bibinfo{volume}{101}}, \bibinfo{pages}{103447}
  (\bibinfo{year}{2025}).

\bibitem{SegTHOR}
\bibinfo{author}{Lambert, Z.}, \bibinfo{author}{Petitjean, C.},
  \bibinfo{author}{Dubray, B.} \& \bibinfo{author}{Kuan, S.}
\newblock \bibinfo{title}{{SegTHOR}: segmentation of thoracic organs at risk in
  {CT} images}.
\newblock In \emph{\bibinfo{booktitle}{Proc. International Conference on Image
  Processing Theory, Tools and Applications}}, pp. \bibinfo{pages}{1--6} (IEEE,
  \bibinfo{year}{2020}).

\bibitem{SegCQ500}
\bibinfo{author}{Spahr, A.}, \bibinfo{author}{St{\r{a}}hle, J.},
  \bibinfo{author}{Wang, C.} \& \bibinfo{author}{Kaijser, M.}
\newblock \bibinfo{title}{Label-efficient deep semantic segmentation of
  intracranial hemorrhages in {CT}-scans}.
\newblock \emph{\bibinfo{journal}{Front. Neuroimag.}}
  \textbf{\bibinfo{volume}{2}}, \bibinfo{pages}{1157565}
  (\bibinfo{year}{2023}).

\bibitem{NSCLCRadiomics}
\bibinfo{author}{Aerts, H. J. W.~L.} et~al.
\newblock \bibinfo{title}{Decoding tumour phenotype by noninvasive imaging
  using a quantitative radiomics approach}.
\newblock \emph{\bibinfo{journal}{Nat. Commun.}} \textbf{\bibinfo{volume}{5}},
  \bibinfo{pages}{4006} (\bibinfo{year}{2014}).

\bibitem{PleThora}
\bibinfo{author}{Kiser, K.~J.} et~al.
\newblock \bibinfo{title}{{PleThora}: pleural effusion and thoracic cavity
  segmentations in diseased lungs for benchmarking chest {CT} processing
  pipelines}.
\newblock \emph{\bibinfo{journal}{Med. Phys.}} \textbf{\bibinfo{volume}{47}},
  \bibinfo{pages}{5941--5952} (\bibinfo{year}{2020}).

\bibitem{WORD}
\bibinfo{author}{Luo, X.} et~al.
\newblock \bibinfo{title}{{WORD}: a large scale dataset, benchmark and clinical
  applicable study for abdominal organ segmentation from {CT} image}.
\newblock \emph{\bibinfo{journal}{Med. Image Anal.}}
  \textbf{\bibinfo{volume}{82}}, \bibinfo{pages}{102642}
  (\bibinfo{year}{2022}).

\bibitem{RAOS}
\bibinfo{author}{Luo, X.}, \bibinfo{author}{Li, Z.}, \bibinfo{author}{Zhang,
  S.}, \bibinfo{author}{Liao, W.} \& \bibinfo{author}{Wang, G.}
\newblock \bibinfo{title}{Rethinking abdominal organ segmentation ({RAOS}) in
  the clinical scenario: a robustness evaluation benchmark with challenging
  cases}.
\newblock In \emph{\bibinfo{booktitle}{Proc. International Conference on
  Medical Image Computing and Computer-Assisted Intervention}}, pp.
  \bibinfo{pages}{531--541} (MICCAI, \bibinfo{year}{2024}).

\bibitem{CTICH}
\bibinfo{author}{Hssayeni, M.}
\newblock \bibinfo{title}{Computed tomography images for intracranial
  hemorrhage detection and segmentation} (version 1.3.0).
\newblock \bibinfo{howpublished}{PhysioNet}
  {\urlstyle{same}\url{https://doi.org/10.13026/w8q8-ky94}}
  (\bibinfo{year}{2019}).

\bibitem{MediastinalLN}
\bibinfo{author}{Idris, T.} et~al.
\newblock \bibinfo{title}{Mediastinal lymph node quantification ({LNQ}):
  segmentation of heterogeneous {CT} data} (version 1).
\newblock \bibinfo{howpublished}{The Cancer Imaging Archive}
  {\urlstyle{same}\url{https://doi.org/10.7937/qvaz-ja09}}
  (\bibinfo{year}{2024}).

\bibitem{IRCAD}
\bibinfo{author}{Soler, L.} et~al.
\newblock \bibinfo{title}{{3D} image reconstruction for comparison of algorithm
  database: a patient-specific anatomical and medical image database}.
\newblock \bibinfo{type}{Technical report}, \bibinfo{institution}{IRCAD}
  {\urlstyle{same}\url{https://www.ircad.fr/research/data-sets/liver-segmentation-3d-ircadb-01/}}
  (\bibinfo{year}{2010}).

\bibitem{NSCLCRadiogenomics}
\bibinfo{author}{Bakr, S.} et~al.
\newblock \bibinfo{title}{A radiogenomic dataset of non-small cell lung
  cancer}.
\newblock \emph{\bibinfo{journal}{Sci. Data}} \textbf{\bibinfo{volume}{5}},
  \bibinfo{pages}{180202} (\bibinfo{year}{2018}).

\bibitem{PanTS}
\bibinfo{author}{Li, W.} et~al.
\newblock \bibinfo{title}{{PanTS}: the pancreatic tumor segmentation dataset}.
\newblock In \emph{\bibinfo{booktitle}{Proc. Advances in Neural Information
  Processing Systems}}, vol. 38, pp. \bibinfo{pages}{35806--35828} (Curran
  Associates, \bibinfo{year}{2025}).

\bibitem{PASTAGen30K}
\bibinfo{author}{Lei, W.} et~al.
\newblock \bibinfo{title}{A synthetic data-driven radiology foundation model
  for pan-tumor clinical diagnosis}.
\newblock Preprint at
  {\urlstyle{same}\url{https://doi.org/10.48550/arXiv.2502.06171}}
  (\bibinfo{year}{2025}).

\bibitem{ULS23}
\bibinfo{author}{de~Grauw, M.} et~al.
\newblock \bibinfo{title}{The {ULS23} challenge: a baseline model and benchmark
  dataset for {3D} universal lesion segmentation in computed tomography}.
\newblock \emph{\bibinfo{journal}{Med. Image Anal.}}
  \textbf{\bibinfo{volume}{102}}, \bibinfo{pages}{103525}
  (\bibinfo{year}{2025}).

\bibitem{CTLymphNodes}
\bibinfo{author}{Roth, H.~R.} et~al.
\newblock \bibinfo{title}{A new {2.5D} representation for lymph node detection
  using random sets of deep convolutional neural network observations}.
\newblock In \emph{\bibinfo{booktitle}{Proc. International Conference on
  Medical Image Computing and Computer-Assisted Intervention}}, pp.
  \bibinfo{pages}{520--527} (MICCAI, \bibinfo{year}{2014}).

\bibitem{HCCTACESeg}
\bibinfo{author}{Moawad, A.~W.} et~al.
\newblock \bibinfo{title}{Multimodality annotated hepatocellular carcinoma data
  set including pre- and post-{TACE} with imaging segmentation}.
\newblock \emph{\bibinfo{journal}{Sci. Data}} \textbf{\bibinfo{volume}{10}},
  \bibinfo{pages}{33} (\bibinfo{year}{2023}).

\bibitem{QUBIQ}
\bibinfo{author}{Li, H.~B.} et~al.
\newblock \bibinfo{title}{{QUBIQ}: uncertainty quantification for biomedical
  image segmentation challenge}.
\newblock Preprint at
  {\urlstyle{same}\url{https://doi.org/10.48550/arXiv.2405.18435}}
  (\bibinfo{year}{2024}).

\bibitem{PANORAMA}
\bibinfo{author}{Alves, N.} et~al.
\newblock \bibinfo{title}{Artificial intelligence and radiologists in
  pancreatic cancer detection using standard of care {CT} scans ({PANORAMA}):
  an international, paired, non-inferiority, confirmatory, observational
  study}.
\newblock \emph{\bibinfo{journal}{Lancet Oncol.}}
  \textbf{\bibinfo{volume}{27}}, \bibinfo{pages}{116--124}
  (\bibinfo{year}{2026}).

\bibitem{TotalSegLiverLesion}
\bibinfo{author}{Wasserthal, J.}
\newblock \bibinfo{title}{Training dataset for {TotalSegmentator} task
  "liver\_lesions"}.
\newblock \bibinfo{howpublished}{Zenodo}
  {\urlstyle{same}\url{https://doi.org/10.5281/zenodo.20272572}}
  (\bibinfo{year}{2026}).

\bibitem{MBHSeg}
\bibinfo{author}{Wang, D.} et~al.
\newblock \bibinfo{title}{{MICCAI2025} {MBH-Seg} challenge}.
\newblock \bibinfo{howpublished}{Zenodo}
  {\urlstyle{same}\url{https://doi.org/10.5281/zenodo.15094748}}
  (\bibinfo{year}{2025}).

\bibitem{PediatricCTSEG}
\bibinfo{author}{Jordan, P.} et~al.
\newblock \bibinfo{title}{Pediatric chest/abdomen/pelvic {CT} exams with expert
  organ contours ({Pediatric-CT-SEG})} (version 2).
\newblock \bibinfo{howpublished}{The Cancer Imaging Archive}
  {\urlstyle{same}\url{https://doi.org/10.7937/tcia.x0h0-1706}}
  (\bibinfo{year}{2021}).

\bibitem{VEELA2025}
\bibinfo{author}{Toprak, T.} et~al.
\newblock \bibinfo{title}{{VEELA} challenge - vessel extraction and extrication
  for liver analysis}.
\newblock In \emph{\bibinfo{booktitle}{Proc. IEEE International Workshop on
  Machine Learning for Signal Processing}}, pp. \bibinfo{pages}{1--6} (IEEE,
  \bibinfo{year}{2025}).

\bibitem{PENGWIN}
\bibinfo{author}{Liu, Y.} et~al.
\newblock \bibinfo{title}{Automatic pelvic fracture segmentation: a deep
  learning approach and benchmark dataset}.
\newblock \emph{\bibinfo{journal}{Front. Med. (Lausanne)}}
  \textbf{\bibinfo{volume}{12}}, \bibinfo{pages}{1511487}
  (\bibinfo{year}{2025}).

\bibitem{AdrenalACC}
\bibinfo{author}{Moawad, A.~W.} et~al.
\newblock \bibinfo{title}{Voxel-level segmentation of pathologically-proven
  adrenocortical carcinoma with {Ki-67} expression ({Adrenal-ACC-Ki67-Seg})}
  (version 1).
\newblock \bibinfo{howpublished}{The Cancer Imaging Archive}
  {\urlstyle{same}\url{https://doi.org/10.7937/1fpg-vm46}}
  (\bibinfo{year}{2023}).

\bibitem{ThreeDRAD}
\bibinfo{author}{Gai, X.}, \bibinfo{author}{Liu, J.}, \bibinfo{author}{Li, Y.},
  \bibinfo{author}{Meng, Z.}, \bibinfo{author}{Wu, J.} \& \bibinfo{author}{Liu,
  Z.}
\newblock \bibinfo{title}{{3D-RAD}: a comprehensive {3D} radiology med-{VQA}
  dataset with multi-temporal analysis and diverse diagnostic tasks}.
\newblock In \emph{\bibinfo{booktitle}{Proc. Advances in Neural Information
  Processing Systems}}, vol. 38, pp. \bibinfo{pages}{188177--188218} (Curran
  Associates, \bibinfo{year}{2025}).

\bibitem{FLARE2025}
\bibinfo{author}{{FLARE-MedFM}}.
\newblock \bibinfo{title}{{FLARE}-{Task5}-{MLLM}-{3D}}.
\newblock Hugging Face
  {\urlstyle{same}\url{https://huggingface.co/datasets/FLARE-MedFM/FLARE-Task5-MLLM-3D}}
  (\bibinfo{year}{2025}).

\bibitem{INSPECT}
\bibinfo{author}{Huang, S.-C.} et~al.
\newblock \bibinfo{title}{{INSPECT}: a multimodal dataset for patient outcome
  prediction of pulmonary embolisms}.
\newblock In \emph{\bibinfo{booktitle}{Proc. Advances in Neural Information
  Processing Systems}}, vol. 36, pp. \bibinfo{pages}{17742--17772} (Curran
  Associates, \bibinfo{year}{2023}).

\end{thebibliography}
\endgroup
\endgroup

\clearpage
\begingroup
    \color{black}
    \arrayrulecolor{black}
    \hypersetup{linkcolor=black}
    \renewcommand{\baselinestretch}{1}\selectfont
    \fontsize{9.5pt}{12pt}\selectfont
    \setlength{\parindent}{0pt}
    \setlength{\parskip}{0pt}
    \setlength{\LTpre}{14pt}
    \setlength{\LTpost}{0pt}
    \setlength{\LTleft}{0pt plus 1fill}
    \setlength{\LTright}{0pt plus 1fill}
    \setlength{\LTcapwidth}{\textwidth}
    \renewcommand{\arraystretch}{1.0}
    \emergencystretch=2em
    \captionsetup[table]{name=Supplementary Table,font=small,labelfont=bf,
        labelsep=tablepipe,justification=raggedright,singlelinecheck=false,skip=7pt}
\clearpage
\begin{longtable}{@{}p{0.055\linewidth}@{\hspace{0.02\linewidth}}>{\raggedright\arraybackslash}p{0.925\linewidth}@{}}
\caption[Prompts for report cleaning, question construction and evidence-mask selection.]{\textbf{Prompts for report cleaning, question construction and evidence-mask selection.} \textbf{a,} BIMCV report cleaning and atomic fact extraction. \textbf{b,} BIMCV question--answer construction. \textbf{c,d,} Supporting-mask assignment for DeepTumorVQA and AMOS-MM-VQA. Bracketed mappings and item fields denote runtime inputs; JSON specifies the requested output format.}\label{tab:preprocessing_prompts}\\
\toprule
\textbf{Part} & \textbf{Prompt and runtime inputs} \\
\midrule
\endfirsthead
\caption*{\textbf{Supplementary Table~\ref{tab:preprocessing_prompts} (continued)\enspace|\enspace Prompts for report cleaning, question construction and evidence-mask selection.}}\\
\toprule
\textbf{Part} & \textbf{Prompt and runtime inputs} \\
\midrule
\endhead
\midrule
\multicolumn{2}{r@{}}{\footnotesize Continued on next page}\\
\endfoot
\bottomrule
\endlastfoot
a & {\normalfont\bfseries BIMCV report cleaning}\par {\normalfont\small BIMCV-RG cleaning and fact extraction}\par\vspace{5pt}{\ttfamily You standardize a noisy English radiology report translated from Spanish into a current-CT report-generation target.\par\vspace{4pt} Produce:\newline 1. a compact list of atomic CT facts;\newline 2. one fluent cleaned report containing the same information.\par\vspace{4pt} Preserve current CT-visible positive findings, explicit negative findings, diagnostic impressions, uncertainty, anatomy, laterality, distribution, severity, devices, postoperative changes, and reported quantitative values.\par\vspace{4pt} Remove symptoms, medical history, physical examination, vital signs, laboratory or PCR results, treatment information, identifiers, administrative or signature text, acquisition technique, contrast status, reconstruction method, scan-protocol-only text, recommendations, follow-up instructions, and requests for clinical correlation.\par\vspace{4pt} Remove references to prior examinations and interval change, but retain the current imaging state.\par\vspace{4pt} Do not introduce a new finding that is unsupported by the report.} \\[8pt]
b & {\normalfont\bfseries BIMCV question construction}\par {\normalfont\small BIMCV-VQA generation}\par\vspace{5pt}{\ttfamily Generate a compact set of natural open-ended question-answer pairs from the provided atomic current-CT facts.\par\vspace{4pt} Use only the supplied facts.\par\vspace{4pt} If there are 1-2 facts, generate 1-2 pairs.\newline If there are 3-5 facts, generate 2-3 pairs.\newline If there are 6 or more facts, generate 3-5 pairs.\par\vspace{4pt} Prefer clinically informative and nonredundant questions.\par\vspace{4pt} Use normal open-ended questions beginning with What, Which, Where, or How.\par\vspace{4pt} Do not generate yes/no questions.\newline Do not generate multiple-choice or true/false questions.\par\vspace{4pt} Preserve laterality, anatomy, uncertainty, negation, numeric values, and units.\par\vspace{4pt} Do not add information absent from the cited facts.\par\vspace{4pt} Cite the fact IDs that directly support each answer.\newline Every answer must have at least one valid evidence ID.} \\[8pt]
c & {\normalfont\bfseries DeepTumorVQA mask matching}\par {\normalfont\small DeepTumorVQA grounding}\par\vspace{5pt}{\ttfamily Select CT masks needed to verify each answer.\par\vspace{4pt} Targets: [ID-to-mask mapping]\par\vspace{4pt} Each item: q=question, a=answer,\newline m=available mask IDs.\par\vspace{4pt} Rules:\newline - Output only IDs in m.\newline - Be strict: mentioned anatomy alone is not enough.\newline - Respect left/right; add both only when both are required.\newline - Organ measurement -\textgreater{} organ mask.\newline - Lesion existence/count/size/volume -\textgreater{} lesion mask; never replace missing lesion mask with organ mask.\newline - Multi-object comparison/sum -\textgreater{} all masks actually needed.\newline - Prefer specific anatomy over parent anatomy.\newline - Negative lesion answer does not require parent organ.\newline - If uncertain, choose fewer.\par\vspace{4pt} Return JSON only:\newline \{"r":[[ids],...]\}\par\vspace{4pt} One result per input item, same order.} \\[8pt]
d & {\normalfont\bfseries AMOS-MM-VQA mask matching}\par {\normalfont\small AMOS-MM-VQA grounding}\par\vspace{5pt}{\ttfamily Select CT masks needed to visually support each answer.\par\vspace{4pt} Targets: [ID-to-mask mapping]\par\vspace{4pt} Semantic synonyms and alternative medical wording count; exact string matching is not required.\par\vspace{4pt} Rules:\newline - Return all targets actually needed.\newline - Respect left/right.\newline - Add both sides only when bilateral involvement is explicit or clearly implied.\newline - Do not infer both sides from generic words such as kidney, renal or adrenal.\newline - Mentioned or nearby anatomy alone is not enough.\newline - Choose the anatomy providing direct visual evidence, not an organ merely affected as a downstream consequence.\newline - If none of the targets applies, return [].\newline - If uncertain, choose fewer.\par\vspace{4pt} Return JSON only:\newline \{"r":[[ids],...]\}\par\vspace{4pt} One result per input item, same order.} \\[8pt]
\end{longtable}

\clearpage
\begin{longtable}{@{}p{0.055\linewidth}@{\hspace{0.02\linewidth}}>{\raggedright\arraybackslash}p{0.925\linewidth}@{}}
\caption[Segmentation, lesion-screening and finding-grounded report generation templates.]{\textbf{Segmentation, lesion-screening and finding-grounded report generation templates.} All 42 entries are shown. Empty braces denote the requested target or finding description. \textbf{a,} Target-level segmentation. \textbf{b,c,} Presence questions and conditional grounding requests. \textbf{d,} Finding-level referring segmentation. \textbf{e,} Multiple-target screening. \textbf{f,} Finding-grounded and text-only report generation prompts. Numbering is local to each bank.}\label{tab:segmentation_prompts}\\
\toprule
\textbf{ID} & \textbf{Template} \\
\midrule
\endfirsthead
\caption*{\textbf{Supplementary Table~\ref{tab:segmentation_prompts} (continued)\enspace|\enspace Segmentation, lesion-screening and finding-grounded report generation templates.}}\\
\toprule
\textbf{ID} & \textbf{Template} \\
\midrule
\endhead
\midrule
\multicolumn{2}{r@{}}{\footnotesize Continued on next page}\\
\endfoot
\bottomrule
\endlastfoot
\multicolumn{2}{@{}p{\linewidth}@{}}{\textbf{a | Target-level segmentation}\par {\footnotesize \texttt{\detokenize{segmentation.pure_questions}}}}\\*[4pt]
1 & {\ttfamily \detokenize{Segment the {} in this CT scan.}} \\[4pt]
2 & {\ttfamily \detokenize{Please segment the {} in this CT volume.}} \\[4pt]
3 & {\ttfamily \detokenize{Generate a 3D mask for the {}.}} \\[4pt]
4 & {\ttfamily \detokenize{Provide the segmentation mask for the {}.}} \\[4pt]
5 & {\ttfamily \detokenize{Delineate the {} in this CT scan.}} \\[4pt]
6 & {\ttfamily \detokenize{Locate and segment the {}.}} \\[4pt]
\multicolumn{2}{@{}p{\linewidth}@{}}{\textbf{b | Single-target presence questions}\par {\footnotesize \texttt{\detokenize{segmentation.classification_questions}}}}\\*[4pt]
1 & {\ttfamily \detokenize{Is the {} present in this CT scan?}} \\[4pt]
2 & {\ttfamily \detokenize{Can you identify the {} in this CT scan?}} \\[4pt]
3 & {\ttfamily \detokenize{Check whether the {} is present in this CT scan.}} \\[4pt]
4 & {\ttfamily \detokenize{Determine whether the {} is present in this CT volume.}} \\[4pt]
5 & {\ttfamily \detokenize{Does this CT scan contain the {}?}} \\[4pt]
6 & {\ttfamily \detokenize{Can you detect the {} in this CT volume?}} \\[4pt]
7 & {\ttfamily \detokenize{Look for the {} in this CT scan and determine whether it is present.}} \\[4pt]
8 & {\ttfamily \detokenize{Assess whether the {} is visible in this CT scan.}} \\[4pt]
9 & {\ttfamily \detokenize{Is there evidence of the {} in this CT volume?}} \\[4pt]
10 & {\ttfamily \detokenize{Can the {} be found in this scan?}} \\[4pt]
\multicolumn{2}{@{}p{\linewidth}@{}}{\textbf{c | Conditional grounding requests}\par {\footnotesize \texttt{\detokenize{segmentation.classification_grounding_requests}}}}\\*[4pt]
1 & {\ttfamily \detokenize{If present, please provide its grounded mask.}} \\[4pt]
2 & {\ttfamily \detokenize{If found, also provide the corresponding segmentation mask.}} \\[4pt]
3 & {\ttfamily \detokenize{If it is present, return its 3D mask.}} \\[4pt]
4 & {\ttfamily \detokenize{If detected, provide its voxel-level segmentation.}} \\[4pt]
5 & {\ttfamily \detokenize{If visible, also ground it with a segmentation mask.}} \\[4pt]
6 & {\ttfamily \detokenize{If the target is present, please segment it.}} \\[4pt]
7 & {\ttfamily \detokenize{If identified, provide the corresponding grounded mask.}} \\[4pt]
8 & {\ttfamily \detokenize{If found, also delineate its spatial extent.}} \\[4pt]
9 & {\ttfamily \detokenize{If present, return the corresponding 3D segmentation.}} \\[4pt]
10 & {\ttfamily \detokenize{If detected, also provide its mask.}} \\[4pt]
\multicolumn{2}{@{}p{\linewidth}@{}}{\textbf{d | Finding-level referring segmentation}\par {\footnotesize \texttt{\detokenize{segmentation.referring_segmentation_questions}}}}\\*[4pt]
1 & {\ttfamily \detokenize{Segment the region described as: "{}"}} \\[4pt]
2 & {\ttfamily \detokenize{Generate a segmentation mask for the following finding: "{}"}} \\[4pt]
3 & {\ttfamily \detokenize{Locate and segment the area corresponding to this description: "{}"}} \\[4pt]
4 & {\ttfamily \detokenize{Provide the segmentation mask for the following text: "{}"}} \\[4pt]
5 & {\ttfamily \detokenize{Identify and segment the target based on this description: "{}"}} \\[4pt]
6 & {\ttfamily \detokenize{Segment the specific pathology or anatomy described here: "{}"}} \\[4pt]
7 & {\ttfamily \detokenize{Ground the following textual description in the 3D volume: "{}"}} \\[4pt]
8 & {\ttfamily \detokenize{Find and segment the region matching this observation: "{}"}} \\[4pt]
9 & {\ttfamily \detokenize{Create a mask for the region described by: "{}"}} \\[4pt]
10 & {\ttfamily \detokenize{Segment the target corresponding to the following report excerpt: "{}"}} \\[4pt]
\multicolumn{2}{@{}p{\linewidth}@{}}{\textbf{e | Multiple-target screening}\par {\footnotesize \texttt{\detokenize{screening_questions}}}}\\*[4pt]
\multicolumn{2}{@{}p{\linewidth}@{}}{\footnotesize \texttt{\detokenize{LiMT_Screening}}}\\*[3pt]
1 & {\ttfamily \detokenize{Screen the CT and {action} the liver and any potential abnormalities: liver cyst, liver hemangioma, metastatic liver cancer, or primary liver cancer.}} \\[4pt]
\multicolumn{2}{@{}p{\linewidth}@{}}{\footnotesize \texttt{\detokenize{AbdomenAtlas4Lesion_Screening}}}\\*[3pt]
1 & {\ttfamily \detokenize{Screen the CT and {action} any potential abnormalities: kidney lesion, liver lesion, pancreas lesion, or colon lesion.}} \\[4pt]
\multicolumn{2}{@{}p{\linewidth}@{}}{\footnotesize \texttt{\detokenize{ICH_Screening}}}\\*[3pt]
1 & {\ttfamily \detokenize{Screen the CT and {action} any potential intracranial hemorrhage: epidural hemorrhage, subdural hemorrhage, subarachnoid hemorrhage, intraparenchymal hemorrhage, or intraventricular hemorrhage.}} \\[4pt]
\multicolumn{2}{@{}p{\linewidth}@{}}{\footnotesize \texttt{\detokenize{R3_Screening}}}\\*[3pt]
1 & {\ttfamily \detokenize{Screen the CT and {action} any potential thyroid tumors: malignant thyroid tumor or benign thyroid tumor.}} \\[4pt]
\multicolumn{2}{@{}p{\linewidth}@{}}{\textbf{f | Finding-grounded and text-only report generation}\par {\footnotesize \texttt{\detokenize{rex_report_questions}}}}\\*[4pt]
\multicolumn{2}{@{}p{\linewidth}@{}}{\footnotesize Grounded}\\*[3pt]
1 & {\ttfamily \detokenize{Identify and segment all abnormal findings in the lung, pleura, trachea and bronchi.}} \\[4pt]
\multicolumn{2}{@{}p{\linewidth}@{}}{\footnotesize Text-only}\\*[3pt]
1 & {\ttfamily \detokenize{Identify all abnormal findings in the lung, pleura, trachea and bronchi.}} \\[4pt]
\end{longtable}

\par
\begin{longtable}{@{}p{0.055\linewidth}@{\hspace{0.02\linewidth}}>{\raggedright\arraybackslash}p{0.925\linewidth}@{}}
\caption[Question-answering and VQA-grounding templates.]{\textbf{Question-answering and VQA-grounding templates.} All 19 question--answer templates (\textbf{a--d}), 20 grounding requests (\textbf{e}) and 16 mask-answer templates (\textbf{f,g}) are shown. Braced fields specify runtime values. The answer template below each question bank applies to the indicated entries. Grounded VQA combines a base question with a grounding request and the corresponding mask-answer wording.}\label{tab:vqa_prompts}\\
\toprule
\textbf{ID} & \textbf{Template} \\
\midrule
\endfirsthead
\caption*{\textbf{Supplementary Table~\ref{tab:vqa_prompts} (continued)\enspace|\enspace Question-answering and VQA-grounding templates.}}\\
\toprule
\textbf{ID} & \textbf{Template} \\
\midrule
\endhead
\midrule
\multicolumn{2}{r@{}}{\footnotesize Continued on next page}\\
\endfoot
\bottomrule
\endlastfoot
\multicolumn{2}{@{}p{\linewidth}@{}}{\textbf{a | Standard multiple choice}\par {\footnotesize \texttt{\detokenize{vqa.std_mcq}}}}\\*[4pt]
1 & {\ttfamily \detokenize{{q}}\newline \detokenize{{opts}}\newline \detokenize{Answer with the full option.}} \\[4pt]
2 & {\ttfamily \detokenize{Based on the CT scan, {q}}\newline \detokenize{{opts}}\newline \detokenize{Answer:}} \\[4pt]
3 & {\ttfamily \detokenize{Question: {q}}\newline \detokenize{Options:}\newline \detokenize{{opts}}\newline \detokenize{Select the correct option.}} \\[4pt]
4 & {\ttfamily \detokenize{Looking at this medical image, which of the following is correct?}\newline \detokenize{{q}}\newline \detokenize{{opts}}} \\[4pt]
5 & {\ttfamily \detokenize{{q}}\newline \detokenize{{opts}}} \\[4pt]
\multicolumn{2}{@{}p{\linewidth}@{}}{\footnotesize Answer template for all entries: \texttt{\detokenize{{ans}}}.}\\[7pt]
\multicolumn{2}{@{}p{\linewidth}@{}}{\textbf{b | Multiple choice with reasoning}\par {\footnotesize \texttt{\detokenize{vqa.reasoning_mcq}}}}\\*[4pt]
1 & {\ttfamily \detokenize{{q}}\newline \detokenize{{opts}}\newline \detokenize{Select the answer and provide your reasoning based on the scan.}} \\[4pt]
2 & {\ttfamily \detokenize{{q}}\newline \detokenize{{opts}}\newline \detokenize{Explain the logic behind your choice.}} \\[4pt]
3 & {\ttfamily \detokenize{Question: {q}}\newline \detokenize{Options:}\newline \detokenize{{opts}}\newline \detokenize{Answer with explanation.}} \\[4pt]
4 & {\ttfamily \detokenize{Based on the findings in the imaging, {q}}\newline \detokenize{{opts}}\newline \detokenize{Give the answer and the reason.}} \\[4pt]
5 & {\ttfamily \detokenize{Analyze the CT image and answer: {q}}\newline \detokenize{{opts}}} \\[4pt]
\multicolumn{2}{@{}p{\linewidth}@{}}{\footnotesize Answer template for all entries: \texttt{\detokenize{{ans}. {reason}}}.}\\[7pt]
\multicolumn{2}{@{}p{\linewidth}@{}}{\textbf{c | Binary answers}\par {\footnotesize \texttt{\detokenize{vqa.binary}}}}\\*[4pt]
1 & {\ttfamily \detokenize{{q} (Yes/No)}} \\[4pt]
2 & {\ttfamily \detokenize{Based on the medical image, {q}? Yes or No?}} \\[4pt]
3 & {\ttfamily \detokenize{Please answer with Yes or No: {q}}} \\[4pt]
4 & {\ttfamily \detokenize{True or False: {q}}} \\[4pt]
\multicolumn{2}{@{}p{\linewidth}@{}}{\footnotesize Answer templates: entries 1--3, \texttt{\detokenize{{ans}}}; entry 4, \texttt{\detokenize{{ans_tf}}}.}\\[7pt]
\multicolumn{2}{@{}p{\linewidth}@{}}{\textbf{d | Open-ended answers}\par {\footnotesize \texttt{\detokenize{vqa.open_ended}}}}\\*[4pt]
1 & {\ttfamily \detokenize{{q}}} \\[4pt]
2 & {\ttfamily \detokenize{Question: {q}}\newline \detokenize{Answer:}} \\[4pt]
3 & {\ttfamily \detokenize{Based on the CT volume, {q}}} \\[4pt]
4 & {\ttfamily \detokenize{Look at the 3D CT scan and answer: {q}}} \\[4pt]
5 & {\ttfamily \detokenize{According to the 3D scan, {q}}} \\[4pt]
\multicolumn{2}{@{}p{\linewidth}@{}}{\footnotesize Answer template for all entries: \texttt{\detokenize{{ans}}}.}\\[7pt]
\multicolumn{2}{@{}p{\linewidth}@{}}{\textbf{e | Optional VQA grounding requests}\par {\footnotesize \texttt{\detokenize{vqa_grounding.requests}}}}\\*[4pt]
1 & {\ttfamily \detokenize{Ground your answer with the mask.}} \\[4pt]
2 & {\ttfamily \detokenize{Provide the grounded mask.}} \\[4pt]
3 & {\ttfamily \detokenize{Include the grounded mask.}} \\[4pt]
4 & {\ttfamily \detokenize{Give the grounded mask.}} \\[4pt]
5 & {\ttfamily \detokenize{Also provide the grounded mask.}} \\[4pt]
6 & {\ttfamily \detokenize{Please provide the grounded mask.}} \\[4pt]
7 & {\ttfamily \detokenize{Please include the grounded mask.}} \\[4pt]
8 & {\ttfamily \detokenize{Please give the grounded mask.}} \\[4pt]
9 & {\ttfamily \detokenize{Ground the answer with the mask.}} \\[4pt]
10 & {\ttfamily \detokenize{Ground your response with the mask.}} \\[4pt]
11 & {\ttfamily \detokenize{Provide the corresponding grounded mask.}} \\[4pt]
12 & {\ttfamily \detokenize{Include the corresponding grounded mask.}} \\[4pt]
13 & {\ttfamily \detokenize{Please respond with the grounded mask.}} \\[4pt]
14 & {\ttfamily \detokenize{Please also provide the grounded mask.}} \\[4pt]
15 & {\ttfamily \detokenize{Please ground your answer with the mask.}} \\[4pt]
16 & {\ttfamily \detokenize{Provide a segmentation mask to ground your answer.}} \\[4pt]
17 & {\ttfamily \detokenize{Include a segmentation mask to ground your answer.}} \\[4pt]
18 & {\ttfamily \detokenize{Please provide a segmentation mask to ground your answer.}} \\[4pt]
19 & {\ttfamily \detokenize{Please accompany your answer with the grounded mask.}} \\[4pt]
20 & {\ttfamily \detokenize{Please accompany your response with the grounded mask.}} \\[4pt]
\multicolumn{2}{@{}p{\linewidth}@{}}{\textbf{f | Single-mask answer wording}\par {\footnotesize \texttt{\detokenize{vqa_grounding.single_mask_answers}}}}\\*[4pt]
1 & {\ttfamily \detokenize{{grounded} is the grounded mask.}} \\[4pt]
2 & {\ttfamily \detokenize{{grounded} is the corresponding mask.}} \\[4pt]
3 & {\ttfamily \detokenize{{grounded} is the grounding mask.}} \\[4pt]
4 & {\ttfamily \detokenize{{grounded} is the corresponding grounded mask.}} \\[4pt]
5 & {\ttfamily \detokenize{The grounded mask is {grounded}.}} \\[4pt]
6 & {\ttfamily \detokenize{The corresponding mask is {grounded}.}} \\[4pt]
7 & {\ttfamily \detokenize{Grounded mask: {grounded}.}} \\[4pt]
8 & {\ttfamily \detokenize{Grounding: {grounded}.}} \\[4pt]
\multicolumn{2}{@{}p{\linewidth}@{}}{\textbf{g | Multiple-mask answer wording}\par {\footnotesize \texttt{\detokenize{vqa_grounding.multiple_mask_answers}}}}\\*[4pt]
1 & {\ttfamily \detokenize{{grounded} are the grounded masks.}} \\[4pt]
2 & {\ttfamily \detokenize{{grounded} are the corresponding masks.}} \\[4pt]
3 & {\ttfamily \detokenize{{grounded} are the grounding masks.}} \\[4pt]
4 & {\ttfamily \detokenize{{grounded} are the corresponding grounded masks.}} \\[4pt]
5 & {\ttfamily \detokenize{The grounded masks are {grounded}.}} \\[4pt]
6 & {\ttfamily \detokenize{The corresponding masks are {grounded}.}} \\[4pt]
7 & {\ttfamily \detokenize{Grounded masks: {grounded}.}} \\[4pt]
8 & {\ttfamily \detokenize{Grounding: {grounded}.}} \\[4pt]
\end{longtable}

\clearpage
\begin{longtable}{@{}p{0.055\linewidth}@{\hspace{0.02\linewidth}}>{\raggedright\arraybackslash}p{0.925\linewidth}@{}}
\caption[Dataset-specific report templates for five source datasets.]{\textbf{Dataset-specific report templates for five source datasets.} All 60 templates from the AMOS-MM (\textbf{a}), RadGenome (\textbf{b}), Merlin (\textbf{c}), AbdomenAtlas (\textbf{d}) and FLARE-2025 (\textbf{e}) banks are shown. Region-list fields specify the annotated anatomical regions requested in the report. INSPECT and BIMCV templates are provided in Supplementary Table~\ref{tab:grounded_report_prompts}. Numbering is local to each bank.}\label{tab:report_prompts}\\
\toprule
\textbf{ID} & \textbf{Template} \\
\midrule
\endfirsthead
\caption*{\textbf{Supplementary Table~\ref{tab:report_prompts} (continued)\enspace|\enspace Dataset-specific report templates for five source datasets.}}\\
\toprule
\textbf{ID} & \textbf{Template} \\
\midrule
\endhead
\midrule
\multicolumn{2}{r@{}}{\footnotesize Continued on next page}\\
\endfoot
\bottomrule
\endlastfoot
\multicolumn{2}{@{}p{\linewidth}@{}}{\textbf{a | AMOS-MM}\par {\footnotesize \texttt{\detokenize{report_generation.AMOS-MM}}}}\\*[4pt]
1 & {\ttfamily \detokenize{ Provide a comprehensive radiological narrative for the following regions: {region_list}.}} \\[4pt]
2 & {\ttfamily \detokenize{ Generate a detailed findings report covering the {region_list}, detailing morphology and any lesions found.}} \\[4pt]
3 & {\ttfamily \detokenize{ Describe the radiological appearance across the {region_list} in a continuous, structured format.}} \\[4pt]
4 & {\ttfamily \detokenize{ What are the detailed clinical observations for the organs and tissues in the {region_list}?}} \\[4pt]
5 & {\ttfamily \detokenize{ Deliver a full account of the findings for the following anatomical areas: {region_list}.}} \\[4pt]
6 & {\ttfamily \detokenize{ Provide an in-depth assessment of the {region_list}, focusing on organ size, shape, and parenchymal density.}} \\[4pt]
7 & {\ttfamily \detokenize{ Write a structured findings section for the 3D CT scan, including {region_list}.}} \\[4pt]
8 & {\ttfamily \detokenize{ Outline the visual characteristics and enhancement patterns observed in the {region_list}.}} \\[4pt]
9 & {\ttfamily \detokenize{ Based on this scan, provide a systematic description of findings within the {region_list}.}} \\[4pt]
10 & {\ttfamily \detokenize{ Construct a detailed report summarizing the status of the {region_list} as captured in this volume.}} \\[4pt]
\multicolumn{2}{@{}p{\linewidth}@{}}{\textbf{b | RadGenome}\par {\footnotesize \texttt{\detokenize{report_generation.Radgenome}}}}\\*[4pt]
1 & {\ttfamily \detokenize{ Conduct a structural evaluation of the {region_list} based on the CT imaging data.}} \\[4pt]
2 & {\ttfamily \detokenize{ Report the anatomical findings for the following regions: {region_list}.}} \\[4pt]
3 & {\ttfamily \detokenize{ Analyze the {region_list} for any pathological, structural, or sequelae changes.}} \\[4pt]
4 & {\ttfamily \detokenize{ Provide a technical description of the status and integrity of the {region_list}.}} \\[4pt]
5 & {\ttfamily \detokenize{ What are the observations regarding the contours and dimensions of the {region_list}?}} \\[4pt]
6 & {\ttfamily \detokenize{ Evaluate the components of the {region_list} as visualized in this study.}} \\[4pt]
7 & {\ttfamily \detokenize{ Characterize the anatomical appearance of the {region_list} in this scan.}} \\[4pt]
8 & {\ttfamily \detokenize{ Detail the findings for the {region_list}, focusing on their internal structures and boundaries.}} \\[4pt]
9 & {\ttfamily \detokenize{ Provide a systematic assessment of the {region_list} in the specified imaging windows.}} \\[4pt]
10 & {\ttfamily \detokenize{ What is the radiological interpretation for the following anatomical sections: {region_list}?}} \\[4pt]
\multicolumn{2}{@{}p{\linewidth}@{}}{\textbf{c | Merlin}\par {\footnotesize \texttt{\detokenize{report_generation.Merlin}}}}\\*[4pt]
1 & {\ttfamily \detokenize{ Assess the radiological status of the following: {region_list}.}} \\[4pt]
2 & {\ttfamily \detokenize{ Provide a concise clinical summary of findings for the {region_list}.}} \\[4pt]
3 & {\ttfamily \detokenize{ What is the condition of the following anatomical structures: {region_list}?}} \\[4pt]
4 & {\ttfamily \detokenize{ Report the imaging observations for the {region_list} in a systematic manner.}} \\[4pt]
5 & {\ttfamily \detokenize{ Summarize the status of the {region_list}, noting any significant abnormalities or normal findings.}} \\[4pt]
6 & {\ttfamily \detokenize{ Document the clinical status for each of the following areas: {region_list}.}} \\[4pt]
7 & {\ttfamily \detokenize{ Provide a structured overview of the scan findings for the {region_list}.}} \\[4pt]
8 & {\ttfamily \detokenize{ Identify and state any notable findings within the {region_list}.}} \\[4pt]
9 & {\ttfamily \detokenize{ What does the 3D CT study reveal regarding the {region_list}?}} \\[4pt]
10 & {\ttfamily \detokenize{ Give a brief and professional assessment of the following regions: {region_list}.}} \\[4pt]
\multicolumn{2}{@{}p{\linewidth}@{}}{\textbf{d | AbdomenAtlas}\par {\footnotesize \texttt{\detokenize{report_generation.Abdomenatlas}}}}\\*[4pt]
1 & {\ttfamily \detokenize{Provide a quantitative analysis of the abdominal organs, reporting volumes in cc and mean HU values.}} \\[4pt]
2 & {\ttfamily \detokenize{Generate a morphometric report detailing the size and density of the pancreas, spleen, liver, and kidneys.}} \\[4pt]
3 & {\ttfamily \detokenize{Extract the volumetric measurements and attenuation characteristics for the major abdominal organs.}} \\[4pt]
4 & {\ttfamily \detokenize{List the quantitative metrics (volume and mean HU) for the abdominal anatomy and summarize the impression.}} \\[4pt]
5 & {\ttfamily \detokenize{Report the objective CT measurements for the patient, including organ volumes and any lesion dimensions.}} \\[4pt]
6 & {\ttfamily \detokenize{Analyze the CT scan to provide a detailed list of organ volumes and densities, noting any enlargements.}} \\[4pt]
7 & {\ttfamily \detokenize{Create a data-driven report focusing on the Hounsfield Unit statistics and volumetric data of the abdomen.}} \\[4pt]
8 & {\ttfamily \detokenize{Identify and measure the volume and density of the liver, spleen, pancreas, and kidneys.}} \\[4pt]
9 & {\ttfamily \detokenize{Provide a technical assessment of the abdominal organs with precise volumetric and densitometric data.}} \\[4pt]
10 & {\ttfamily \detokenize{Detail the dimensions and mean attenuation values for the abdominal structures and identified lesions.}} \\[4pt]
11 & {\ttfamily \detokenize{Perform a quantitative volumetric assessment of the abdominal viscera and summarize the findings.}} \\[4pt]
12 & {\ttfamily \detokenize{Write a report containing only the objective measurements (Volume/HU) and a concluding impression.}} \\[4pt]
13 & {\ttfamily \detokenize{Calculate and report the size and density parameters for the pancreas, liver, spleen, and kidneys.}} \\[4pt]
14 & {\ttfamily \detokenize{Assess the abdominal cavity quantitatively, specifying the volume and mean HU for each major organ.}} \\[4pt]
15 & {\ttfamily \detokenize{Generate a structured report highlighting organ volumes, HU deviations, and lesion measurements.}} \\[4pt]
16 & {\ttfamily \detokenize{Provide the volumetric segmentation statistics and density analysis for the abdominal CT.}} \\[4pt]
17 & {\ttfamily \detokenize{Report the calculated volumes and mean attenuation values for the visceral organs.}} \\[4pt]
18 & {\ttfamily \detokenize{Summarize the quantitative features of the abdomen, including total kidney volume and liver lesion sizes.}} \\[4pt]
19 & {\ttfamily \detokenize{Present the morphometric data of the patient's abdominal organs, followed by a diagnostic impression.}} \\[4pt]
20 & {\ttfamily \detokenize{Detail the quantitative findings, specifically focusing on organ volumes (cc) and radiodensity (HU).}} \\[4pt]
\multicolumn{2}{@{}p{\linewidth}@{}}{\textbf{e | FLARE-2025}\par {\footnotesize \texttt{\detokenize{report_generation.FLARE-2025}}}}\\*[4pt]
1 & {\ttfamily \detokenize{ Systematically evaluate the following systems and regions for any evidence of abnormality: {region_list}.}} \\[4pt]
2 & {\ttfamily \detokenize{ Provide a structured report assessing the morphology and density of the {region_list}.}} \\[4pt]
3 & {\ttfamily \detokenize{ Are there any significant findings in the following areas: {region_list}? Please describe.}} \\[4pt]
4 & {\ttfamily \detokenize{ Report on the status of the {region_list}, noting any pathological or normal features identified.}} \\[4pt]
5 & {\ttfamily \detokenize{ Describe the appearance of the {region_list} focusing on size, morphology, and regularity.}} \\[4pt]
6 & {\ttfamily \detokenize{ Summarize the imaging observations for the following anatomical groups: {region_list}.}} \\[4pt]
7 & {\ttfamily \detokenize{ Examine the {region_list} and state the findings for each system.}} \\[4pt]
8 & {\ttfamily \detokenize{ Provide the findings for the structures within the {region_list} regarding their parenchymal integrity.}} \\[4pt]
9 & {\ttfamily \detokenize{ Check for abnormalities in the {region_list} and provide a standardized findings report.}} \\[4pt]
10 & {\ttfamily \detokenize{ Document the radiological status of the {region_list} in this CT volume.}} \\[4pt]
\end{longtable}

\par
\begin{longtable}{@{}p{0.055\linewidth}@{\hspace{0.02\linewidth}}>{\raggedright\arraybackslash}p{0.925\linewidth}@{}}
\caption[INSPECT, BIMCV and anatomy-grounded report generation templates.]{\textbf{INSPECT, BIMCV and anatomy-grounded report generation templates.} \textbf{a,b,} The 20 INSPECT and 5 BIMCV report instructions. \textbf{c,} All 40 RadGenome anatomy-grounded report prompts, jointly requesting masks and report text in a specified anatomical order. Region-order fields specify the requested sequence of anatomical regions. Numbering is local to each bank.}\label{tab:grounded_report_prompts}\\
\toprule
\textbf{ID} & \textbf{Template} \\
\midrule
\endfirsthead
\caption*{\textbf{Supplementary Table~\ref{tab:grounded_report_prompts} (continued)\enspace|\enspace INSPECT, BIMCV and anatomy-grounded report generation templates.}}\\
\toprule
\textbf{ID} & \textbf{Template} \\
\midrule
\endhead
\midrule
\multicolumn{2}{r@{}}{\footnotesize Continued on next page}\\
\endfoot
\bottomrule
\endlastfoot
\multicolumn{2}{@{}p{\linewidth}@{}}{\textbf{a | INSPECT}\par {\footnotesize \texttt{\detokenize{report_generation.INSPECT}}}}\\*[4pt]
1 & {\ttfamily \detokenize{Generate a numbered radiological impression for this CT scan.}} \\[3pt]
2 & {\ttfamily \detokenize{Write the diagnostic impression for this CTPA as a numbered list.}} \\[3pt]
3 & {\ttfamily \detokenize{Evaluate the scan for pulmonary embolism and list all findings in a numbered impression.}} \\[3pt]
4 & {\ttfamily \detokenize{Provide a structured, numbered clinical impression based on this CT volume.}} \\[3pt]
5 & {\ttfamily \detokenize{Summarize the CTPA findings into a formal, numbered impression.}} \\[3pt]
6 & {\ttfamily \detokenize{Generate the final IMPRESSION section for this scan in a standard numbered format.}} \\[3pt]
7 & {\ttfamily \detokenize{Detail the PE status and any incidental findings in a numbered impression.}} \\[3pt]
8 & {\ttfamily \detokenize{Write a concise, numbered diagnostic summary for this chest CT.}} \\[3pt]
9 & {\ttfamily \detokenize{Formulate the clinical impression for this scan as a numbered list.}} \\[3pt]
10 & {\ttfamily \detokenize{Provide the final radiological impression, noting all findings in numerical order.}} \\[3pt]
11 & {\ttfamily \detokenize{Assess the CT volume and generate a numbered list of diagnostic impressions.}} \\[3pt]
12 & {\ttfamily \detokenize{Summarize all primary and incidental findings into a numbered CTPA impression.}} \\[3pt]
13 & {\ttfamily \detokenize{Draft the IMPRESSION section, listing all radiological findings systematically.}} \\[3pt]
14 & {\ttfamily \detokenize{Generate a comprehensive, numbered impression for this thoracic CT.}} \\[3pt]
15 & {\ttfamily \detokenize{Provide a standard numbered impression evaluating the pulmonary vasculature and secondary findings.}} \\[3pt]
16 & {\ttfamily \detokenize{List the diagnostic conclusions for this CTPA in a numbered impression format.}} \\[3pt]
17 & {\ttfamily \detokenize{Write the final numbered impression, including any clinical recommendations or summaries.}} \\[3pt]
\pagebreak
18 & {\ttfamily \detokenize{Evaluate this 3D CT and summarize the impression in a numbered list.}} \\[3pt]
19 & {\ttfamily \detokenize{Provide a formal, numbered summary of all relevant findings in this scan.}} \\[3pt]
20 & {\ttfamily \detokenize{Generate the clinical impression, systematically listing all thoracic and abdominal findings.}} \\[3pt]
\multicolumn{2}{@{}p{\linewidth}@{}}{\textbf{b | BIMCV}\par {\footnotesize \texttt{\detokenize{report_generation.BIMCV}}}}\\*[4pt]
1 & {\ttfamily \detokenize{Generate a radiology report for this CT scan.}} \\[3pt]
2 & {\ttfamily \detokenize{Please write a radiology report for the given CT volume.}} \\[3pt]
3 & {\ttfamily \detokenize{Describe the radiological findings in this 3D CT scan.}} \\[3pt]
4 & {\ttfamily \detokenize{Provide a concise radiology report for this CT volume.}} \\[3pt]
5 & {\ttfamily \detokenize{Generate a clinical imaging report based on the given 3D CT scan.}} \\[3pt]
\multicolumn{2}{@{}p{\linewidth}@{}}{\textbf{c | RadGenome anatomy-grounded report generation}\par {\footnotesize \texttt{\detokenize{radgenome_grounded_report_questions}}}}\\*[4pt]
1 & {\ttfamily \detokenize{Please analyze the CT scan and generate segmentation masks alongside a descriptive report. I need the findings organized in this specific flow: {region_order}.}} \\[3pt]
2 & {\ttfamily \detokenize{Starting with {region_order}, please segment the anatomical structures in this 3D CT volume and provide the corresponding radiology narratives for each.}} \\[3pt]
3 & {\ttfamily \detokenize{I'd like a detailed anatomical segmentation and a written summary for this CT study. Please process the regions according to this sequence: {region_order}.}} \\[3pt]
4 & {\ttfamily \detokenize{For the provided CT data, can you output both the segmentation masks and the clinical report? It}\textnormal{\textquoteright}\detokenize{s important to follow the order of: {region_order}.}} \\[3pt]
5 & {\ttfamily \detokenize{Using the 3D CT volume, please perform a region-by-region segmentation and descriptive write-up, adhering to the structure: {region_order}.}} \\[3pt]
6 & {\ttfamily \detokenize{Could you generate the anatomical masks and report text from this CT? I need the output to follow the progression from {region_order}.}} \\[3pt]
7 & {\ttfamily \detokenize{Run an anatomical segmentation on this CT dataset and draft a findings report. Please ensure every part follows the requested order: {region_order}.}} \\[3pt]
8 & {\ttfamily \detokenize{I need the CT findings and masks delivered in a structured format. Please follow this anatomical priority: {region_order}.}} \\[3pt]
9 & {\ttfamily \detokenize{Map out the segmentation masks for the CT scan and write a narrative report for each segment, following the list: {region_order}.}} \\[3pt]
10 & {\ttfamily \detokenize{From this 3D CT, I need a combined output of segmented masks and diagnostic text. Please structure it by: {region_order}.}} \\[3pt]
11 & {\ttfamily \detokenize{Can you segment the following regions in the CT scan}\textnormal{\textemdash}\detokenize{{region_order}}\textnormal{\textemdash}\detokenize{and provide a clinical interpretation for each in that exact order?}} \\[3pt]
12 & {\ttfamily \detokenize{Generate a segment-by-segment report and mask set from this CT volume, strictly adhering to the specified sequence: {region_order}.}} \\[3pt]
13 & {\ttfamily \detokenize{I'm looking for a radiology report paired with segmentation masks. Please arrange the findings and masks based on: {region_order}.}} \\[3pt]
14 & {\ttfamily \detokenize{Process this CT study to yield anatomical segmentations and narrative descriptions, ensuring the workflow follows: {region_order}.}} \\[3pt]
15 & {\ttfamily \detokenize{Please extract the anatomical masks and corresponding report entries from this CT, keeping the order consistent with: {region_order}.}} \\[3pt]
16 & {\ttfamily \detokenize{For this CT examination, deliver a set of segmentation masks and the associated report, structured according to the list: {region_order}.}} \\[3pt]
17 & {\ttfamily \detokenize{I need a comprehensive analysis of the CT scan. Provide masks and report text, following the anatomical layout of {region_order}.}} \\[3pt]
18 & {\ttfamily \detokenize{Take the 3D CT data and generate a structured report with masks. Please stick to this sequence for the output: {region_order}.}} \\[3pt]
\pagebreak
19 & {\ttfamily \detokenize{Analyze the CT volume and provide localized segmentation and text descriptions. The order of operations should be: {region_order}.}} \\[3pt]
20 & {\ttfamily \detokenize{Could you provide a radiology report and mask output for this CT? Please make sure the structures are addressed as: {region_order}.}} \\[3pt]
21 & {\ttfamily \detokenize{Using the provided CT, create a detailed map of masks and report text. The sequence of interest is {region_order}.}} \\[3pt]
22 & {\ttfamily \detokenize{I'd appreciate it if you could segment the CT anatomy and write the report findings, following the specific order of {region_order}.}} \\[3pt]
23 & {\ttfamily \detokenize{Please handle the CT segmentation and report generation. I need it organized structure-by-structure: {region_order}.}} \\[3pt]
24 & {\ttfamily \detokenize{Your task is to segment the CT and write a descriptive summary. Please follow the anatomical sequence: {region_order}.}} \\[3pt]
25 & {\ttfamily \detokenize{Can you provide masks and descriptive sentences for the anatomical regions in this CT scan? The desired order is {region_order}.}} \\[3pt]
26 & {\ttfamily \detokenize{Generate an anatomical segmentation for this 3D CT. The masks and report text should be presented in the order of {region_order}.}} \\[3pt]
27 & {\ttfamily \detokenize{For the input CT, I need a mask-based analysis and a written report. Please follow this specific path: {region_order}.}} \\[3pt]
28 & {\ttfamily \detokenize{Perform a region-specific analysis of this CT volume. Produce masks and text findings following: {region_order}.}} \\[3pt]
29 & {\ttfamily \detokenize{I need a clinical report and anatomical masks derived from this CT. Ensure the segments follow the list: {region_order}.}} \\[3pt]
30 & {\ttfamily \detokenize{Working with the provided CT, please output masks and findings descriptions in the following anatomical order: {region_order}.}} \\[3pt]
31 & {\ttfamily \detokenize{Generate both the segmentation masks and the radiology report text. Please align your output with: {region_order}.}} \\[3pt]
32 & {\ttfamily \detokenize{I need the CT dataset processed into anatomical masks and descriptive text. Follow this precise order: {region_order}.}} \\[3pt]
33 & {\ttfamily \detokenize{For the given CT, provide segmentation and matching narrative reports, structured as: {region_order}.}} \\[3pt]
34 & {\ttfamily \detokenize{Please produce anatomical masks and the corresponding imaging report, maintaining the sequence: {region_order}.}} \\[3pt]
35 & {\ttfamily \detokenize{Extract the segmentation results and write the report for this CT, following the structural sequence: {region_order}.}} \\[3pt]
36 & {\ttfamily \detokenize{Can you analyze the CT scan and provide masks and clinical text for each region? Please follow the sequence: {region_order}.}} \\[3pt]
37 & {\ttfamily \detokenize{Deliver a report and segmentation masks from the CT scan. The structures must be analyzed in this order: {region_order}.}} \\[3pt]
38 & {\ttfamily \detokenize{I'm requesting a mask-based segmentation and a clinical report for this CT scan, ordered by: {region_order}.}} \\[3pt]
39 & {\ttfamily \detokenize{Please segment the anatomy and draft the findings for this CT study, following the specified list of {region_order}.}} \\[3pt]
40 & {\ttfamily \detokenize{Process the 3D CT to output masks and a descriptive report, ensuring the sections follow: {region_order}.}} \\[3pt]
\end{longtable}

\clearpage
\begin{longtable}{@{}p{0.055\linewidth}@{\hspace{0.02\linewidth}}>{\raggedright\arraybackslash}p{0.925\linewidth}@{}}
\caption[Clinical report cleaning, translation and anatomical structuring.]{\textbf{Clinical report cleaning, translation and anatomical structuring.} \textbf{a,} Prompt template for Chinese reports from centre~5. \textbf{b,} Thoracic and abdominal region lists. Placeholders denote runtime inputs; the JSON block specifies the requested output format.}\label{tab:clinical_report_prompt}\\
\toprule
\textbf{Part} & \textbf{Prompt and runtime inputs} \\
\midrule
\endfirsthead
\caption*{\textbf{Supplementary Table~\ref{tab:clinical_report_prompt} (continued)\enspace|\enspace Clinical report cleaning, translation and anatomical structuring.}}\\
\toprule
\textbf{Part} & \textbf{Prompt and runtime inputs} \\
\midrule
\endhead
\midrule
\multicolumn{2}{r@{}}{\footnotesize Continued on next page}\\
\endfoot
\bottomrule
\endlastfoot
a & {\normalfont\bfseries Clinical report processing}\par {\normalfont\small Cleaning, translation and anatomical structuring template}\par\vspace{5pt}{\ttfamily You are an expert radiology report editor.\par\vspace{4pt} Clean, translate and structure the provided Chinese CT report into professional English.\par\vspace{4pt} Instructions:\newline - Remove patient identifiers, administrative information, non-imaging clinical details, acquisition protocols, recommendations and prior-study comparisons.\newline - Preserve current CT findings, diagnostic impressions, anatomy, laterality, morphology, severity, uncertainty and all reported measurements.\newline - Translate using standardized radiological terminology without adding unsupported findings.\newline - Assign each finding to the most appropriate anatomical region from the provided list.\newline - Consolidate findings and impressions within each region, avoiding unnecessary repetition.\newline - Preserve explicit normal findings; do not infer normality from an unmentioned region.\newline - Return valid JSON containing the region-wise descriptions.\par\vspace{4pt} Required regions:\newline \{region\_list\}\par\vspace{4pt} Original Chinese report:\newline \{original\_report\}\par\vspace{4pt} Output:\newline \{\newline \hspace*{1.0em}"process": \{\newline \hspace*{2.0em}"\textless{}region\_name\textgreater{}": "\textless{}English description\textgreater{}"\newline \hspace*{1.0em}\}\newline \}} \\[8pt]
b & {\normalfont\bfseries Region-list values}\par {\normalfont\small Chest and abdomen use their corresponding list}\par\vspace{5pt}{\ttfamily Chest (10): mediastinum, abdomen, lung, bone, heart, trachea and bronchi, esophagus, pleura, breast, thyroid.\par\vspace{4pt} Abdomen (13): lower thorax, liver and biliary tree, gallbladder, spleen, pancreas, adrenal glands, kidneys and ureters, gastrointestinal tract, peritoneum, pelvic, vasculature, lymph nodes, musculoskeletal.}\par\vspace{6pt}{\normalfont The relevant list populates \texttt{\detokenize{{region_list}}}; the source report populates \texttt{\detokenize{{original_report}}}. Region-wise English descriptions are concatenated to form the structured whole-report reference.} \\[8pt]
\end{longtable}
\clearpage
\endgroup

\end{document}